\documentclass[english,ngerman]{KITreprt}
\usepackage{url}
\usepackage{siunitx}
\usepackage{paralist}
\usepackage{bm}
\usepackage[section]{placeins}
\usepackage[export]{adjustbox}
\usepackage[english, boxed]{algorithm2e}
\usepackage[multiple]{footmisc}
\usepackage{caption}
\usepackage{float}
\usepackage{booktabs}
\usetikzlibrary{positioning}

\newcommand{\tvspace}[1]{\renewcommand{\arraystretch}{#1}}
\newcommand{\thspace}[1]{\renewcommand{\tabcolsep}{#1}}

\definecolor{red}{RGB}{246,0,1}
\definecolor{green}{RGB}{26,111,0}
\definecolor{blue}{RGB}{0,0,255}

\hypersetup{
  pdfauthor={Matthias Plappert},
  pdftitle={Classification of Human Whole-Body Motion using Hidden Markov Models},
  pdfsubject={Not set},
  pdfkeywords={Not set}
}

\renewcommand{\myname}{Matthias Plappert}
\renewcommand{\mythesis}{\bachelorsthesis}
\renewcommand{\mytitle}{Classification of Human Whole-Body Motion using Hidden Markov Models}
\renewcommand{\myshorttitle}{}
\renewcommand{\timestart}{May 1\textsuperscript{st}, 2015}
\renewcommand{\timeend}{August 31\textsuperscript{th}, 2015}
\newcommand{\advisor}{Dipl.-Inform. Christian Mandery}
\newcommand{\referee}{Prof. Dr.-Ing. Tamim Asfour}
\newcommand{\refereetwo}{Prof. Dr.-Ing. Rüdiger Dillmann}

\graphicspath{{./images/}}

\renewcommand{\vec}[1]{\mathbf{#1}}
\newcommand{\vecsym}[1]{\boldsymbol{#1}}
\newcommand{\abs}[1]{\ensuremath{\left\vert#1\right\vert}}
\newcommand{\norm}[1]{\ensuremath{\lVert#1\rVert}}

\DeclareMathOperator*{\argmax}{arg\,max}
\DeclareMathOperator*{\precision}{precision}
\DeclareMathOperator*{\recall}{recall}
\DeclareMathOperator*{\accuracy}{accuracy}
\DeclareMathOperator*{\tp}{TP}
\DeclareMathOperator*{\tn}{TN}
\DeclareMathOperator*{\fp}{FP}
\DeclareMathOperator*{\fn}{FN}

\begin{document}
\selectlanguage{english}
\maketitle

\thispagestyle{empty}
\newpage
\selectlanguage{ngerman}
\vspace*{\fill}
\noindent
\textbf{Erkl\"arung:}\\
\\
\noindent
Ich versichere hiermit, dass ich die Arbeit selbstst\"andig verfasst habe, keine anderen als die
angegebenen Quellen und Hilfsmittel benutzt habe, die w\"ortlich oder inhaltlich \"ubernommenen
Stellen als solche kenntlich gemacht habe und die Satzung des Karlsruher Instituts f\"ur Technologie
zur Sicherung guter wissenschaftlicher Praxis beachtet habe.\\
\\
\\
\noindent
Karlsruhe, den 31. August 2015
\begin{flushright}

\myname
\end{flushright}

\vspace*{\fill}
\cleardoublepage

\thispagestyle{empty}
\newpage
\selectlanguage{ngerman}
\vspace*{\fill}
\noindent
\textbf{Kurzzusammenfassung:}\\
\\
\noindent
Diese Bachelorarbeit befasst sich mit der Klassifikation menschlicher Ganzkörperbewegung mittels Hidden Markov Modellen (HMMs). Die menschlichen Bewegungen wurden mittels eines optisches Motion-Capture-Systems mit passiven Markern aufgenommen. Im Laufe der Arbeit werden zunächst Merkmale (\emph{features}) erläutert, die die so aufgenommenen menschlichen Ganzkörperbewegungen beschreiben. Weiterhin wurden neue Merkmale aus den vorhandenen Rohdaten abgeleitet. Ein weiterer Schwerpunkt der Arbeit findet sich in der Diskussion von verschiedenen Ansätzen zur Lösung des Multi-Label-Klassifikationsproblems, also der Zuweisung von mehreren Klassen auf eine Bewegung. Hierbei wurden zwei grundsätzliche Ansätze erarbeitet: Beim ersten Ansatz wird das Multi-Label-Problem zu einem Multi-Class-Problem transformiert, welches sich dann einfach lösen lässt. Der zweite Ansatz beschäftigt sich mit Möglichkeiten, das Multi-Label-Problem ohne vorhergehende Transformation zu lösen, indem die Bewertungen einer unbekannten Bewegung durch die HMMs zu einer Gesamtvorhersage zusammengesetzt werden. Alle Aspekte des Klassifikators wurden anschließend auf einem aus $454$ Bewegungen bestehenden Datensatz evaluiert. Hierbei wurden verschiedene Parameter und Konfigurationen verglichen. Weiterhin fand ein Vergleich zwischen HMMs und Factorial Hidden Markov Modellen (FHMMs) statt. Die zwei zuvor genannten Ansätze wurden in zwei verschiedenen Systemen realisiert und quantitativ miteinander verglichen. Hierbei erkannte das erste System $98,02\%$ und das zweite System $93,39\%$ der menschlichen Bewegungen auf dem Testdatensatz korrekt.
\vspace*{\fill}

\cleardoublepage

\selectlanguage{english}
\begingroup
  \renewcommand*\chapterpagestyle{empty}
  \tableofcontents
\endgroup
\thispagestyle{empty}
\newpage

\setcounter{page}{1}
\chapter{Introduction}
Human whole-body motion plays an important role in many fields, including sports, medicine, entertainment,
computer graphics and robotics. Motion capture technologies are readily available to relatively
easily record vast amounts of data. Today, whole characters in movies
and video games are created using computer-generated imagery (CGI) and the recorded movements of an actor.
An impressive example is the completely computer-generated character \emph{Gollum} from the \emph{Lord
of the Rings} movies. In sports and medicine, motion data can be used for gait analysis of humans.
For example, stroke patients have been recorded using motion capture techniques to analyze their
disease-induced walking disorders. The insights from this analysis can be used to better understand
the symptoms of the patient and potentially allows the development of new rehabilitation therapies.
Gait analysis is also used in professional sports to analyze the performance of athletes and to optimize
their training. In humanoid robotics, human whole-body motion plays an important role
in the construction and control of biped humanoid robots. The use of motion in the field of robotics will be discussed
in-depth in the \emph{Related Works} chapter.

Given the interest in human motion and the availability of the recording equipment, a growing number of motion
data is recorded. A number of human whole-body motion databases exist today to make this data available to artists, physicians and researchers.
Retrieval of information from these databases is often realized through so-called tags which are associated with a motion record.
A motion can potentially have many tags that describe it. For example, a motion of a human playing tennis
might be labeled with the tags \emph{tennis}, \emph{forehand} and \emph{right hand}. These tags can be used
to query the database for motion data of interest. However, annotating a motion with these tags is usually
done by hand which is both an error-prone and slow process. The results are highly subjective since different
annotators will use different tags to label the same data. This may be because different annotators have a different
understanding of a tag or simply because they are not aware of the full set of available tags. Additionally,
labeling every motion by hand becomes infeasible as more data is recorded.

Autonomous motion recognition and classification can be used to automatically label new motions without
human involvement. This approach solves the two main problems of labeling by hand. Firstly, the classification algorithm
produces objective and reproducible results. Secondly, the system can be scaled to handle more and more
data by increasing the available computational resources. The objective of this thesis is to develop
a system that can be used to perform such autonomous classification of human whole-body motion.

This thesis is organized as follows: Chapter~\ref{chapter:related-work} provides a brief overview of the
relevant literature and important authors in the field. Some fundamental concepts are introduced in chapter~\ref{chapter:basics} which forms
the basis for all following chapters. This includes an in-depth discussion of motion capture systems and
ways to represent this recorded data. The \emph{KIT Whole-Body Human Motion Database} is introduced since
the system developed during this thesis will be used to classify motions stored in this database. Additionally,
the foundations of \emph{Hidden Markov Models} (HMMs) and \emph{Factorial Hidden Markov Models} (FHMMs) are discussed
since they play a key role for the devised classifier. Chapter~\ref{chapter:features}
discusses possible features to discriminate motions. The discussion includes different representations
of motion data, the extraction of additional features from the data and important preprocessing steps
before the features can be used in a classifier. Chapter~\ref{chapter:classification} is concerned
with the problem of classification. In the chapter the previously discussed basics of Hidden Markov Models
are extended so that they can be used for motion recognition. The remainder of the chapter is devoted
to the use of multiple HMMs for the classification of human motion. The theoretical concepts
discussed in the previous chapters are then evaluated in chapter~\ref{chapter:evaluation}. Besides a description
of the used tools and dataset, the evaluation includes the selection of features, a comparison between different
HMM configurations as well as the evaluation of different classification approaches. Finally, the best
results from each area are used to perform an end-to-end evaluation of the entire system. Chapter~\ref{chapter:conclusion}
summarizes the work and describes possible improvements for future work.

\cleardoublepage

\chapter{Related Work}
\label{chapter:related-work}
In robotics, a promising idea is to use human motion as an intuitive way to
instruct and program machines. This approach is commonly referred to as \emph{Programming by Demonstration}
(PbD)~\cite{dillmann:2000learning, Billard:2008kb}. In PbD, a human instructor teaches a machine how to complete a
given task by performing the necessary steps themselves. The machine then observes the instructor
and attempts to also complete the task by imitating what it perceives.
Human motion can also be used to gain a better understanding of how different parts of the human body work together to complete a task or goal.
For example, observing how a human reacts to counter balance perturbations can potentially be used to transfer this knowledge to biped humanoid robots~\cite{mandery:2015analyzing, babivc:2014effects}.
Since motion plays a key role in humanoid robotics most of the hereafter mentioned literature has a background in this
area of research.

Different approaches exist for representing motions. Ogata~et~al.~\cite{ogata:2005open} used
\emph{Recurrent Neural Networks} (RNNs) for interactive learning. In their work, RNNs were
used in a cooperative navigation task with a humanoid robot and a human partner. Taylor~et~al.~\cite{taylor:2006modeling, taylor:2009factored}
proposed \emph{Conditional Restricted Boltzmann Machines} (CRBMs) to learn from and then generate human whole-body motion. The proposed
model is capable of generating continuous motion sequences (e.g. walking) and also allows to smoothly transition between them by adding higher-order layers to the model.
\emph{Nonlinear Oscillators} were used by Nakanishi~et~al.~\cite{nakanishi:2004learning} in a framework for learning
biped locomotion. Breazeal~et~al.~\cite{breazeal:2005learning} represented motion as a path through a \emph{directed weighted graph} where
each node represents a pose. The edges of the graph define transitions between poses that are physically possible and safe.
Calinon~et~al.~\cite{calinon:2007learning} used a \emph{Mixture Model of Gaussian and Bernoulli distributions} (GMM/BMM)
to encode motion data. Yamane~et~al.~\cite{Yamane:RSS09} represented continuous motions in \emph{binary trees} which can be used for
motion recognition and generation. Lastly, \emph{Hidden Markov Models} (HMMs) have been a popular choice to represent human whole-body motion.
Since this work is concerned with Hidden Markov Models, the following paragraphs review some works in this domain in greater detail.

Takano~at~al.~\cite{Takano:2006primitive} developed a system for recognizing and generating human motion
for primitive nonverbal communication. Their approach uses a hierarchy of Hidden Markov Models for human
whole-body motion recognition and motion generation. The lower layer represents motion primitives, also referred to as \emph{proto symbols},
whereas the upper layer models the transitions between the motion primitives and therefore represents higher-level
interactions. The lower layer of the system was trained on joint angle data recorded with an optical motion capture system.
Multi-dimensional scaling was used to construct a multi-dimensional space of proto symbols (the \emph{proto symbol space})
on which the Hidden Markov Model in the upper layer was trained. The authors evaluated their approach by
recording a kickboxing match between two humans. One of the human subjects was then replaced with
a humanoid robot. The model trained on the recorded data was used by the robot to generate and perform motions in response to
the actions of its human counterpart.

Kulić~et~al.~\cite{Kulic:2007bf, Kulic:2007clustering, Kulic:2008ib} proposed a system for learning, clustering
and hierarchy formation of human whole-body motion in humanoid robots. The authors used Hidden Markov Models and Factorial Hidden Markov Models to represent
motion as a sequence of \emph{motion primitives}. Additionally, the system described by the authors is capable of on-line learning. This was achieved
by two essential properties of the devised system: sequential training of FHMMs and incremental hierarchical
formation of the motion primitives by clustering. The sequential training algorithm allowed Kulić et al. to initially encode an observed motion
into a simple Hidden Markov Model. As more and more data is observed, additional chains can be added and trained on-line, transforming the
HMM into an FHMM. Secondly, newly observed motions are dynamically organized into an hierarchical tree structure, the \emph{motion symbol tree}.
This can be done efficiently by performing a tree search and placing the new motion into the node that is most similar.
Local clustering is performed to split groups into new subgroups as new knowledge is added. As a result, specialized motions
are placed at the leaves of the tree, whereas more generalized motions can be found near the root. Both properties allow
a humanoid robot to incrementally and efficiently build, organize and access knowledge during operation. The authors evaluated
their work with a database of recorded human whole-body motion data. The data set contained 28 motions for walking, 15 cheering motions,
7 dancing motions, 19 kicking motions, 14 punching motions, 13 sumo leg raise motions, 13 squatting motions, 13 throwing motions, and 15 bowing
motions. Each motion was represented in a humanoid model with 20 degrees of freedom. The results indicated that Factorial
Hidden Markov Models outperform single-chain Hidden Markov Models in their discriminative and generative
properties.

The work by Kulić et al. was extended in \cite{takano:2010organization} and \cite{Kulic:2011incremental}. The authors used the motion symbol tree to perform efficient classification. This was achieved
by traversing the tree from the root and only recursively considering the subtree with the highest likelihood. As soon as a leave node is reached, the classification
is complete. The approach greatly decreased the computational cost since fewer comparisons are required in order to classify an unknown motion. The authors
further introduced the concept of a \emph{motion symbol graph}. This directed graph allowed the authors to model likely transitions between motion symbols when observing continuous
motion. The motion symbol graph was used to predict motion patterns
and to generate motions in humanoid robots that consists of sequences of motion primitives.

In \cite{Takano:2015ca, takano:2015construction, takano:2015statistical}, Takano et al. proposed a system for mapping between motion symbols and word labels. Motions were encoded into HMMs
and the distances between all models was calculated. Like in the earlier work of Takano, the distance measures was used to construct a multi-dimensional space, the \emph{motion symbol space}. Multiple word labels
associated with the motion primitive were encoded into a binary vector, which can be seen as a point in \emph{word label space}. Finally, a linear mapping between the motion symbol space
and the word label space was learned using Canonical Correlation Analysis (CCA). CCA attempts to find a mapping in such a way that the correlation of the positions of motion symbols and
word labels is maximized. An advantage of this model is that it can be used to map from motion symbol to word label and
vice versa. This means that the system is both capable of motion classification given an unknown motion and motion retrieval given a query of word labels. In the latter case, since motion symbol
space and word label space are metric spaces, it is also possible to calculate the distance between a word label query and a motion symbol, making it easy to quantify the similarity.

An interesting extension to Hidden Markov Models was proposed in \cite{wilson:1999phmm}. Wilson et al. used \emph{Parametric Hidden Markov Models} (PHMMs)
to recognize parameterized gestures. An example of such a gesture is the movement of the hands that accompanies the speech ``I saw a robot \emph{this} big!''.
Here, \emph{this} is a parameter of the gesture, namely the scalar size of the observed robot. The authors showed that traditional HMM-based recognition
cannot adequately model this spatial variance. Furthermore, HMMs do not allow to estimate the parameter (e.g. the size of the robot) from an unknown gesture.
The PHMM devised by the authors can solve both problems efficiently. It works by weighting a parameter vector and adding it to the mean
of the emission distribution of each hidden state. A modified version of the Baum-Welch algorithm was used to estimate the weights of the parameter vector. Recognition with PHMMs is complicated
by the fact that the parameter vector is unknown. This was solved by estimating the parameter vector (using an EM algorithm) for the observed sequence and each PHMM. The PHMM with the highest likelihood
was then selected. Furthermore, the authors extended PHMMs to a non-linear mapping from parameter vector to the means of the emission distributions. In this case, gradient ascent techniques were used to estimate the necessary parameters.

Herzog~et~al.~\cite{herzog:2008motion} used a variation of PHMMs to recognize and imitate motions in humanoid robots. Their approach differed from the model proposed by Wilson et al. The basic idea
proposed by the authors is to use linear interpolation of HMMs that were trained on known parameters to generate a new HMM for new parameters. In their work the authors further discussed
how a humanoid robot can generate motions from such a model. The authors evaluated their approach on pointing and reaching motions and were able to show that, in those cases, PHMMs outperform traditional HMMs in classification.

Krüger~et~al.~\cite{kruger:2010learning} used PHMMs for action recognition. Their work builds on the idea that
an action can be represented by a sequence of \emph{action primitives}. The authors proposed a system that
used unsupervised segmentation to discover the action primitives. PHMMs were then used to encode and recognize them,
as well as synthesis motions with a desired effect (e.g. grabbing an object).

\chapter{Basics}
\label{chapter:basics}
Classification of human whole-body motion first and foremost requires motion data. This chapter therefore
starts with a brief discussion of motion capture (section~\ref{motion-capture}) and the Master Motor Map as
a framework for representing motion (section~\ref{mmm}). Section~\ref{database} gives an overview
of the KIT Whole-Body Human Motion Database, which plays an important role in this work since it
stores all motion data and also provides structures for labeling motions. Hidden Markov Models are
introduced in section~\ref{hmm}, which are used to learn and recognize motions in this work. Lastly,
an extension of HMMs, Factorial Hidden Markov Models, are discussed (section~\ref{fhmm}).

\section{Motion Capture}
\label{motion-capture}
For recording motion data, the \emph{VICON MX} motion capture system can be used. The system uses
passive optical markers that can be attached to both humans and objects. Cameras that are
positioned at multiple locations around the scene record the position of the markers within line of
sight. To do so, each camera features a ring of LEDs that surrounds its lens. The LEDs emit light in
the infrared spectrum, which is then reflected by the markers. Each camera records this reflected
light and (depending on the mode of operation) reports the 2D coordinates of the markers. The final 3D
coordinates for each marker are calculated by triangulation using the data from each camera~\cite{vicon}.

\begin{figure}[h]
    \centering
    \includegraphics[width=0.6\textwidth]{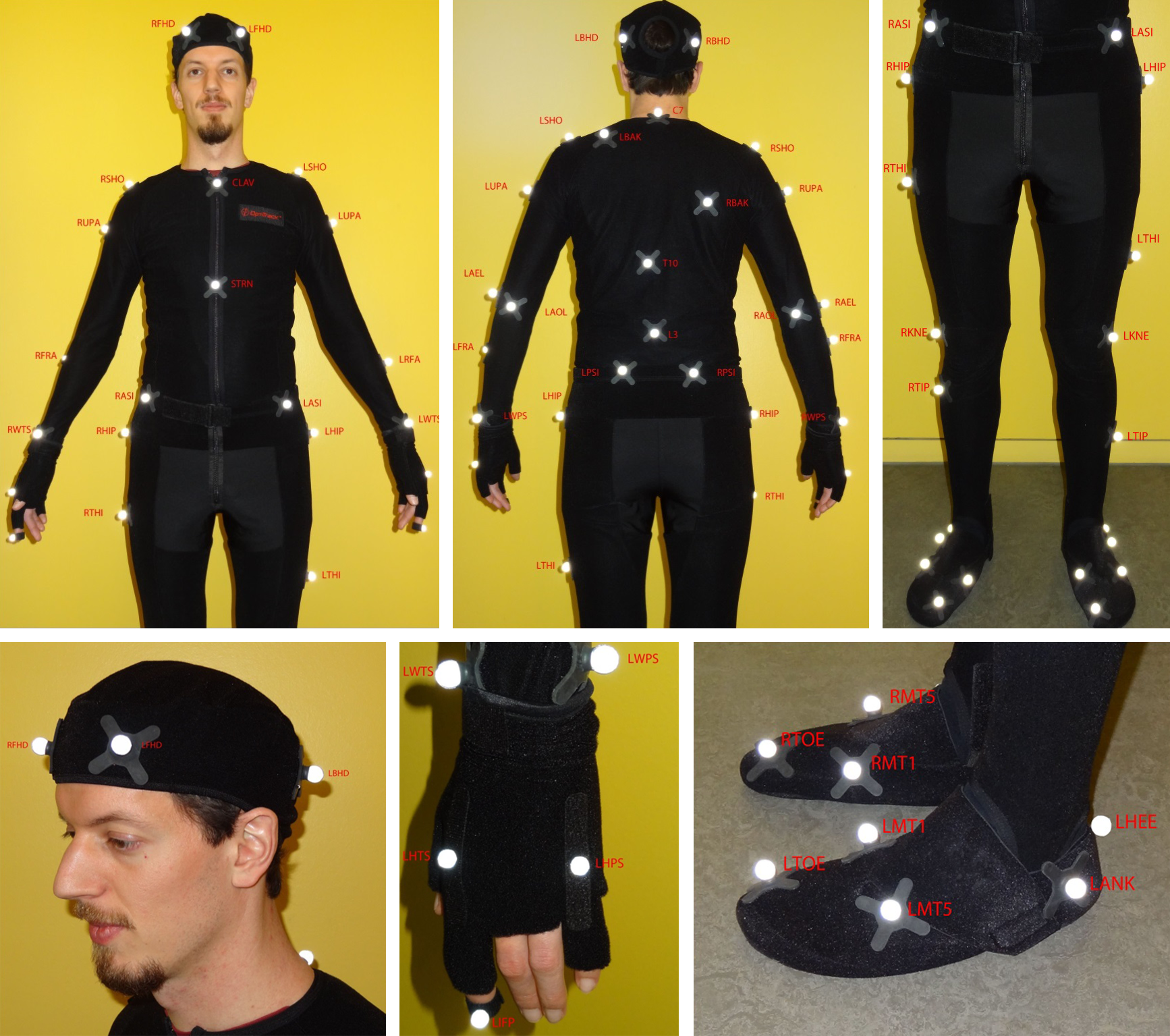}
    \caption{Marker placement on the human body~\cite{Mandery:2015db}.}
    \label{fig:markers}
\end{figure}

The motions are recorded by eight stationary and two portable
\emph{VICON T10} cameras. Each camera records with a sampling rate of \SI{100}{\Hz}. A total of 56 markers
are placed onto the human as depicted in figure~\ref{fig:markers}.

All recorded motion data is stored in the \emph{C3D format}. The C3D format is a binary file format under
public domain and is considered an industry standard. Besides storing marker coordinates in 3D space,
it also allows to store information about the human (e.g. body size and weight), the experiment setup
(e.g. marker positions) as well as additional data (e.g. data from additional sensors like force sensors)~\cite{c3d}.

\section{Master Motor Map}
\label{mmm}
\begin{figure}[h]
    \centering
    \includegraphics[width=0.8\textwidth]{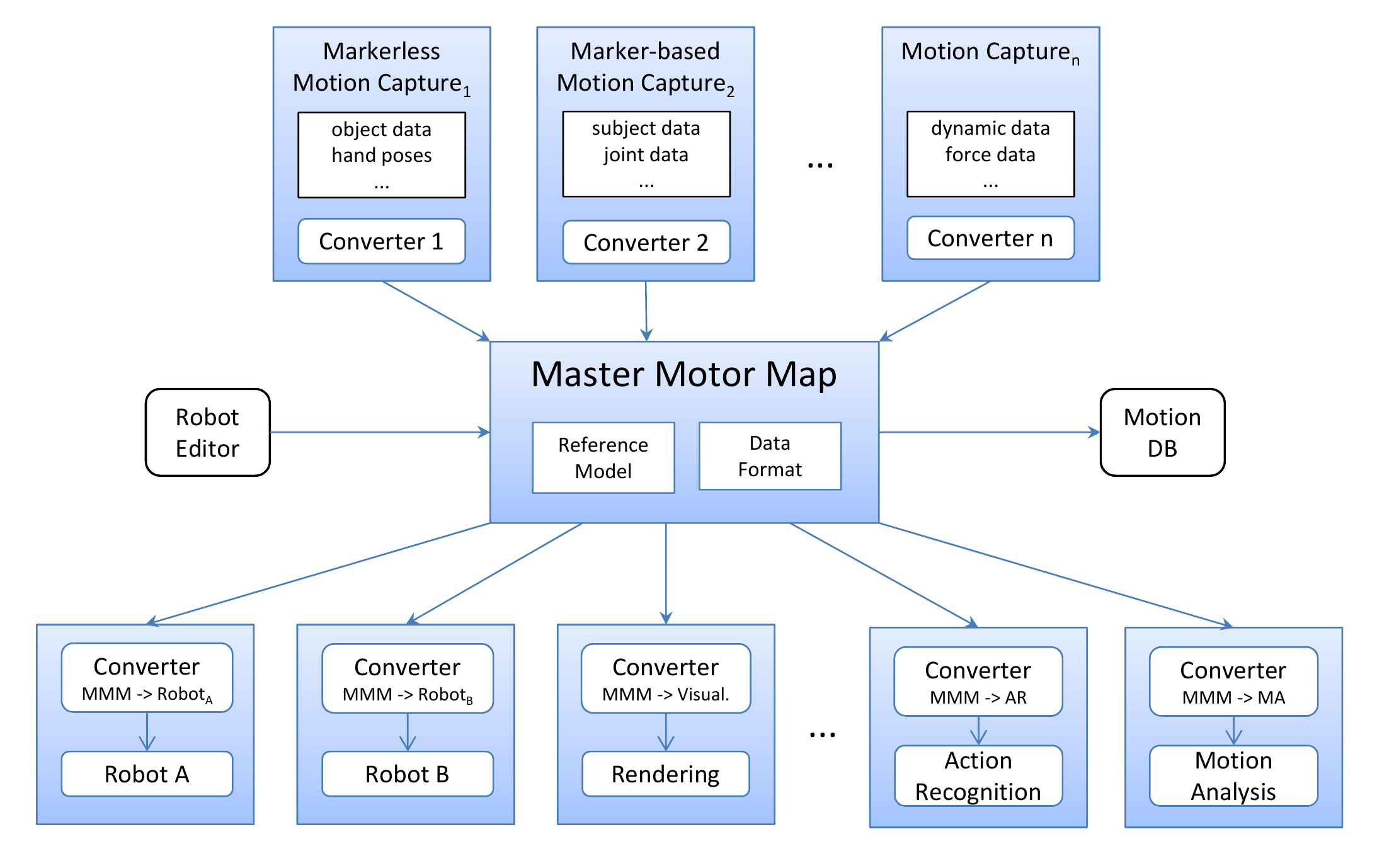}
    \caption{Architecture of the Master Motor Map framework~\cite{Terlemez:2014master}.}
    \label{fig:mmm_architecture}
\end{figure}

\emph{Master Motor Map} (MMM)~\cite{Terlemez:2014master, Azad:2007unifiedrepr} is a framework for representation, mapping and reproduction of human motions on humanoid robots. The fundamental goal of MMM is to map and unify different motions performed by different
humans and recorded with different motion capture systems to the MMM reference model. Motions represented
under this reference model can then be converted to different outputs, e.g. to map a human motion onto
a humanoid robot like \emph{ARMAR-III}~\cite{Asfour:2006armar}. The architecture of the MMM framework is depicted in
figure~\ref{fig:mmm_architecture}. The framework includes different command-line and graphical user interface (GUI) tools
and is open source\footnote{\url{http://h2t.anthropomatik.kit.edu/752.php}}.

At the core of the framework is the MMM reference model. It consists of a model of the human body with a
normalized height and weight, as well as kinematic and dynamic properties.
These properties are based on the research conducted by Winter~et~al.~\cite{Winter:1979biomechanics, Winter:2009biomechanics}
and Buchholz~et~al.~\cite{Buchholz:1992anthropometric}.
The kinematics of the MMM model consist of 104 degrees of freedom (DoF): 6 DoF cover the model pose,
23 DoF are assigned to each hand, and the remaining 52 DoF are distributed on arms, legs, head, eyes
and body. The reference coordinate system in every joint is chosen in such a way that the $x$-axis points to the right of the model,
the $y$-axis to the front and the $z$-axis upwards. If a joint has multiple DoF it is
split into multiple joints with a single DoF each. The MMM reference model also specifies upper and
lower limits for each joint. It is important to note that not all joints in the model must be used
to represent motion. For example, the movement of individual fingers might not be of interest
when recording a whole-body walking motion. In this case, the unspecified joints will simply remain
in their initial positions~\cite{Terlemez:2014master}.

The command-line tool \emph{MMMConverter} can be used to convert data recorded with a motion capture system to the
MMM reference model, i.e. reconstruct joint angles of the MMM model from motion data. This is accomplished by placing virtual markers onto the reference model and finding
a mapping from the position of the physical markers (as recorded by the motion capture system) to
virtual markers. The optimization problem can be solved efficiently by minimizing the distance between the position
of physical and virtual markers for each frame. Details on this mapping procedure are given in~\cite{Terlemez:2014master}.
The converted motion is then stored in a XML-based file format. Such a XML file can contain multiple motions, which
is useful for scenes where a human interacts with objects or scenes with multiple humans in them. Each motion
is referenced by a unique name and consists of two parts: a preamble and the actual motion data.
The preamble specifies a model (e.g. the MMM reference model or a model of an object)
and can contain additional information about the human or object (e.g. body size and weight). The
motion data is encoded in a list of frames. Each frame has a relative time step in seconds (the first
frame starts at time step $t = 0$) and contains values for all properties that are specified by the model.
For example, each frame of human motion under the MMM reference model contains the root position ($x$, $y$, $z$ coordinates),
the root rotation (roll, pitch, yaw angles) and a list of joint angles. Additionally, the velocity and
acceleration information for each of the above properties as well as dynamic data (e.g. center of mass, angular momentum) can be stored for each frame~\cite{mmm:dataformat}.

\begin{figure}[h]
    \centering
    \includegraphics[width=0.8\textwidth]{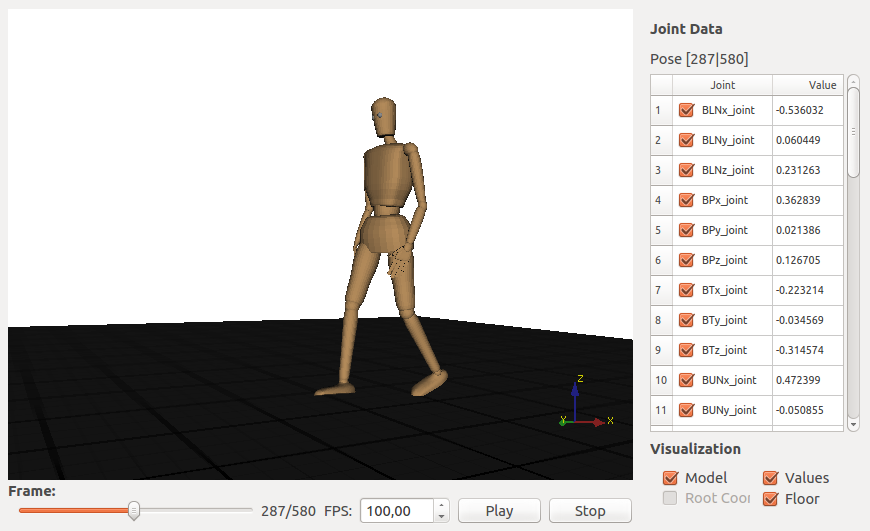}
    \caption{Visualization of a walking motion in \emph{MMMViewer}.}
    \label{fig:mmm_viewer}
\end{figure}

The GUI tool \emph{MMMViewer} can be used to visualize motions.
The whole motion can be played back or each frame can be inspected individually. The camera can be moved freely
to view the motion from different angles. Additionally, the joint angles of the currently visible frame
are displayed. Figure~\ref{fig:mmm_viewer} shows the tool during a visualization.

\section{KIT Whole-Body Human Motion Database}
\label{database}
The \emph{KIT Whole-Body Human Motion Database}\footnote{\url{https://motion-database.humanoids.kit.edu}}
contains motion data of both humans and objects that have been recorded using a marker-based approach
as described in section~\ref{motion-capture}. Each entry in the database has a unique ID, belongs to a project,
and references the subjects and objects that participated in the recording. When recording motions,
multiple trials are usually performed. For each trial, the raw motion data is stored in the C3D format (see
section~\ref{motion-capture}) and uploaded. Recorded data of additional sensors (e.g. force measurements for push recovery)
as well as video footage can be uploaded as well. The database system automatically converts the
C3D files to a subject-independent representation under the MMM model (see section~\ref{mmm})
and, optionally, estimates dynamic properties like the center of mass. Log files grant insight into
the conversion process. Furthermore, the database is capable of storing data related to subjects and
objects. Size, weight, gender and other anthropometric measurements can be stored in the record for
each subject. For objects, a 3D model of the object alongside a custom description can be
saved~\cite{Mandery:2015db}.

\begin{figure}[h]
    \centering
    \includegraphics[width=0.9\textwidth]{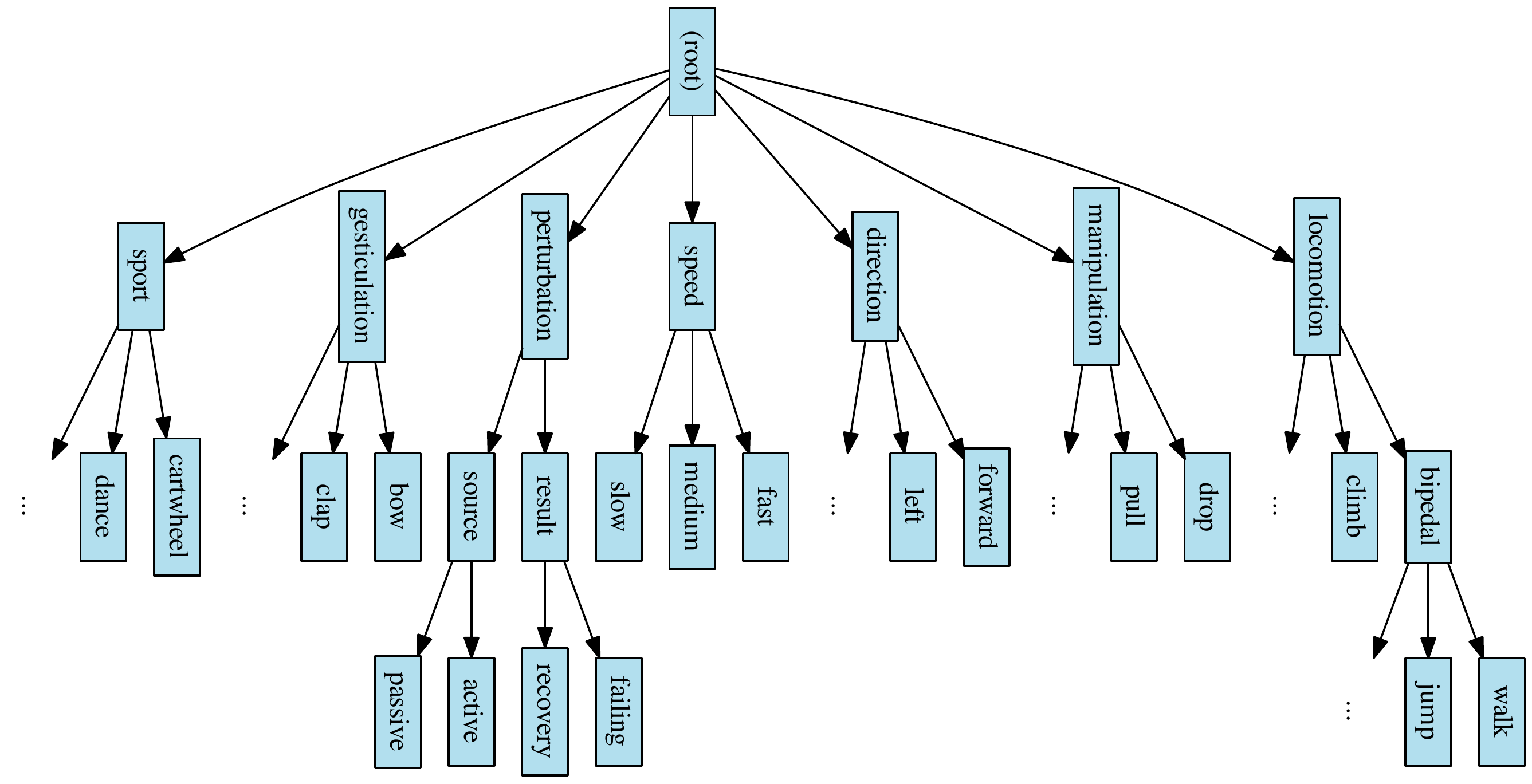}
    \caption{The Motion Description Tree (excerpt)~\cite{Mandery:2015db}.}
    \label{fig:motion-description-tree}
\end{figure}

Each motion is classified within the \emph{Motion Description Tree}. The tree consists of a hierarchical declaration
of tags describing motion types (e.g. \emph{walk}, \emph{kick}, \emph{run}) and additional tags for other properties like the
direction of a movement (e.g. \emph{left}, \emph{right}, \emph{forward}, \emph{backward}). The tree is organized in such a way
that the parent of a node has a broader semantic meaning than its children. For example, the tags \emph{clap}
and \emph{bow} are both child nodes of the tag \emph{gesticulation}. An excerpt of the Motion
Description Tree is depicted in figure~\ref{fig:motion-description-tree}. An important property of
this classification approach is that a motion can be associated with an arbitrary number of nodes of
the Motion Description Tree. For example, a motion of a human that trips while walking to the left
with high speed can be categorized using the following tags:
\begin{inparaenum}[(1)]
  \item \emph{locomotion}~$\rightarrow$~\emph{bipedal}~$\rightarrow$~\emph{walk},
	\item \emph{speed}~$\rightarrow$~\emph{fast},
	\item \emph{direction}~$\rightarrow$~\emph{left},
	\item \emph{perturbation}~$\rightarrow$~\emph{result}~$\rightarrow$~\emph{failing}, and
	\item \emph{perturbation}~$\rightarrow$~\emph{source}~$\rightarrow$~\emph{passive}.
\end{inparaenum}
The whole tree is managed by the KIT Whole-Body Human Motion Database and can be extended if necessary~\cite{Mandery:2015db}.

The database can be accessed through a web interface or an application programming interface (API).
The web interface and the API are available publicly. For each motion the raw files as well as the processed files can
be downloaded. For convenience, bulk download options are available. The web interface is also used to
modify existing or upload new motions. These operations are restricted to registered accounts. The API
allows direct access to the database. This allows the integration of the database into existing
tools. The API is build on top of the \emph{Internet Communications Engine} (Ice)~\cite{ice}.
Ice is a \emph{remote procedure call} (RPC) framework and allows for easy integration with a wide variety
of platforms and programming languages~\cite{Mandery:2015db}.

At the time of this writing, the database contains $4\,457$ motions performed by $49$ different subjects. All motions
in total have a length of approximately $9$ hours and $20$ minutes, with the average length of a recording being
approximately $7.56$ seconds.

\section{Hidden Markov Models}
\label{hmm}
A \emph{Hidden Markov Model} (HMM)~\cite{Elliott:1995HMM, Rabiner:1989hs} is a statistical model popular for learning
sequential data. This is due to the fact that HMMs have the ability to have some degree of invariance
to local warping (compression and stretching) of the time axis~\cite{Bishop:2006pattern}. The methods
discussed in this section are applicable to all forms of sequential data. However, since this work deals
with temporal sequences, this section and all following chapters use notation and phrases that imply temporal
sequences. Concretely, a \emph{temporal sequence} of length $T$ is denoted by $\vec{o}_1, \vec{o}_2, \ldots, \vec{o}_T$,
where each $\vec{o}_t$ is a multi-dimensional \emph{observation}. The following discussion of HMMs and
the underlaying concepts are based on Bishop~et~al.~\cite{Bishop:2006pattern}.

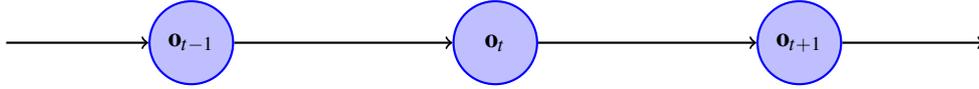
\begin{figure}[h]
  \centering
  \begin{tikzpicture}[->,scale=1.,auto=left,thick,every node/.style={circle,thick,minimum size=1.1cm}]
    \tikzstyle{observation}=[draw=blue!100,fill=blue!25]

    \node              (x1) at ( 0,0) {};
    \node[observation] (x2) at ( 3,0) {$\vec{o}_{t-1}$};
    \node[observation] (x3) at ( 7,0) {$\vec{o}_t$};
    \node[observation] (x4) at (11,0) {$\vec{o}_{t+1}$};
    \node              (x5) at (14,0) {};

    \foreach \from/\to in {x1/x2, x2/x3, x3/x4, x4/x5}
      \draw (\from) -- (\to);
  \end{tikzpicture}
  \caption{Graphical representation of a first-order Markov chain. Each observation $\vec{o}_t$ is only dependent on the previous
    observation~\cite{Bishop:2006pattern}.}
  \label{fig:markov-chain}
\end{figure}

To understand Hidden Markov Models, it is helpful to first consider a simpler \emph{Markov model}: the
\emph{Markov chain}. In a Markov chain (of first order), given a sequence $\vec{o}_1, \ldots, \vec{o}_T$, the conditional
probability of an observation $\vec{o}_t$ (that is the observation at time $t$) is assumed to be independent
of all past (and, of course, future) observations except for observation $\vec{o}_{t-1}$. The conditional
distribution in a such a model is given by
\begin{equation} 
	p(\vec{o}_t \mid \vec{o}_1, \ldots, \vec{o}_{t-1}) = p(\vec{o}_t \mid \vec{o}_{t-1}).
\end{equation}
Consequently, the joint distribution is given by
\begin{equation} 
	p(\vec{o}_1, \ldots, \vec{o}_T) = p(\vec{o}_1) \prod\limits_{t=2}^T{p(\vec{o}_t \mid \vec{o}_{t-1})}.
\end{equation}

A graphical representation of a first-order Markov chain is depicted in figure~\ref{fig:markov-chain}.
The assumption that an observation is only dependent on its previous observation is rather strong.
This can be easily seen by considering an example: If one attempts to predict the weather for the next
hour by only considering the current weather situation instead of using the data of the last
24 hours, the prediction would be severely limited. The assumption can be relaxed by generalizing
the Markov chain to be of $M$-th order. Here, each observation in a sequence is dependent on the
past $M$ observations. However, in such a model the number of parameters grows exponentially with M,
so that this approach becomes impractical for large values of $M$.

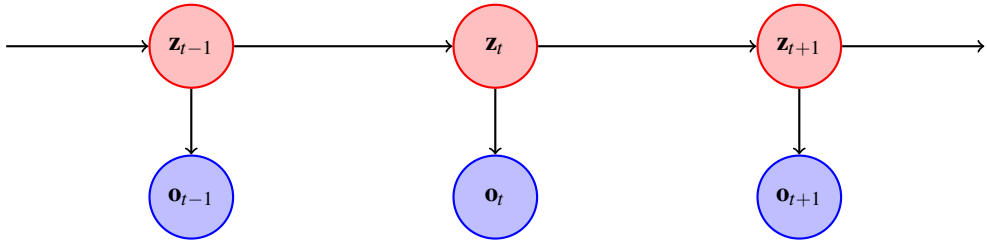
\begin{figure}[h]
  \centering
  \begin{tikzpicture}[->,scale=1.,auto=left,thick,every node/.style={circle,thick,minimum size=1.1cm}]
    \tikzstyle{hidden}=[draw=red!100,fill=red!25]
    \tikzstyle{observation}=[draw=blue!100,fill=blue!25]

    \node         (z1) at ( 0,2) {};
    \node[hidden] (z2) at ( 3,2) {$\vec{z}_{t-1}$};
    \node[hidden] (z3) at ( 7,2) {$\vec{z}_t$};
    \node[hidden] (z4) at (11,2) {$\vec{z}_{t+1}$};
    \node         (z5) at (14,2) {};

    \node              (x1) at ( 0,0) {};
    \node[observation] (x2) at ( 3,0) {$\vec{o}_{t-1}$};
    \node[observation] (x3) at ( 7,0) {$\vec{o}_t$};
    \node[observation] (x4) at (11,0) {$\vec{o}_{t+1}$};
    \node              (x5) at (14,0) {};

    \foreach \from/\to in {z1/z2, z2/z3, z3/z4, z4/z5}
      \draw (\from) -- (\to);
    \foreach \from/\to in {z2/x2, z3/x3, z4/x4}
      \draw (\from) -- (\to);    
  \end{tikzpicture}
  \caption{Graphical representation of a Hidden Markov Model. Each observation $\vec{o}_t$ (in {\color{blue}blue}) is conditioned
  on the state of its respective hidden variable $\vec{z}_t$ (in {\color{red}red}). The hidden variables form a first-order
  Markov chain~\cite{Bishop:2006pattern}.}
  \label{fig:markov-hidden-chain}
\end{figure}

To solve this problem, \emph{hidden} (sometimes also referred to as \emph{latent}) variables are
introduced. Concretely, each observation variable $\vec{o}_t$ is conditioned on the state of its hidden variable
$\vec{z}_t$. The hidden variables form a first-order Markov chain. Such a model is known as a \emph{state space model},
which is visualized in figure~\ref{fig:markov-hidden-chain}. The joint distribution for this model is given by
\begin{equation} 
	p(\vec{o}_1, \ldots, \vec{o}_T, \vec{z}_1, \ldots, \vec{z}_T) = p(\vec{z}_1) \prod\limits_{t=2}^T{p(\vec{z}_t \mid \vec{z}_{t-1})}
	\prod\limits_{t=1}^T{p(\vec{o}_t \mid \vec{z}_t)}.
\end{equation}
An important property of this model is that any pair of observed variables $\vec{o}_i$ and $\vec{o}_j$
are connected via the hidden variables. It can be shown that the predictive distribution
$p(\vec{o}_{t+1} \mid \vec{o}_1,\ldots,\vec{o}_t)$ for observation $\vec{o}_{t+1}$ is dependent on all past
observations $\vec{o}_1,\ldots,\vec{o}_t$. This model is therefore not
constrained by the strong independence assumption of a Markov chain. If the hidden variables in a
space state model as described above are discrete, the Hidden Markov Model is obtained.

Formally, a Hidden Markov Model is defined by the set of parameters that govern the model~\cite{Bishop:2006pattern, Rabiner:1989hs}:
\begin{itemize}
    \item $K$ is the number of the individual hidden states. The appropriate number of states depends on the problem
    at hand. The current state at time $t$ is denoted by $q_t$ and the set of possible states is $S = \{s_1, \ldots, s_K\}$.
    Notice the relationship of $K$ and the hidden variables: Each of the $K$ states describes a possible value
    for the hidden variables $\vec{z}_t$. One way to represent this is through the $1$-of-$K$ coding scheme, hence $\vec{z}_t \in \{0,1\}^K$ and
    $\norm{\vec{z}_t}_1 = 1$ for each $t$. For example, the notation $q_t = s_2$ is equivalent to $\vec{z}_t = (0, 1, 0, \ldots, 0)$.
	
	\item $\vec{A} = (a_{ij}) \in [0,1]^{K \times K}$ is a matrix that consists of \emph{transition probabilities}. Concretely,
	each entry defines the probability to transition to state $s_j$ given the current state $s_i$:
	\begin{equation} 
		a_{ij} = p(q_{t+1}=s_j \mid q_t=s_i).
	\end{equation}
	Since $\vec{A}$ is a probability distribution, it must hold that $\forall j: \sum\nolimits_k{a_{kj}}=1$. By setting
	$a_{ij}=0$, it is possible to ``disable'' that specific transition from $s_i$ to $s_j$.

	\item $\vecsym{\pi}=(\pi_{1}, \ldots, \pi_{K}) \in [0,1]^K$ is the \emph{initial state distribution} where
	\begin{equation} 
		\pi_{i} = p(q_1=s_i).
	\end{equation}
	It must hold that $\norm{\vecsym{\pi}}_1 = 1$.

	\item $\phi$ describes the parameters of the conditional distributions of the observed variables:
	\begin{equation} 
		p(\vec{o}_t \mid \vec{z}_t, \phi).
	\end{equation}
	These probabilities are known as \emph{emission probabilities} and can be given by different
	distributions. For example, if the observed values are discrete, a conditional probability table
	can be used. For observations with continuous values, a Gaussian distribution is a often a good choice.
	Other distributions are possible and picking an appropriate distribution depends on the observations.
\end{itemize}
Since $K$ is already encoded by the shape of $\vec{A}$, an HMM is fully described by the following set of parameters:
$\theta = \{\vec{A}, \vecsym{\pi}, \phi\}$. In this work, an HMM with parameters $\theta$ is denoted by $\lambda_\theta$.

Three fundamental problems can be identified when working with HMMs:
\begin{inparaenum}[(1)]
    \item The \emph{evaluation problem},
	\item the \emph{decoding problem}, and
	\item the \emph{optimization problem}.
\end{inparaenum}
The problems and their description are all based on the work of Rabiner~et~al.~\cite{Rabiner:1989hs}.
The first problem, the evaluation problem, is concerned with calculating the probability of a given
sequence under a given model. Formally, given a sequence $\vec{O} = (\vec{o}_1,\ldots,\vec{o}_T)$, how can
$p(\vec{O} \mid \lambda_\theta)$ be calculated efficiently. This can also be viewed as scoring how well
a model matches the given observations. The \emph{forward-backward algorithm}~\cite{Baum:1967bf1, Baum:1968bf2}
can be used to solve this problem (strictly speaking, only the forward pass is necessary to solve this first problem).
The second problem, the decoding problem, is concerned with finding the state sequence of the hidden
variables. Formally, given a sequence $\vec{O} = (\vec{o}_1,\ldots,\vec{o}_T)$ and a model $\lambda_\theta$, find a sequence
$\vec{Z} = (\vec{z}_1, \ldots, \vec{z}_T)$ of hidden states that is optimal. Different criteria of an optimal
state sequence exist, e.g. choosing the states that are individually most likely. The most popular criterion
is to find the single best state sequence. This is equivalent to maximizing $p(\vec{Z} \mid \vec{O}, \lambda_\theta)$, which is solved
efficiently by the \emph{Viterbi algorithm}~\cite{Viterbi:1967decoding}.
The third problem, the optimization problem, is concerned with adjusting the parameters of the model.
Formally, given a sequence $\vec{O} = (\vec{o}_1,\ldots,\vec{o}_T)$, find parameters $\theta$ such that $p(\vec{O} \mid \lambda_\theta)$
is maximized. Solving this problem corresponds with learning the parameters, that is ``training'' the model.
The \emph{Baum-Welch algorithm}~\cite{Baum:1970maximization} solves this problem efficiently. However, since
Baum-Welch is a specific case of the \emph{expectation maximization algorithm} (EM algorithm), it does
not necessarily find a global maximum.

On a final note, another important property of HMMs is that they are \emph{generative} models. This means that a model
that has been trained on some data can be used to generate new samples. This is especially interesting
for human motions and humanoid robots. Here, a motion can be learned by observation and later be reproduced in a robot
by sampling from the model~\cite{Takano:2006primitive}.

\section{Factorial Hidden Markov Models}
\label{fhmm}
A sever limitation of HMMs is that they cannot represent a lot of information about the history of a time
sequence. \emph{Factorial Hidden Markov Models} (FHMMs) are a generalization of HMMs and offer a way to overcome this limitation.
For example, representing $30$ bit of information about the history requires $2^{30}$ hidden states in a
standard HMM whereas an FHMM can represent the same information with only $30$ binary state variables~\cite{Ghahramani:1997id}.
The discussion in this section is based on the work of Ghahramani~et~al.~\cite{Ghahramani:1997id}.

\begin{figure}[h]
  \centering
  \begin{tikzpicture}[->,scale=1.,auto=left,thick,every node/.style={circle,thick,minimum size=1.1cm}]
    \tikzstyle{hidden1}=[draw=red!100,fill=red!25]
    \tikzstyle{hidden2}=[draw=green!100,fill=green!25]
    \tikzstyle{hidden3}=[draw=purple!100,fill=purple!25]      
    \tikzstyle{observation}=[draw=blue!100,fill=blue!25]

    \node          (z11) at ( 0,6) {};
    \node[hidden1] (z12) at ( 3,6) {$\vec{z}^{(1)}_{t-1}$};
    \node[hidden1] (z13) at ( 7,6) {$\vec{z}^{(1)}_t$};
    \node[hidden1] (z14) at (11,6) {$\vec{z}^{(1)}_{t+1}$};
    \node          (z15) at (14,6) {};

    \node          (z21) at ( 0,4) {};
    \node[hidden2] (z22) at ( 3,4) {$\vec{z}^{(2)}_{t-1}$};
    \node[hidden2] (z23) at ( 7,4) {$\vec{z}^{(2)}_t$};
    \node[hidden2] (z24) at (11,4) {$\vec{z}^{(2)}_{t+1}$};
    \node          (z25) at (14,4) {};

    \node          (z31) at ( 0,2) {};
    \node[hidden3] (z32) at ( 3,2) {$\vec{z}^{(3)}_{t-1}$};
    \node[hidden3] (z33) at ( 7,2) {$\vec{z}^{(3)}_t$};
    \node[hidden3] (z34) at (11,2) {$\vec{z}^{(3)}_{t+1}$};
    \node          (z35) at (14,2) {};

    \node              (x1) at ( 0,0) {};
    \node[observation] (x2) at ( 3,0) {$\vec{o}_{t-1}$};
    \node[observation] (x3) at ( 7,0) {$\vec{o}_t$};
    \node[observation] (x4) at (11,0) {$\vec{o}_{t+1}$};
    \node              (x5) at (14,0) {};

    \foreach \from/\to in {z11/z12, z12/z13, z13/z14, z14/z15}
      \draw (\from) -- (\to);
    \foreach \from/\to in {z21/z22, z22/z23, z23/z24, z24/z25}
      \draw (\from) -- (\to);
    \foreach \from/\to in {z31/z32, z32/z33, z33/z34, z34/z35}
      \draw (\from) -- (\to);
    \foreach \from/\to in {z12/x2, z13/x3, z14/x4}
      \path (\from) edge [bend left=60] (\to);
    \foreach \from/\to in {z22/x2, z23/x3, z24/x4}
      \path (\from) edge [bend left=40] (\to);
    \foreach \from/\to in {z32/x2, z33/x3, z34/x4}
      \draw (\from) -- (\to);    
  \end{tikzpicture}
  \caption{Graphical representation of a Factorial Hidden Markov Model. Each observation $\vec{o}_t$ (in {\color{blue}blue}) is conditioned
  on the state of all its respective hidden variables $\vec{z}^{(1)}_t$ (in {\color{red}red}), $\vec{z}^{(2)}_t$ (in {\color{green}green}) and
  $\vec{z}^{(3)}_t$ (in {\color{purple}purple}). The hidden variables $\vec{z}^{(m)}_t$ form a first-order Markov chain each.~\cite{Ghahramani:1997id}.}
  \label{fig:fhmm}
\end{figure}
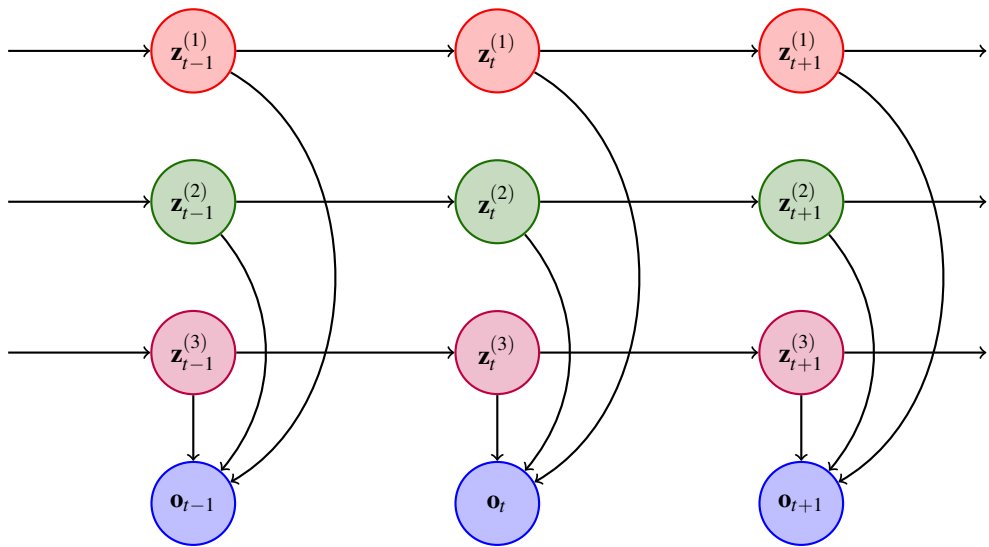

In an FHMM, the current state is generalized by letting the state be represented by a collection of $M$
state variables:
\begin{equation} 
  \vec{z}_t = (\vec{z}^{(1)}_t,\ldots,\vec{z}^{(M)}),
\end{equation}
where each state variable $\vec{z}^{(m)}_t$ can take on $K$ different values. Each state
variable is constrained in such a way that it evolves according to its own dynamics and is therefore
uncoupled from the other state variables:
\begin{equation} 
  p(\vec{z}_t \mid \vec{z}_{t-1}) = \prod\limits_{m=1}^M{p(\vec{z}^{(m)}_t \mid \vec{z}^{(m)}_{t-1}}).
\end{equation}
This can be seen as $M$ independent first-order Markov chains that all contribute to the observation. The distribution of the observed variable
$\vec{o}_t$ is conditional on the states of all hidden variables $\vec{z}^{(1)}_t, \ldots, \vec{z}^{(M)}_t$
for each time step $t$. A simple way to represent this dependency for continuous observations is a
multivariate Gaussian. Concretely, given that each $\vec{z}^{(m)}_t$ uses the $1$-of-$K$ coding scheme as described
in section \ref{hmm} and that each observation $\vec{o}_t \in \mathbb{R}^D$, the conditional distribution
is given by
\begin{equation} 
  p(\vec{o}_t \mid \vec{z}_t) = \abs{\vecsym{\Sigma}}^{-\frac{1}{2}} (2 \pi)^{-\frac{D}{2}} \exp\left(-\frac{1}{2}
  (\vec{o}_t-\vecsym{\mu}_t)^T \vecsym{\Sigma}^{-1} (\vec{o}_t-\vec{\mu}_t)\right),
  \label{eq:fhmm-gaussian}
\end{equation}
where
\begin{equation} 
  \vecsym{\mu}_t = \sum\limits_{m=1}^M{\vec{W}^{(m)}\vec{z}^{(m)}_t}.
  \label{eq:fhmm-mean}
\end{equation}
Here, each $\vec{W}^{(m)}$ is a $D \times K$ matrix that weights the contributions to the mean for each
$\vec{z}^{(m)}_t$, $\vecsym{\Sigma}$ is the $D \times D$ covariance matrix and $\abs{\cdot}$ denotes the matrix determinant.
Also note that the scalar $\pi$ in equation \ref{eq:fhmm-gaussian} is \emph{not} to be confused with the initial probability vector
$\vecsym{\pi}$ from
section \ref{hmm}. In words,
at each time step, the state of all chains are weighted, summed and output through an expectation function
(here equation \ref{eq:fhmm-gaussian}) to produce the observation~\cite{Kulic:2008ib}.

Like an HMM, a Factorial Hidden Markov Model is defined by the set of parameters that govern the model:
$\theta = \{\vec{A}^{(1)}, \ldots, \vec{A}^{(M)}, \vecsym{\pi}^{(1)}, \ldots, \vecsym{\pi}^{(M)}, \phi\}$. This is a simple extension of the parameters of a standard normal HMM: The transition probabilities $\vec{A}^{(1)}, \ldots, \vec{A}^{(M)} \in \mathbb{R}^{K \times K}$ and initial probabilities
$\vecsym{\pi}^{(1)}, \ldots, \vecsym{\pi}^{(M)} \in \mathbb{R}^K$ must be given for each of the $M$ Markov chains. $\phi$ still defines
the necessary parameters for the emission probability distribution, which is $\phi = \{\vec{W}^{(1)}, \ldots, \vec{W}^{(M)}, \vecsym{\Sigma}\}$ for
a Gaussian FHMM as described above.

A problem with FHMMs is learning their parameters. This is because although at each time step the hidden variables are marginally
independent, they become conditionally dependent given the observation sequence. This can be easily seen by considering
equation \ref{eq:fhmm-gaussian} and \ref{eq:fhmm-mean} that makes the mean and therefore the entire Gaussian
a function of all states. As a result, exact inference becomes infeasible. Concretely, the backward-forward algorithm
used in the E step of the Baum-Welch algorithm has time complexity $\mathcal{O}(T M K^{M+1})$, where $T$ is the length of the sequence,
$K$ is the number of states and $M$ is the number of Markov chains. Note however that the M step for FHMMs is
completely tractable and can therefore be calculated exactly. To work around the infeasibility of inference, several approximations
of the E step have been proposed: Ghahramani~et~al. devised \emph{inference using Gibbs sampling},
\emph{completely factorized variational inference} and \emph{structured variational inference}.
A fourth approach, the \emph{generalized backfitting algorithm}, was described by Jacobs~et~al.~\cite{Jacobs:2002fhmmbackfitting}.

\chapter{Features}
\label{chapter:features}
As already mentioned in section~\ref{motion-capture}, motions can be recorded using an optical marker-based
motion capture system. The following section discusses different approaches to represent such motions
and describes possible features that are used to recognize and classify them in later chapters.

\section{Marker Representation}
\label{section:marker}
\begin{figure}[h]
    \centering
    \subfigure{\includegraphics[width=0.18\textwidth, cfbox=gray 0.5pt 1pt]{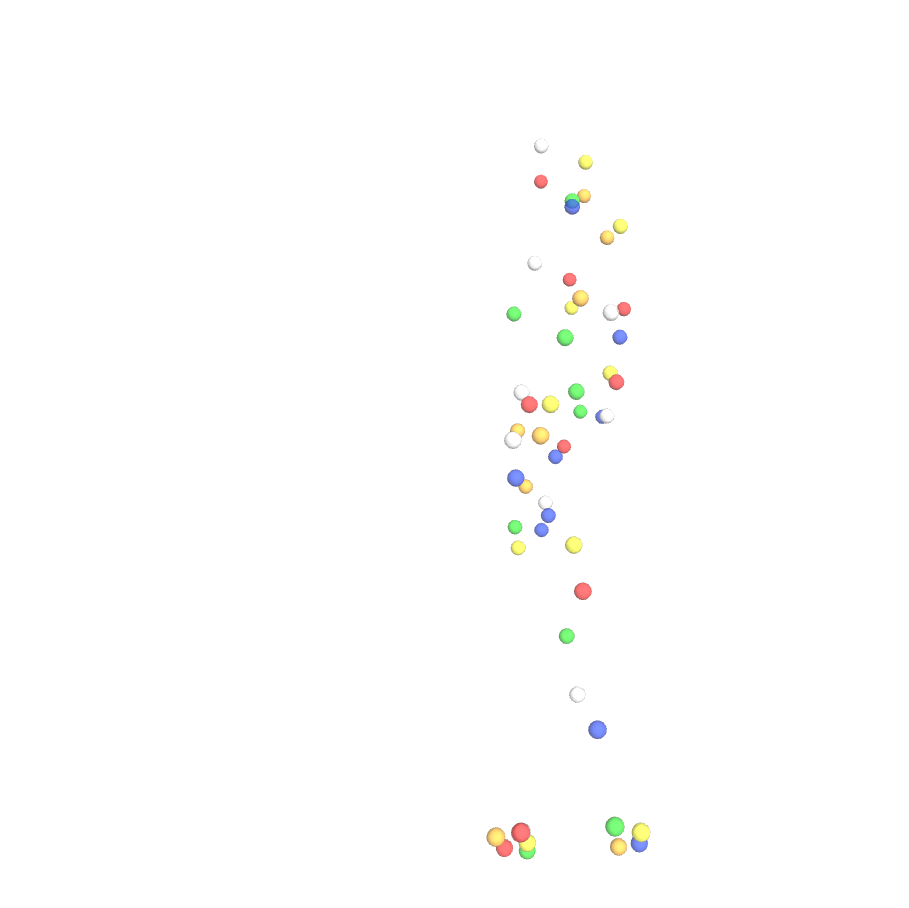}} 
    \subfigure{\includegraphics[width=0.18\textwidth, cfbox=gray 0.5pt 1pt]{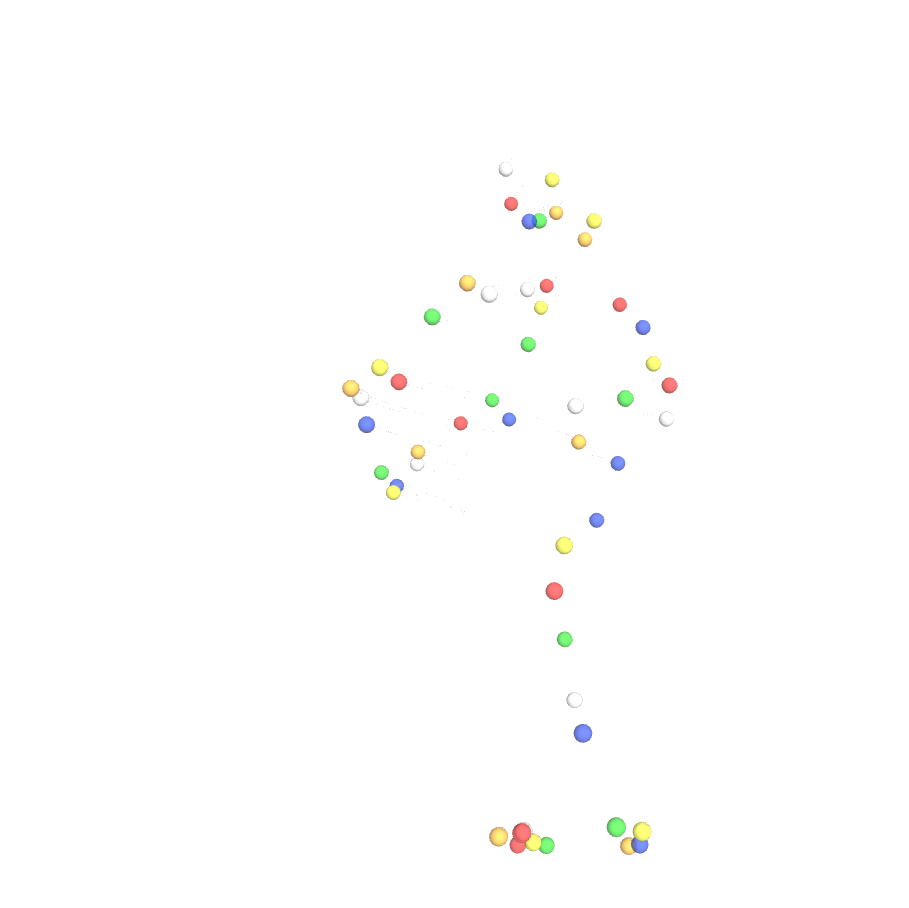}} 
    \subfigure{\includegraphics[width=0.18\textwidth, cfbox=gray 0.5pt 1pt]{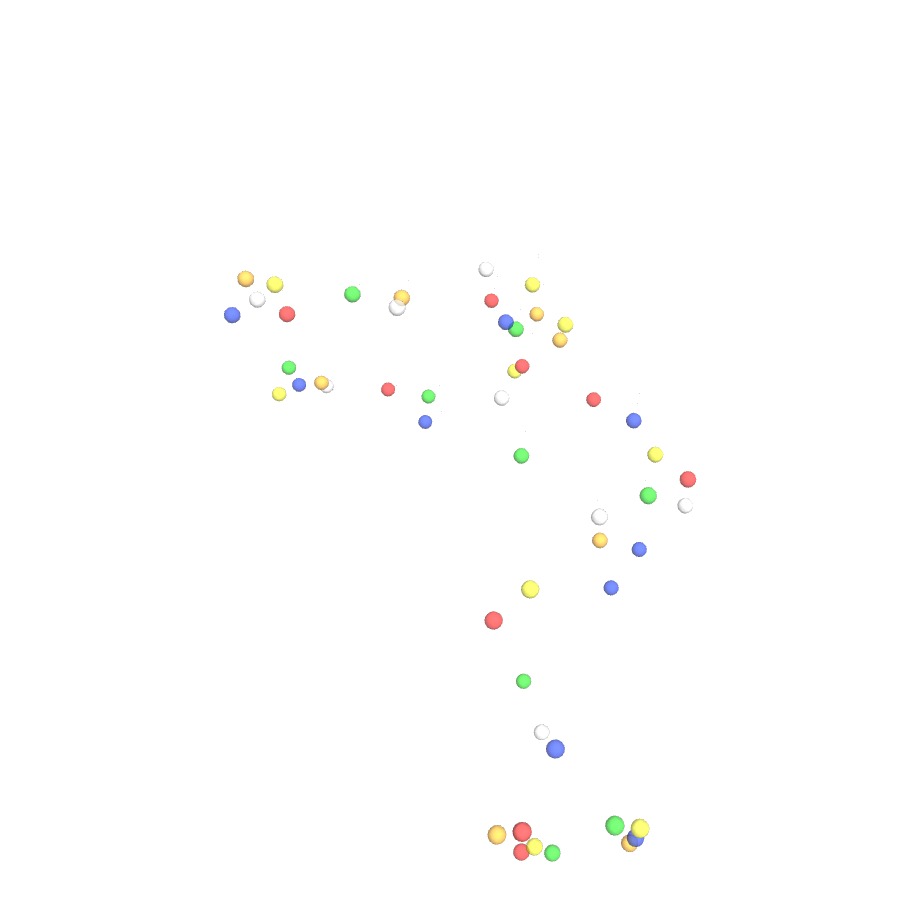}} 
    \subfigure{\includegraphics[width=0.18\textwidth, cfbox=gray 0.5pt 1pt]{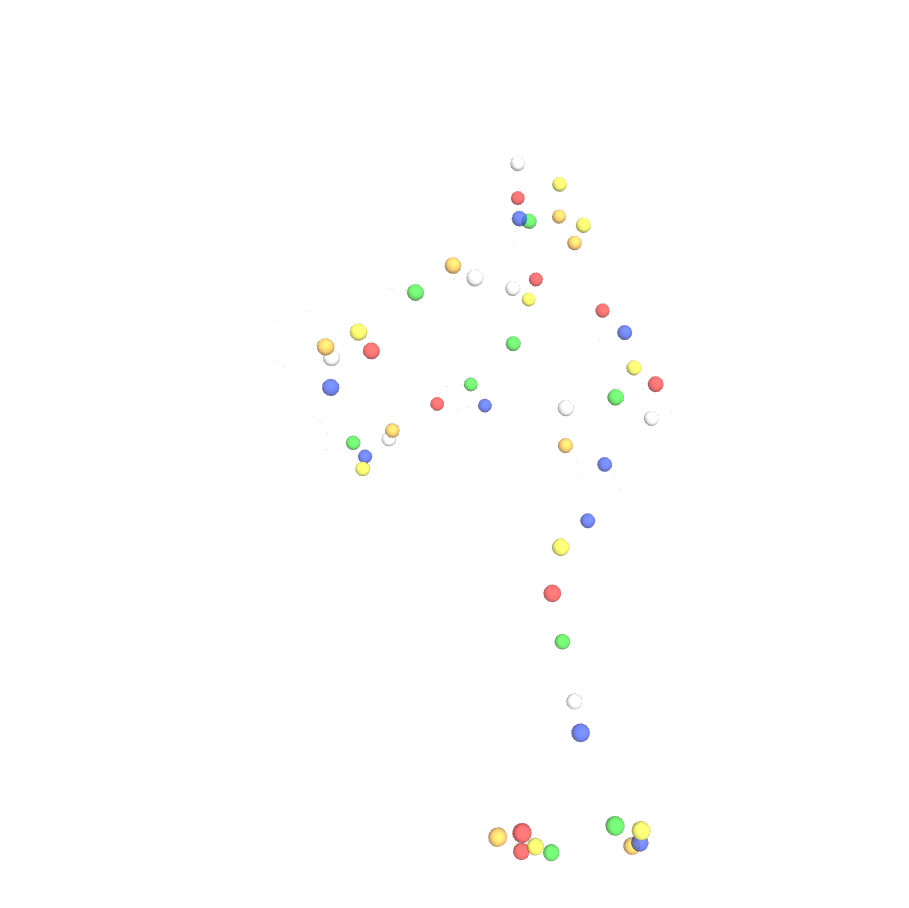}}
    \subfigure{\includegraphics[width=0.18\textwidth, cfbox=gray 0.5pt 1pt]{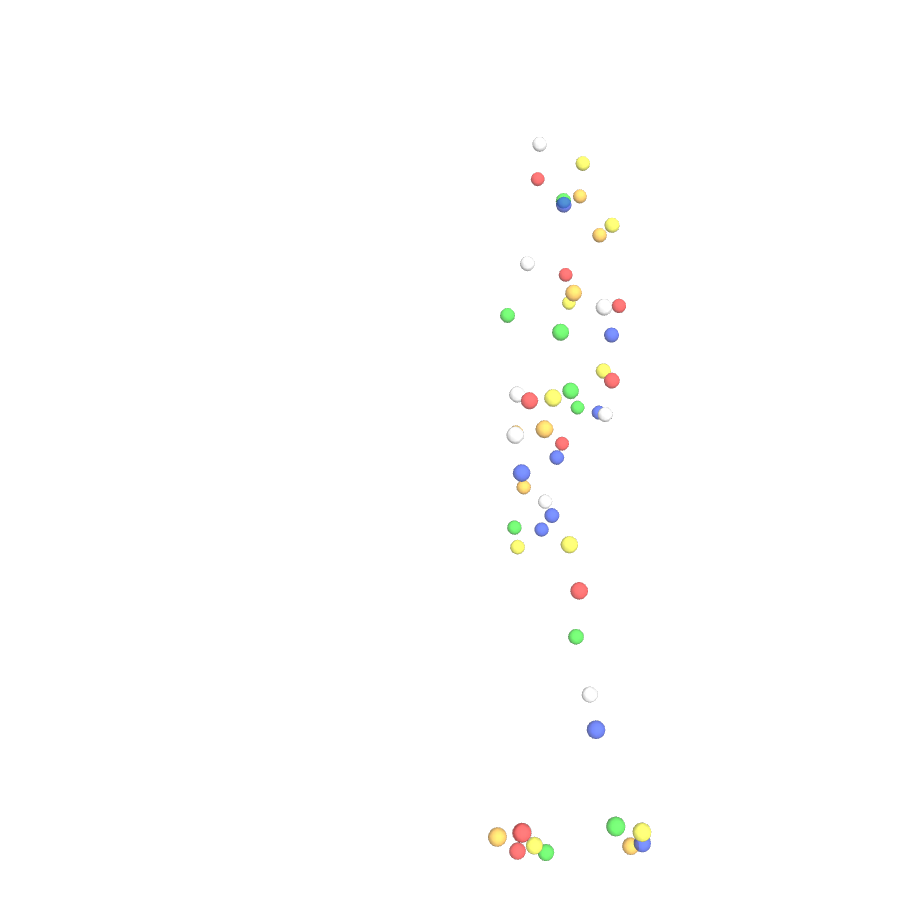}}
    \caption{Five key frames of a squatting motion with the location of the individual markers visualized. The subject looks to the left and is depicted from the side.}
    \label{fig:c3d-frames}
\end{figure}

A natural and obvious way to represent the recorded data is in 3-dimensional Cartesian space. For each
time sample $t$, the system records the location of each marker $n$:
\begin{equation} 
	\vec{r}_t^{(n)} = (x_t^{(n)},y_t^{(n)},z_t^{(n)}) \in \mathbb{R}^3.
\end{equation}
A complete motion or \emph{observation sequence} $\vec{O}$ is then represented by the all marker locations for all sampled time steps. A way to write this
is to ``unroll'' all marker locations for a given time step $t$ into the $t$-th row of an observation matrix:
\begin{equation} 
	\vec{O}_{cartesian} = \begin{bmatrix}
				x_1^{(1)} & y_1^{(1)} & z_1^{(1)} & \dots  & x_1^{(N)} & y_1^{(N)} & z_1^{(N)} \\ 
				x_2^{(1)} & y_2^{(1)} & z_2^{(1)} & \dots  & x_2^{(N)} & y_2^{(N)} & z_2^{(N)} \\
				\vdots    & \vdots    & \vdots    & \ddots & \vdots    & \vdots    & \vdots    \\
				x_T^{(1)} & y_T^{(1)} & z_T^{(1)} & \dots  & x_T^{(N)} & y_T^{(N)} & z_T^{(N)}
			  \end{bmatrix} \in \mathbb{R}^{T \times 3N},
\end{equation}
where $T$ is the number of time samples and $N$ is the number of markers. A visualization of a motion and
the respective marker locations is depicted in figure~\ref{fig:c3d-frames}.

A problem with this representation is that an absolute coordinate system is used. Consider for example
two running motions. Assume that in the first case the subject moves towards a defined point and in the second
case turns 45 degrees and repeats the motion almost identically. However,
since an absolute coordinate system is used, the values in the observation matrix $\vec{O}$ will
be very different for the two almost identical motions. The same problem occurs if the start location of two motions is offset.
Again, similar motions will have different marker positions in the observation matrix. In short, the representation in an absolute
Cartesian coordinate system is neither invariant to translation nor rotation. A coordinate system that is relative to the
recorded subject is desirable to allow for robust motion recognition.

\section{Joint Angle Representation}
\begin{figure}[h]
    \centering
    \includegraphics[height=300pt]{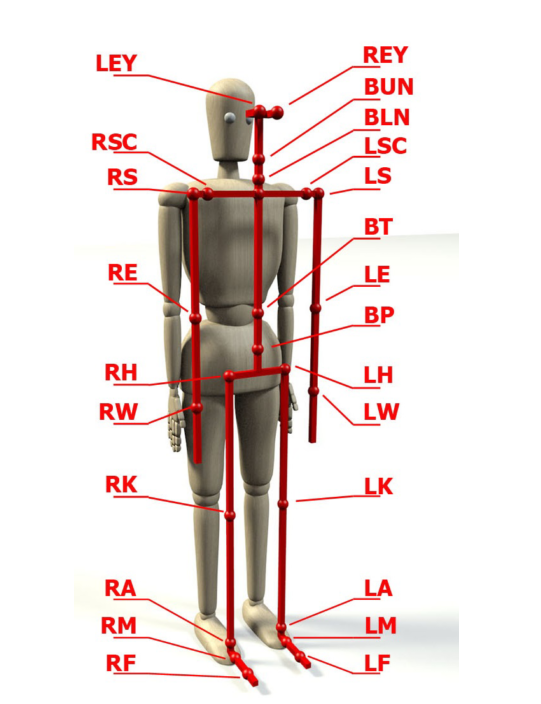}
    \caption{The kinematics of the MMM reference model~\cite{Terlemez:2014master}.}
    \label{fig:mmm_kinematics}
\end{figure}

The Master Motor Map (MMM,~\cite{Terlemez:2014master}) framework (see also section~\ref{mmm}) uses a relative coordinate system.
This is achieved by mapping the position of the \emph{physical} markers onto \emph{virtual} markers
on a reference model. To do so, the squared error between the physical and virtual markers is minimized by varying the pose
of the subject (as defined by its position and rotation in space as well as its joint angles) while maintaining the constraints of the reference
model. The optimization problem is solved by the reimplementation of the \emph{Subplex algorithm} as provided by the NLOpt
library~\cite{mandery:2015analyzing}. Figure~\ref{fig:mmm_kinematics} depicts the kinematics and shows the location and labels of all joints.

Some joints have multiple degrees of freedom (DoF). Take, for example, the body lower neck (BLN) joint that
has 3 DoF (to convince yourself that this is indeed the case, nod, shake your head and move your head from shoulder to shoulder).
In robotics, this is usually handled by splitting a joint with multiple degrees of freedom into multiple joints with a single
DoF each. In the case of the exemplary BLN joint, this means that the joints BLN\textsubscript{x}, BLN\textsubscript{y} and BLN\textsubscript{z} will replace it.
After this step, each joint has a single DoF and can therefore be represented as a scalar value that defines
the joint angle in radians. The complete joint configuration at time step $t$ can therefore be written as
\begin{equation} 
	\vecsym{\theta}_t = (\theta_t^{(1)},\theta_t^{(2)},\ldots,\theta_t^{(N)}) \in \mathbb{R}^N,
\end{equation}
where $N$ is the number of joints with a single DoF each. Similar to a representation in Cartesian space, a complete observation sequence
can then be written as:
\begin{equation} 
	\vec{O}_{mmm} = \begin{bmatrix}
				\theta_1^{(1)} & \theta_1^{(2)} & \dots  & \theta_1^{(N)} \\ 
				\theta_2^{(1)} & \theta_2^{(2)} & \dots  & \theta_2^{(N)} \\ 
				\vdots    	   & \vdots    		& \ddots & \vdots   	  \\
				\theta_T^{(1)} & \theta_T^{(2)} & \dots  & \theta_T^{(N)}
			  \end{bmatrix} \in \mathbb{R}^{T \times N},
\end{equation}
where $T$ denotes the number of time samples.

It is important to stress again that the the joint angle representation is relative to the subject.
However, some important information is lost: the position of the subject in space. To overcome this,
the absolute root position and the root rotation at each time step $t$ are included in the MMM framework as well:
\begin{align}
	\vec{r}_t^{(root)} &= (x_t,y_t,z_t) \in \mathbb{R}^3, \\
	\vecsym{\theta}_t^{(root)} &= (\theta_t^{(roll)}, \theta_t^{(pitch)}, \theta_t^{(yaw)}) \in \mathbb{R}^3.
\end{align}
However, these properties are once again given in an absolute coordinate system, suffering from the same
problem described above. Luckily, this is resolved easily by an affine transformation of the coordinate system
translating it such that the root position at $t = 0$ starts at the origin and rotating it such
that the $y$ axis points away from the front of the subject.
The following two equations describe the necessary calculations for each time step $t$:
\begin{align}
	\Delta\vec{r}_t^{(root)} &= \vec{r}_t^{(root)} - \vec{r}_0^{(root)}, \\
	\vec{\hat{r}}_t^{(root)} &= \vec{R}^{-1}~\Delta\vec{r}_t^{(root)}.
	\label{eq:normalization-root}
\end{align}
The first equation describes the translation, whereas the second equation describes the rotation. A
rotation matrix for roll, pitch and yaw angles is given in~\cite{craig:2005introduction}:
\begin{equation}
	\vec{R} = 
	\begin{bmatrix}
		\cos\alpha \cos\beta & \cos\alpha \sin\beta \sin\gamma - \sin\alpha \cos\gamma & \cos\alpha \sin\beta \cos\gamma + \sin\alpha \sin\gamma \\
		\sin\alpha \cos\beta & \sin\alpha \sin\beta \sin\gamma + \cos\alpha \cos\gamma & \sin\alpha \sin\beta \cos\gamma - \cos\alpha \sin\gamma \\
		-\sin\beta & \cos\beta \sin\gamma & \cos\beta \cos\gamma
	\end{bmatrix},
\end{equation}
where $\alpha := \theta_0^{(yaw)}$, $\beta := \theta_0^{(pitch)}$ and $\gamma := \theta_0^{(roll)}$.

\begin{figure}[h]
    \centering
    \subfigure[The first running motion \emph{without} normalization.]{\label{fig:absolute-relative-comparison1}\includegraphics[width=0.45\textwidth]{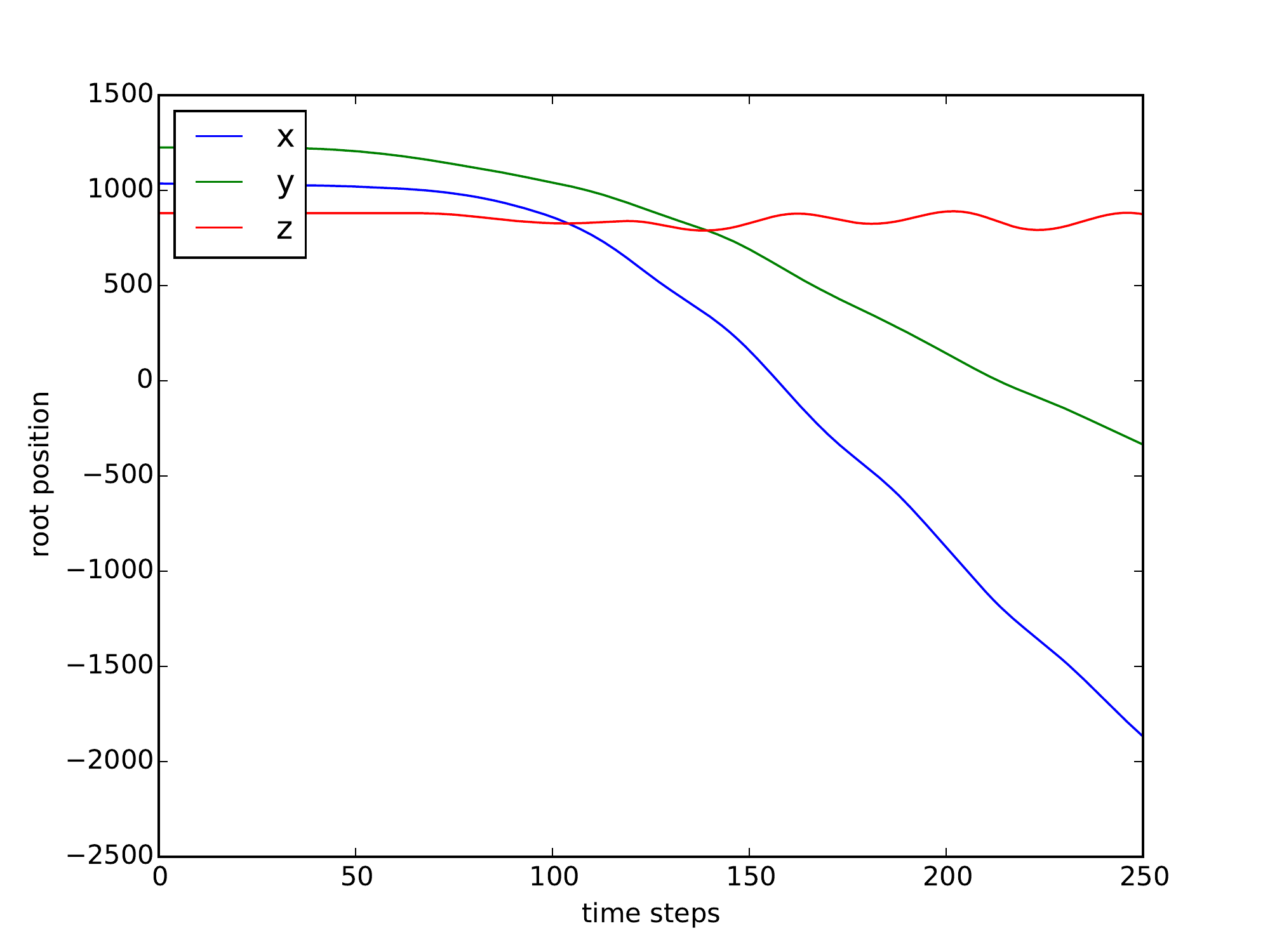}}
    \subfigure[The second running motion \emph{without} normalization.]{\includegraphics[width=0.45\textwidth]{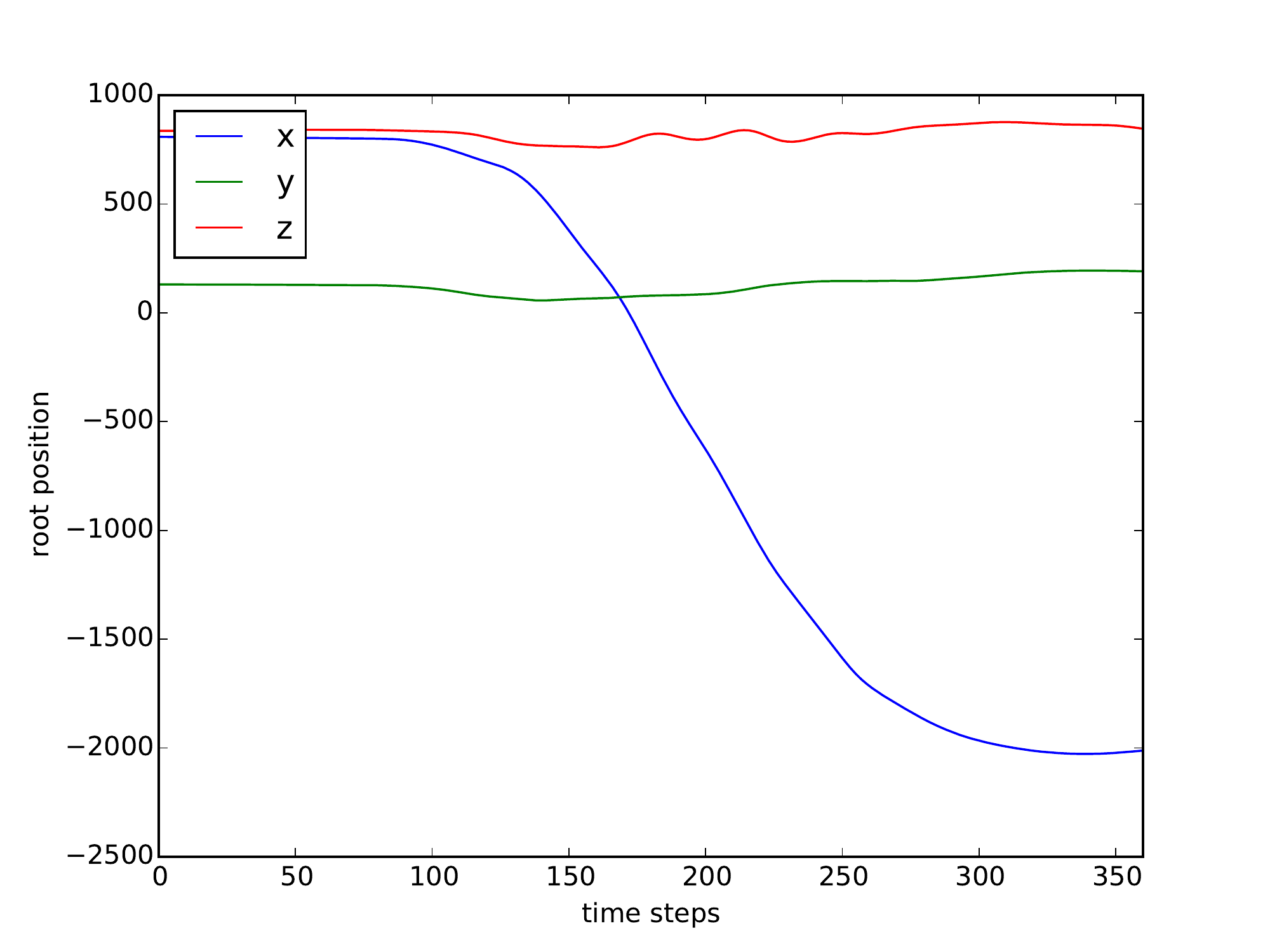}}
    \subfigure[The first running motion \emph{with} normalization.]{\label{fig:absolute-relative-comparison3}\includegraphics[width=0.45\textwidth]{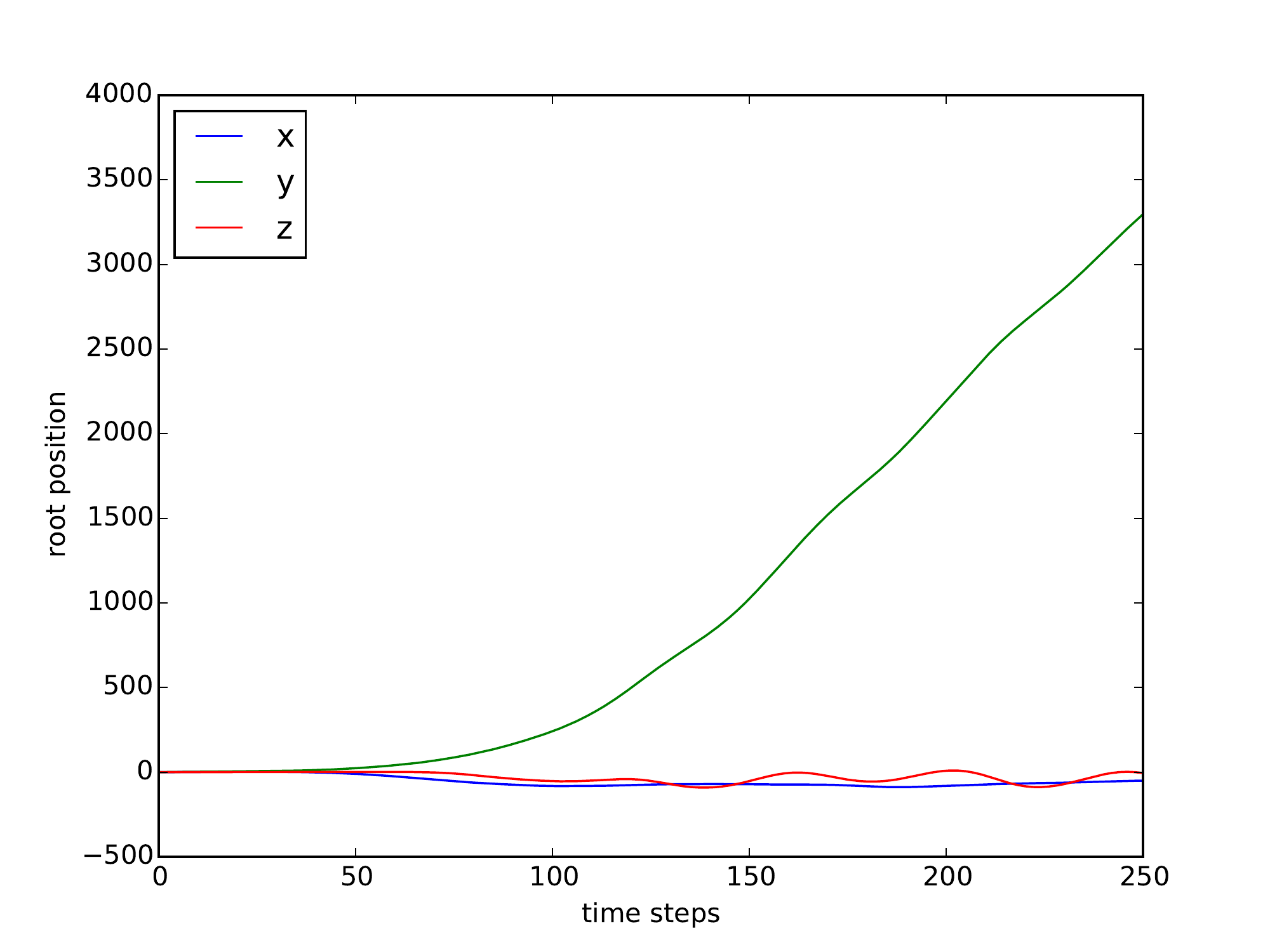}}
    \subfigure[The second running motion \emph{with} normalization.]{\includegraphics[width=0.45\textwidth]{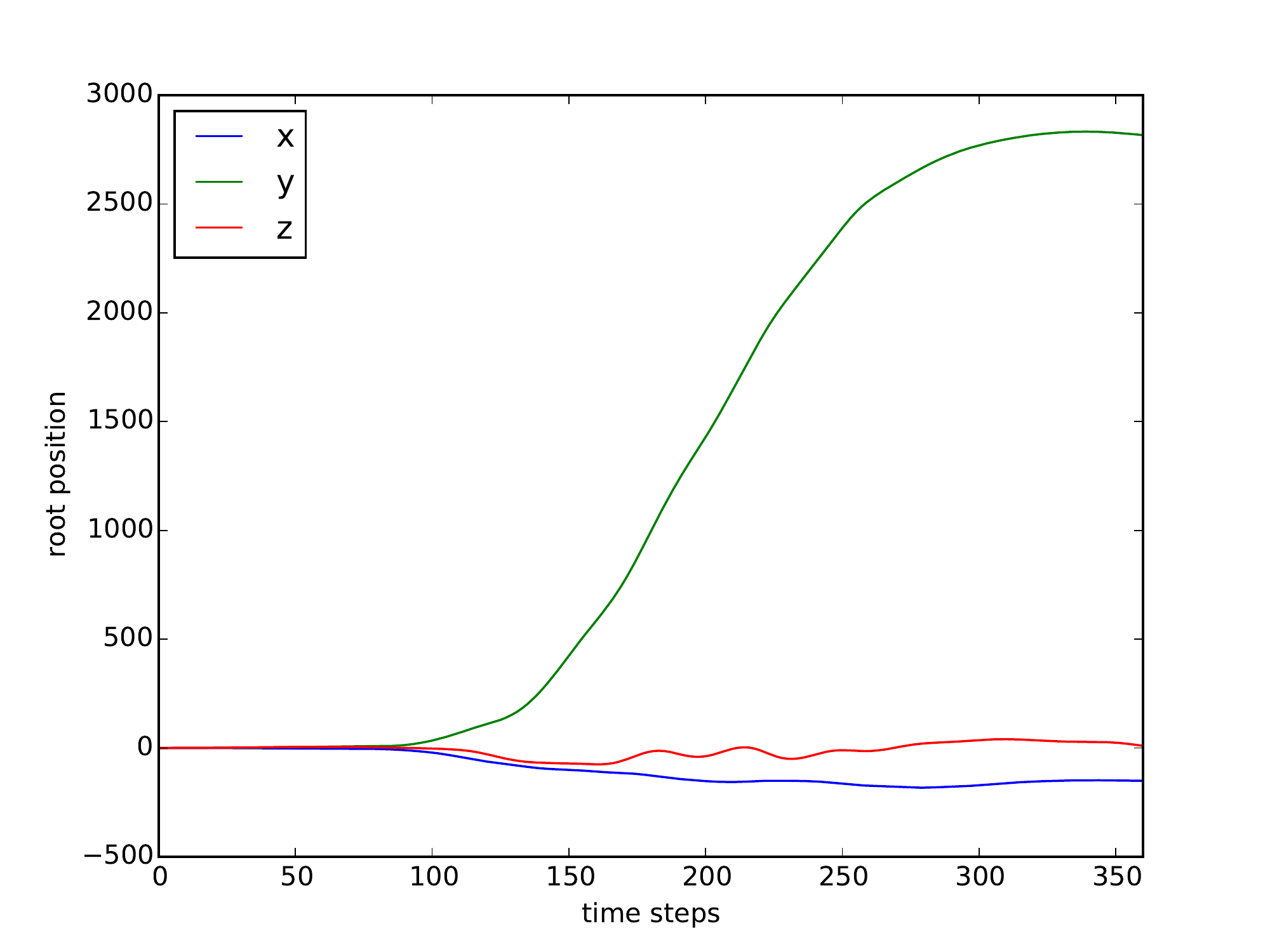}}

    \caption{Comparison between the root positions of two different running motions before and after normalization.}
    \label{fig:absolute-relative-comparison}
\end{figure}

Figure~\ref{fig:absolute-relative-comparison} plots the root position of two motions where the subject runs forward. The
same movements are depicted once before any normalization has been applied and once after normalization. Notice that without normalization,
the features are neither translation nor rotation invariant. This can be especially well seen in figure~\ref{fig:absolute-relative-comparison1}:
Although the subject is running only forward, the movement is split between the $x$ and $y$ components. This is because the subject moves
at an approximately 45 degree angle between the $x$ and $y$ axis of the absolute coordinate system. After normalization (figure~\ref{fig:absolute-relative-comparison3}),
the movement direction happens only in the direction of the $y$ axis of the transformed and now relative coordinate system.
Hence rotation and translation invariant features are obtained after normalization.

The root rotation must be normalized as well to make it comparable. Since the rotation is given in
angles, the normalization is straightforward:
\begin{equation}
	\vecsym{\hat{\theta}}_t^{(root)} = \vecsym{\theta}_t^{(root)} - \vecsym{\theta}_0^{(root)}.
\end{equation}
Notice that this work assumes that the roll, pitch and yaw angles are \emph{not} limited to the interval $[-\pi,\pi]$.
If necessary, this is easily achieved by correcting overflows by adding $2\pi$ to all following angles
(and similarly subtracting $2\pi$ for underflows).

Another interesting usage of the MMM reference model is that it allows normalization of the marker positions.
Recall that the marker positions were previously given in an absolute coordinate system (see section~\ref{section:marker}).
However, since the initial pose of the subject is known in the MMM reference model, this information can
be used to normalize the marker positions as well. This is done analogously to the normalization of
the root position for the position of each marker (see equation~\ref{eq:normalization-root}).

\section{Derived Features}
\label{section:features-derived}
Multiple additional features are computed under the MMM reference model. Firstly, an obvious extension
is to calculate the velocities and accelerations of all features that describe positions in Cartesian coordinate
space. Given that the velocity is the first derivative of the position and the acceleration is the first
derivative of the velocity, both properties are easily calculated by approximating the respective derivatives:
\begin{equation}
	\vec{v}_t = \frac{\vec{r}_{t+1} - \vec{r}_{t-1}}{2 \Delta t} \qquad \text{and} \qquad \vec{a}_t = \frac{\vec{v}_{t+1} - \vec{v}_{t-1}}{2 \Delta t},
\end{equation}
where $\vec{v}$ denotes the velocity, $\vec{a}$ the acceleration and $\Delta t$ is the time difference
between two subsequent samples (which is assumed to be equidistant over all samples). The velocity
and acceleration must be normalized as well. The normalization works similarly to
equation~\ref{eq:normalization-root} but normalizes each sample with the current pose of the respective
segment instead of normalizing each sample with the initial root pose.
An interesting modification of the velocities and accelerations is to reduce them to simple scalar values
by using their norm instead, e.g. the Euclidean one.

\begin{figure}[h]
    \centering
    \subfigure{\includegraphics[width=0.45\textwidth]{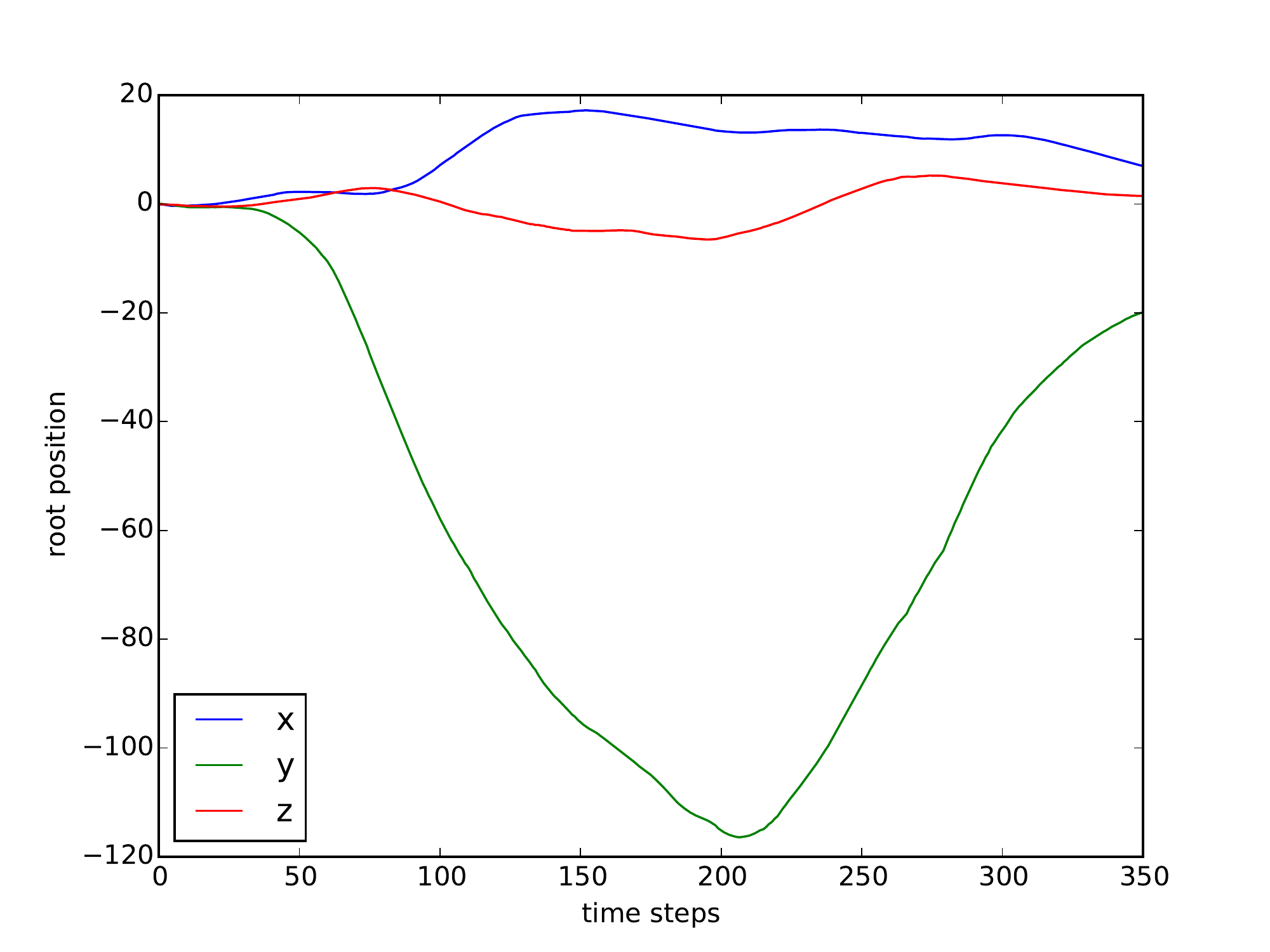}}
    \subfigure{\includegraphics[width=0.45\textwidth]{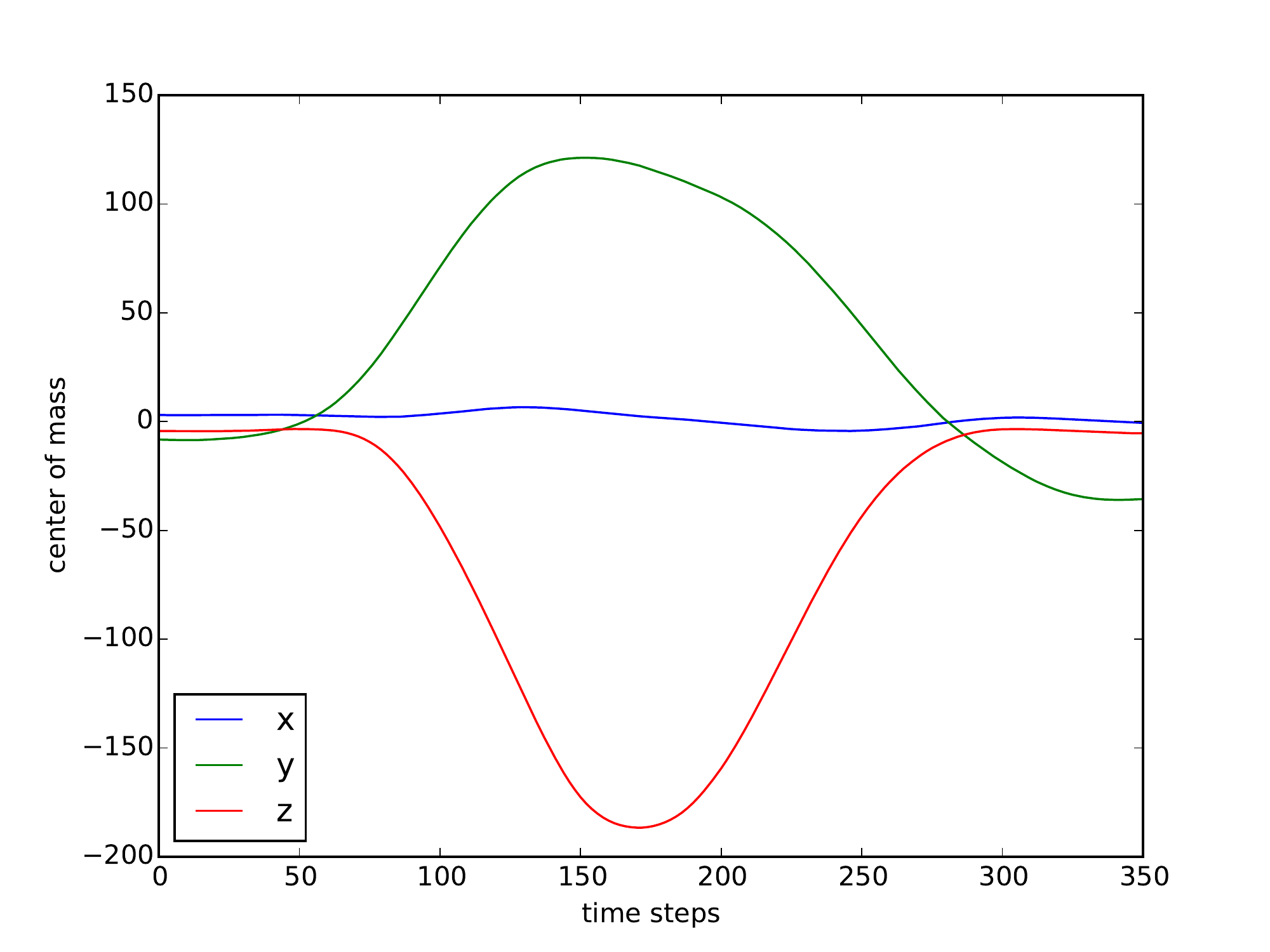}}
    \caption{Comparison between the root position (left) and center of mass (right) of the same bowing motion.}
    \label{fig:root-vs-com}
\end{figure}

Secondly, since the MMM reference model provides additional information about the subject, more advanced
features are computed as well. Two interesting dynamic properties are the \emph{center of mass} (CoM) and
the \emph{angular momentum}~\cite{popovic:2004angular}.
The CoM is conceptually similar to the root position discussed earlier in the sense
that it describes the position of a subject in 3-dimensional Cartesian space. However, while the root position is always
at a fixed point on the reference model, the center of mass is the barycenter of the subject. More concretely,
the center of mass is the average over the CoM positions of body segments weighted by their respective masses. 
To clarify this, consider a motion where the subject performs a deep bow. A bowing motion is interesting in this case
since the lower body remains mostly fixed in space while the upper body moves down. Figure~\ref{fig:root-vs-com} plots
the root position and the CoM of such a motion. Notice that the $z$ component of the root position
remains approximately constant although the upper body moves down during the bow. The $z$ component of the CoM on the other
hand first decreases while the subject bows down and then increases again as the subject comes back up.
Now consider the $y$ component of the root position: It decreases as the subject bows down since the hip of the subject moves backwards. Compare
this to the $y$ component of the CoM that instead increases as the subject bows down since that shifts the center of mass forward.
Like the root position, the CoM is given in an absolute coordinate system. The
normalization to a coordinate system relative to the subject is done analogous to the normalization
of the root position. The velocity and acceleration of the CoM are computed as well.

The angular momentum is a physical measure for the rotational configuration of an object or a system in 3D space.
In the MMM framework, the the angular momentum is calculated with respect to the center of mass at
each time step $t$ in all three spatial directions. The whole-body angular momentum is calculated as follows:
\begin{equation}
	\vec{L}_t = \sum_{i=1}^M \left( m^{(i)} (\vec{r}_t^{(i)} \times \vec{v}_t^{(i)}) + \vec{I}_t^{(i)} \vecsym{\omega}_t^{(i)} \right) \in \mathbb{R}^3 ,
\end{equation}
with
\begin{equation}
	\vec{r}_t^{(i)} = \vec{r}_t^{(CoM_i)} - \vec{r}_t^{(CoM)} \qquad\text{and}\qquad \vec{v}_t^{(i)} = \vec{\dot{r}}_t^{(CoM_i)} - \vec{\dot{r}}_t^{(CoM)}.
\end{equation}
The first part of the sum considers the angular momenta created by the orbital rotation of each segment around the whole-body center of mass. For each segment $i \in \{1,\ldots,M\}$,
$m^{(i)}$ describes its mass, $\vec{r}_t^{(i)}$ its position at time step $t$ w.r.t. the CoM and $\vec{v}_t^{(i)}$ its velocity at time step $t$ w.r.t. the CoM. The cross product is denoted as $\times$.
The second part of the sum takes the spin of each segment into account by computing the product of its inertia tensor $\vec{I}_t^{(i)}$ and its angular velocity $\vecsym{\omega}_t^{(i)}$.
The velocity $\vec{v}_t^{(i)}$ and the difference in CoM $\vec{r}_t^{(i)}$ must be normalized as previously described.

Thirdly, the position of body segments are used as features as well. Consider for example a waving
motion. In this case, the positions of the hands are an interesting feature. Similarly, the position
of the feet are interesting for other motions, e.g. a kick. The positions of the extremities must be
normalized. The velocities and accelerations are computed as previously described.

\section{Smoothing}
\label{section:feature-smoothing}
The described features can be noisy or contain errors. For example, noise is introduced during the
recording process. In addition, approximating the derivative for the computation of velocities and accelerations
can amplify errors and inaccuracies in the recorded data. Smoothing is used to reduce the impact
of these interferences. A signal can be smoothed using a wide variety of different filters. For a more complete
discussion, see~\cite{simonoff:2012smoothing}.

In this work, only a simple filter is briefly discussed: the \emph{moving average} or \emph{sliding window} filter.
In such a filter of length $W$, the mean of the surrounding $W$ data points is used instead of the individual data point:
\begin{equation}
	\vec{\hat{x}}_t = \frac{1}{W+1} \sum\limits_{j=-W/2}^{W/2}{\vec{x}_{t+j}},
\end{equation}
where $\vec{x}_t$ denotes a (potentially multi-dimensional) data point at time step $t$ and $\vec{\hat{x}}$ is the
smoothed version thereof. Including future samples into the average avoids introducing a time delay in the signal.
Notice that averaging over future samples is possible if the smoothing is applied off-line.

\begin{figure}[h]
    \centering
    \subfigure{\includegraphics[width=0.45\textwidth]{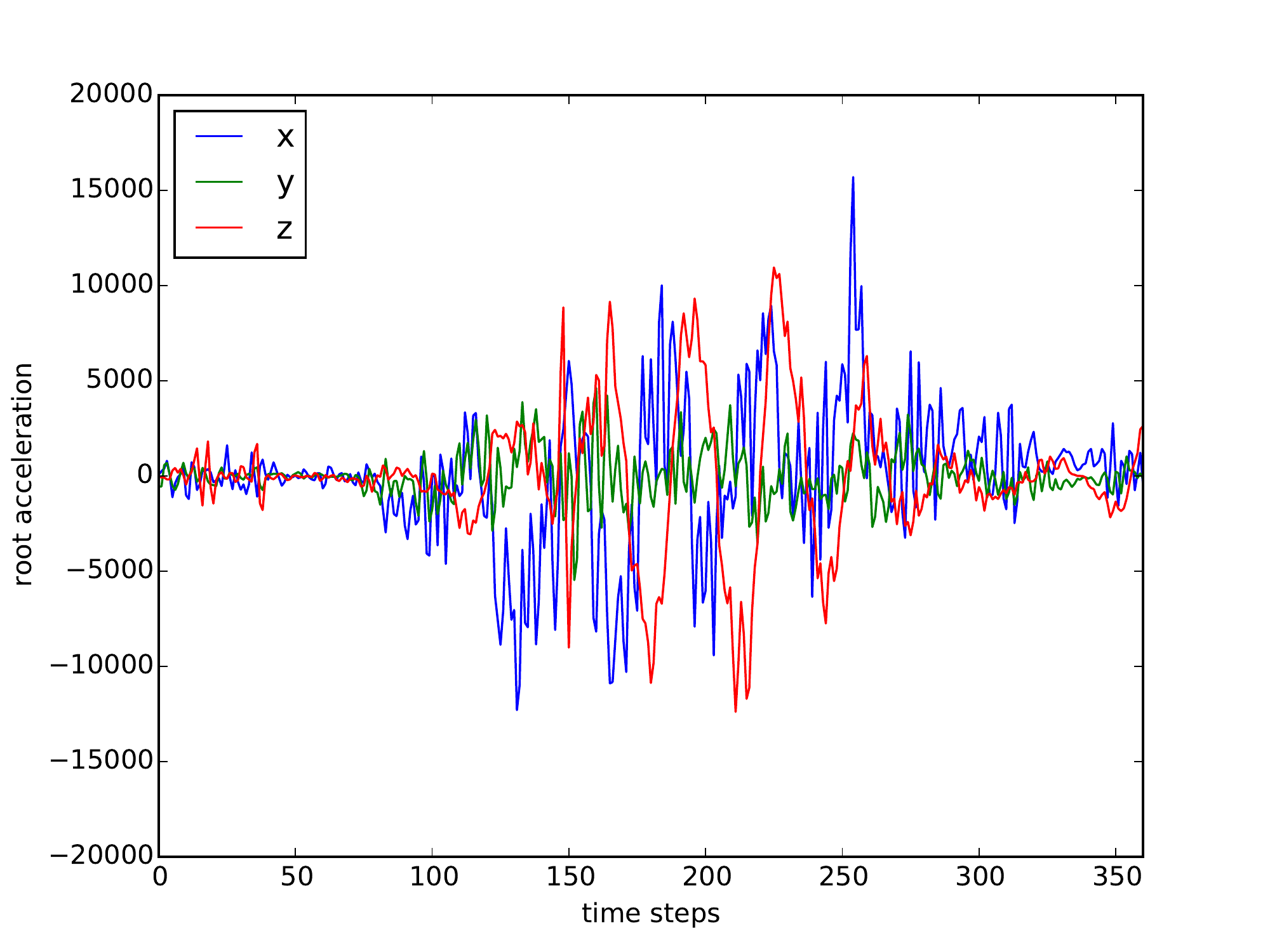}} 
    \subfigure{\includegraphics[width=0.45\textwidth]{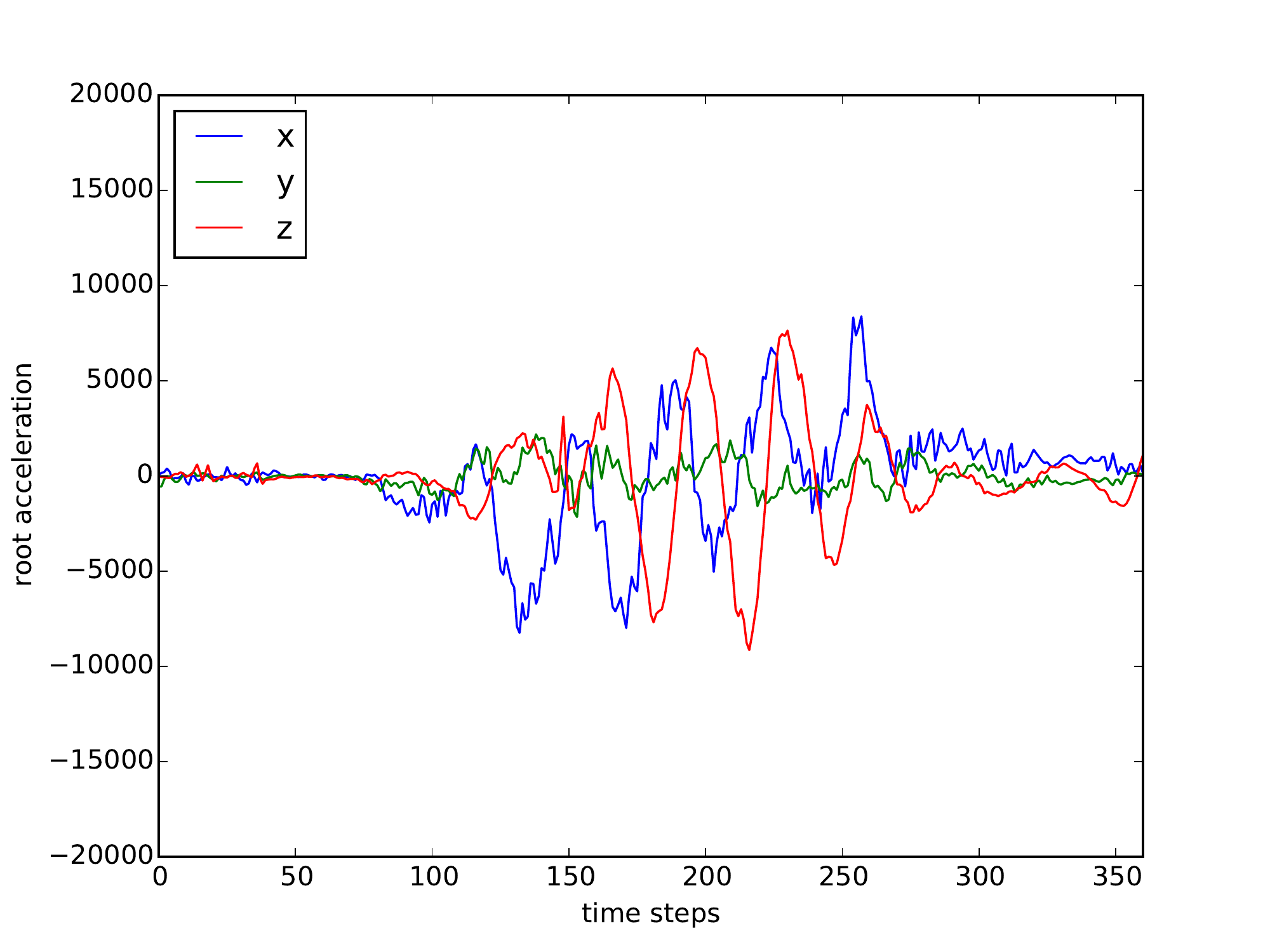}}
    \caption{The components of a subject's root acceleration during a running motion before (left) and
    after (right) a moving average filter with $W=3$ was applied.}
    \label{fig:smoothing-before-after}
\end{figure}

Figure~\ref{fig:smoothing-before-after} compares the root acceleration of
a subject before and after applying a moving average filter. Notice that the the original signal
is very noisy. The smoothed signal maintains the overall structure of the signal but reduces the
amount of jitter. Smoothing is a useful preprocessing step before feeding the features into
a model.

\section{Scaling}
\label{section:feature-scaling}
Feature scaling is another preprocessing step. Take for example the joint angles and
the root position from the previous sections. The joint angles are physically constrained to a very narrow
value range, whereas the root position can potentially grow very large if the subject travels a large
distance from the start position. It should be obvious from this example that features are on very different scales.
This difference in scale becomes a problem if $k$-means clustering is used to initialize an HMM's emission
distribution parameters. If the data is on very different scales across dimensions, $k$-means will not
find clusters that properly fit the data because the same distance measure is minimized across dimensions.
This, in turn, results in bad estimates of the emission distribution parameters which results in vanishing probabilities
and numerical instabilities during inference. To counter this, \emph{feature scaling} is performed. A very simple but effective strategy is to scale the features
such that each feature's values are in the same range, e.g. $[-1,1]$. This is achieved by applying
the following equation to each individual feature $x$ over all $T$ time steps:
\begin{equation}
	\hat{x}_t = 2 \cdot \frac{x_t - \min\limits_{i\in\{1,\dots,T\}}{x_i}}{\max\limits_{i\in\{1,\dots,T\}}{x_i} - \min\limits_{i\in\{1,\dots,T\}}{x_i}} - 1.
\end{equation}
Notice that each feature needs to be scaled over all samples, not per-sample. Furthermore, if feature
scaling is used, the computation of the minimum and maximum are computed on the training data.
New samples (e.g. from the test dataset when evaluating or unknown motions when used productively)
are then simply scaled using the previously computed values. Otherwise the features that were
used to train the model and the features that are used to recognize unknown motions would end up on a different scales.

\cleardoublepage

\chapter{Classification}
\label{chapter:classification}
This chapter discusses methods to recognize and classify motions represented by features as described
in the previous chapter. In the first section, the general concepts of Hidden Markov Models introduced
in chapter~\ref{mmm} are concretely discussed for the case of \emph{motion recognition}. In motion
recognition, the goal is to encode a motion into a Hidden Markov Model and compute a measure that describes
how likely an unknown motion is under the model. In the second section, motion recognition is extended
to perform \emph{multi-class classification}. In contrast to motion recognition, the goal is now
to assign an unknown motion exactly one class out of a (potentially large) set of possible classes.
Lastly, methods for performing \emph{multi-label classification} are introduced. In contrast to the
multi-class classification task, an unknown motion can have many labels that assign it to multiple classes.

To avoid confusion and to emphasize the distinction between multi-class (where one label assigns each motion one class)
and multi-label (where multiple labels assign each motion multiple classes), multi-class classification
is referred to as \emph{single-label classification} hereinafter.

\section{Motion Recognition}
\label{section:motion-recognition}
Hidden Markov Models are a popular choice for encoding human whole-body motions~\cite{Takano:2006primitive, Kulic:2007clustering, Kulic:2008ib}.
This section describes some properties of HMMs in depth and discusses properties and problems
that are especially relevant when dealing with motions.

\subsection{Emission Distribution}
Recall that the emission distribution models the observed data. Since in this case motions are observed, and all previously discussed features
are continuous, the emission distribution must also be continuous. Typically, a Gaussian distribution
or a mixture model thereof is used to model this case~\cite{Kulic:2008ib}. Since this work uses a
multivariate Gaussian distribution, the following discussion focuses on this distribution.

A multivariate Gaussian or normal distribution is defined by two parameters, its mean vector $\vecsym{\mu} \in \mathbb{R}^D$ and its
covariance matrix $\vecsym{\Sigma} \in \mathbb{R}^{D\times D}$, where $D$ is the dimension of the feature vector.
The probability density function (pdf) is then given by
\begin{equation}
	f(\vec{x}) = \abs{\vecsym{\Sigma}}^{-\frac{1}{2}} (2 \pi)^{-\frac{D}{2}} \exp\left(-\frac{1}{2} (\vec{x}-\vecsym{\mu})^T \vecsym{\Sigma}^{-1} (\vec{x}-\vecsym{\mu})\right).
\end{equation}
In an HMM, the emission of each state $k$ is governed by its mean vector $\vecsym{\mu}_k$ and its covariance matrix
$\vecsym{\Sigma}_k$. When dealing with motions, the covariance matrices are often constrained to be diagonal to avoid numerical problems~\cite{Kulic:2007clustering, Kulic:2007bf}.

\subsection{Topologies}
\label{section:topologies}
An important property of a Hidden Markov Models is that it uses hidden states. The transition probabilities
between the $K$ states are given by the transition matrix $\vec{A} \in [0,1]^{K \times K}$, while $\vecsym{\pi} \in [0,1]^K$
defines the start probabilities for each state. By constraining the transition matrix (and as a result the start probabilities),
different \emph{topologies} can be realized. This is easily done by initializing the transition matrix and the
start probabilities with some entries set to zero. During training, all probabilities that were initially
set to zero will remain at zero~\cite{Rabiner:1989hs}.

\begin{figure}[h]
  \centering
  \subfigure{\includegraphics[width=0.45\textwidth]{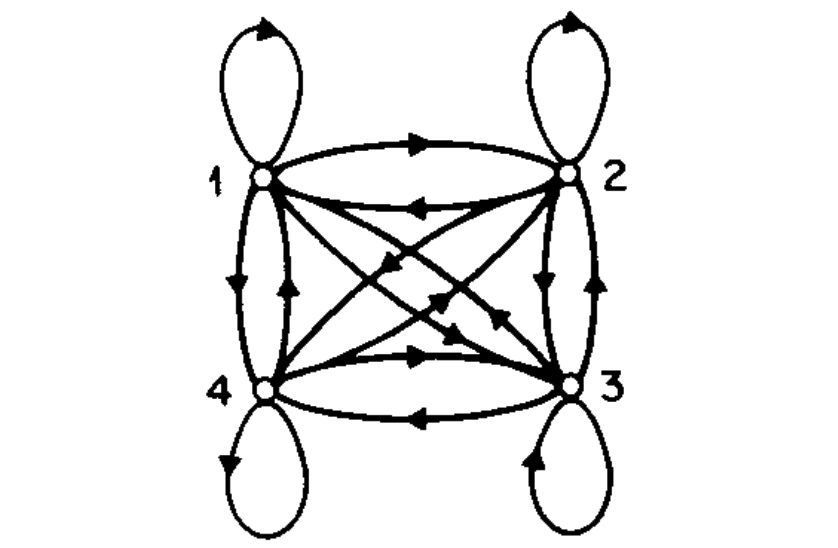}} 
  \subfigure{\includegraphics[width=0.45\textwidth]{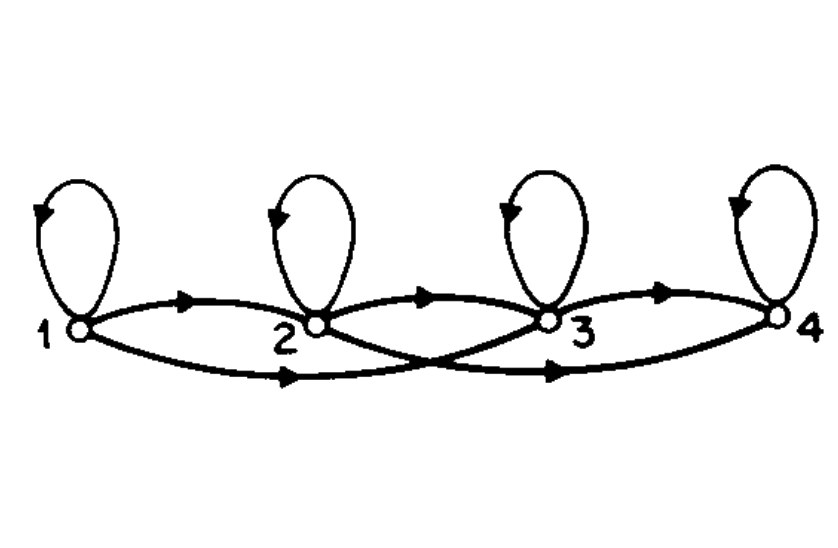}}
  \caption{Illustration of a $4$-state ergodic HMM (left) and a $4$-state left-to-right HMM with with $\Delta = 2$ (right)~\cite{Rabiner:1989hs}.}
  \label{fig:hmm-topologies}
\end{figure}

If the transition matrix is not constrained, a transition from any given state to every other state can occur.
Such an HMM is usually referred to as \emph{fully connected} or \emph{ergodic}. Another popular topology is
the \emph{left-to-right topology} or \emph{Bakis topology}~\cite{bakis:1976continuous}. In such a topology,
the states are thought to be aligned sequentially from left to right. At each state, only a transition to a
state that is right of the current state or
a self-transition is allowed. The model can thus only be traversed from left to right, hence the name.
In a left-to-right model, the start probabilities are set to $\vecsym{\pi} = (1, 0, \ldots, 0)$ while 
the transition matrix takes the following shape:
\begin{equation}
	\vec{A} = 
	\begin{bmatrix}
		a_{1,1} & a_{1,2} & \dots  & a_{1,(K-1)}     & a_{1,K} \\
		0 	    & a_{2,2} & \dots  & a_{2,(K-1)}     & a_{2,K} \\
		\vdots  & \vdots  & \ddots & \vdots           & \vdots \\ 
		0	    & 0		  & \dots  & a_{(K-1),(K-1)} & a_{(K-1),K} \\ 
		0 	    & 0       & \dots  & 0               & a_{K,K} \\
	\end{bmatrix}.
\end{equation}
The left-to-right constraint can thus be written as:
\begin{equation}
	a_{i,j} = 0, \qquad j < i.
\end{equation}
To avoid skipping too many states while traversing from left to right, an additional constraint is introduced:
\begin{equation}
	a_{i,j} = 0, \qquad j > i + \Delta.
\end{equation}
$\Delta$ limits the maximum number of allowed skips~\cite{Rabiner:1989hs}.
The left-to-right topology is frequently used for the recognition of motions~\cite{Takano:2006primitive, Kulic:2007clustering}.
Both, the ergodic and the left-to-right topology are visualized in figure~\ref{fig:hmm-topologies}. Note
that, due to the variate of possible constraints for the transition matrix, other topologies are possible which are not discussed here.

\subsection{Parameter Initialization}
\label{section:parameter-initialization}
An interesting problem that arises is how to initialize the values of the transition matrix and the start probabilities as well as the
means and covariance matrices of the emission distributions. Since the Baum-Welch algorithm
does not necessarily converge to a global maximum but rather to a local one~(see chapter~\ref{mmm}), a proper initial estimate is important
to increase the chances of finding the global maximum during training.
According to~\cite{Rabiner:1989hs}, the start probabilities and transition matrix can either be initialized by a random
(while maintaining the constraint that the respective probabilities must sum to one)
or uniform estimation. Note that the constraints imposed by the topology choice must also be maintained.

The initialization of the mean vectors and covariance matrices of the emission distribution is more complicated. While a random
initialization is possible, it is beneficial to perform an initial estimate of the underlaying distribution.
A popular approach works as follows: The basic idea is to find $K$ clusters that correspond to the $K$ states of the Hidden Markov Model.
The mean  $\vecsym{\mu}_k$ and covariance matrix $\vecsym{\Sigma}_k$ for each state $k$ are then estimated over the respective
$k$-th cluster~\cite{Rabiner:1989hs}.
The necessary clustering can be performed by hand. Alternatively an unsupervised clustering algorithm like the \emph{k-means algorithm}
can be used to automatically cluster the data. The k-means algorithm works by alternately assigning data samples
to a cluster given the current parameters and then updating the cluster center means such that the distance from the previously
associated samples is minimized. \emph{Lloyd's algorithm} is often used to solve the clustering
problem efficiently~\cite{kanungo:2002efficient}.

\subsection{Training and Recognition}
After deciding on the \emph{hyperparameters} of the model (namely the number of states, the topology
and the initialization method for the parameters of the model), the model can be trained. The training of an
HMM is performed efficiently using the Baum-Welch algorithm~\cite{Baum:1970maximization}: For each training sequence
$\vec{O}$ the parameters of the model are updated by first computing the expected likelihood given the current parameters (expectation step)
and then updating the parameters such that the expected quantity from the expectation step is maximized (maximization step). When performed iteratively, the likelihood
is maximized. The E and M steps are repeated until convergence or until a fixed number of iterations
have been performed.

A common problem when using the Baum-Welch algorithm are numerical instabilities. This is due
to the fact that the probabilities during the forward or backward pass can become extremely small since they
are multiplied at each time step. This becomes worse as the sequences grow longer. Since motions are usually
at least a couple of hundred samples long, this becomes a very real problem when training HMMs on motions.
The problem is mitigated by either using a scaling technique~\cite{Rabiner:1989hs}
or by adapting the algorithm such that it uses the logarithm of the probabilities instead~\cite{mann:2006numerically}.

Another important property is that the training is \emph{unsupervised}. This means that no target value
like a label is necessary to learn the parameters of the HMM. However, if multiple HMMs are used to classify
motions into classes, \emph{supervised learning} becomes important. This will be discussed in the next section.
Finally, HMMs can be trained \emph{off-line} and \emph{on-line}. In off-line training, the HMM is trained
only once and the parameters are kept fixed even if previously unseen observation sequences become available. Most
discussion in the literature assume off-line learning, e.g.~\cite{Rabiner:1989hs}.
In on-line learning on the other hand, the HMMs are trained incrementally as new data becomes available.
Kulić~et~al.~\cite{Kulic:2011incremental} describe such a system. This work only considers the off-line approach.

After the model has been trained, recognition is performed by calculating the likelihood under the
model $\lambda$ for an unknown observation sequence $\vec{O}$:
\begin{equation}
	p(\vec{O} \mid \lambda).
\end{equation}
This is done efficiently by the forward algorithm~\cite{Rabiner:1989hs}. Since the forward algorithm is used during training
as well, the same underflow issues as discussed earlier apply. Notice that the likelihood can become larger
than $1$ by definition, hence $0 \leq p(\vec{O} \mid \lambda) < \infty$. Since the likelihood can become both very small (that is, very close to zero but \emph{not} negative)
and very large, the logarithmic likelihood (\emph{loglikelihood}) is usually computed and presented.
A strongly positive loglikelihood thus indicates a motion that has been strongly recognized by the model whereas
a strongly negative loglikelihood indicates that the motion has not been recognized by the model at at. Finding
such a \emph{decision boundary} will be discussed later in this chapter.

\subsection{Extension to Factorial Hidden Markov Models}
\label{section:recognition-fhmm}
The previously discussed concepts apply equally to Factorial Hidden Markov Models. However, three additional
considerations are of interest: the number of Markov chains as an additional hyperparameter, efficient training of the FHMM
and computation of the likelihood under the model.

Firstly, the number of chains is an important hyperparameter since it directly controls the complexity
of time series that an FHMM can represent. However it also comes at the cost of increasing the computational
complexity of both the training and the evaluation given an unknown observation sequence (see chapter~\ref{fhmm}).
In the literature, few chains are usually used. For example, Kulić~et~al.~\cite{Kulic:2008ib} use
only two chains in their work. This does not seem like a lot at first. However, assume that an FHMM with
$15$ states and $2$ chains is used. This results in $15^2 = 225$ possible state combinations. Compared to
an HMM with $15$ states, even such a relatively small FHMM can already represent vastly more
history than its HMM counterpart.

Exact training of FHMMs is computationally very expensive and can be, depending
on the parameters, even infeasible. Four different approximations were already briefly discussed in chapter~\ref{fhmm}.
However, Kulić~et~al.\cite{Kulic:2008ib} proposed a fifth approach that the authors used to train FHMMs
on motion data. The authors further showed that their approach is at least as good as and in most cases
even better than the exact inference algorithm when using it to train FHMMs on motion data. This makes
their approach especially interesting for this work. The algorithm devised by the authors works as follows:
The FHMM is trained \emph{sequentially}. This means that each FHMM starts with a single chain. Inference
is then performed using the standard Baum-Welch algorithm on the training data. For the next chain $m$,
the \emph{residual error} between the previously trained chains and the $n$-th training sample $\vec{O}^{(n)} \in \mathbb{R}^{T \times D}$ is computed for
each time step $t \in \{1,\ldots,T\}$:
\begin{equation}
	\vec{e}_t^{(n)} = \frac{1}{W} \left(\vec{o}_t^{(n)} - \sum\limits_{i=1}^{m-1}{W \vec{c}_t^{(i)}}\right) \in \mathbb{R}^D,
\end{equation}
where $\vec{e}_t^{(n)}$ is the residual error between the frame at time step $t$ of the $n$-th training 
sample and the summed contributions of the previous $m-1$ chains. Each chain's contribution
is weighted by $W = 1/M$, where $M$ is the number of all chains.
Finally, the contribution for each chain $m$ at time step $t$ is computed as follows:
\begin{equation}
	\vec{c}_t^{(m)} = \sum\limits_{k=1}^{K}{\vecsym{\mu}_k^{(m)}\gamma_{t,k}^{(m)}} \in \mathbb{R}^D,
\end{equation}
where $K$ denotes the number of states, $\vecsym{\mu}_k^{(m)}$ is the $D$-dimensional mean vector of the
emission distribution of the \emph{already trained}
chain $m$ in state $k$. Furthermore,  $\gamma_{t,k}^{(m)}$ denotes the probability that state $k$ in chain $m$ is active
at time step $t$. Algorithm~\ref{alg:fhmm-seq} summarizes the described training procedure.

\begin{algorithm}[h]
  \SetAlgorithmName{Algorithm}{} 
  
  initialize first chain \\
  train first chain on time series $\vec{O}^{(n)} = (\vec{o}_t^{(n)})$ using the Baum-Welch algorithm \\
  \For{$m = 2,\ldots,M$}{
    initialize next chain $m$ \\
    compute residual errors $\vec{e}_1^{(n)}, \ldots, \vec{e}_T^{(n)}$ \\
    train chain $m$ on time series $\vec{E}^{(n)} = (\vec{e}_t^{(n)})$ using the Baum-Welch algorithm \\
  }
  \caption{The sequential training algorithm for FHMMs in pseudo code~\cite{Kulic:2008ib}.}
  \label{alg:fhmm-seq}
\end{algorithm}

A noteworthy and convenient property of the sequential training algorithm is that it uses procedures that are
already used when training HMMs. More concretely, the training of each chain is done by an unmodified
version of the well-known Baum-Welch algorithm. Computing $\gamma_{t,k}^{(m)}$ for the residual error
is achieved just as easily by using the standard forward-backward algorithm.

Lastly, the likelihoods under the FHMM must be calculated as well. This is done using the
exact algorithm given in~\cite{Ghahramani:1997id}. In the more concrete case of the sequential training algorithm,
the necessary means and covariances are computed as follows:
\begin{equation}
  \vecsym{\mu}_{k_1,\dots,k_M}    = W \sum\limits_{m=1}^{M}{\vecsym{\mu}_{k_m}^{(m)}} \qquad\text{and}\qquad
  \vecsym{\Sigma}_{k_1,\dots,k_M} = W^2 \sum\limits_{m=1}^{M}{\vecsym{\Sigma}_{k_m}^{(m)}}.
\end{equation}
Notice that this must be done for each possible combination of states across all $M$ chains, as indicated
by the index $k_1,\ldots,k_M$. Take, for example, an FHMM with $3$ states and $2$ chains. This would result in $3^2 = 9$~combinations, with the following mean vectors:
\begin{align}
  \vecsym{\mu}_{1,1} =& W \left(\vecsym{\mu}_1^{(1)} + \vecsym{\mu}_1^{(2)} \right), &
  \vecsym{\mu}_{1,2} =& W \left(\vecsym{\mu}_1^{(1)} + \vecsym{\mu}_2^{(2)} \right), &
  \vecsym{\mu}_{1,3} =& W \left(\vecsym{\mu}_1^{(1)} + \vecsym{\mu}_3^{(2)} \right), \\
  \vecsym{\mu}_{2,1} =& W \left(\vecsym{\mu}_2^{(1)} + \vecsym{\mu}_1^{(2)} \right), &
  \vecsym{\mu}_{2,2} =& W \left(\vecsym{\mu}_2^{(1)} + \vecsym{\mu}_2^{(2)} \right), &
  \vecsym{\mu}_{2,3} =& W \left(\vecsym{\mu}_2^{(1)} + \vecsym{\mu}_3^{(2)} \right), \\
  \vecsym{\mu}_{3,1} =& W \left(\vecsym{\mu}_3^{(1)} + \vecsym{\mu}_1^{(2)} \right), &
  \vecsym{\mu}_{3,2} =& W \left(\vecsym{\mu}_3^{(1)} + \vecsym{\mu}_2^{(2)} \right), &
  \vecsym{\mu}_{3,3} =& W \left(\vecsym{\mu}_3^{(1)} + \vecsym{\mu}_3^{(2)} \right).
\end{align}
The $9$ covariance matrices would be calculated analogously. The emission distribution for each
state combination is then given by a multivariate Gaussian distribution with mean $\vecsym{\mu}_{k_1,\dots,k_M}$ and
covariance $\vecsym{\Sigma}_{k_1,\dots,k_M}$~\cite{Jacobs:2002fhmmbackfitting, Kulic:2008ib}.

\section{Single-Label Classification}
\label{section:single-label-classification}
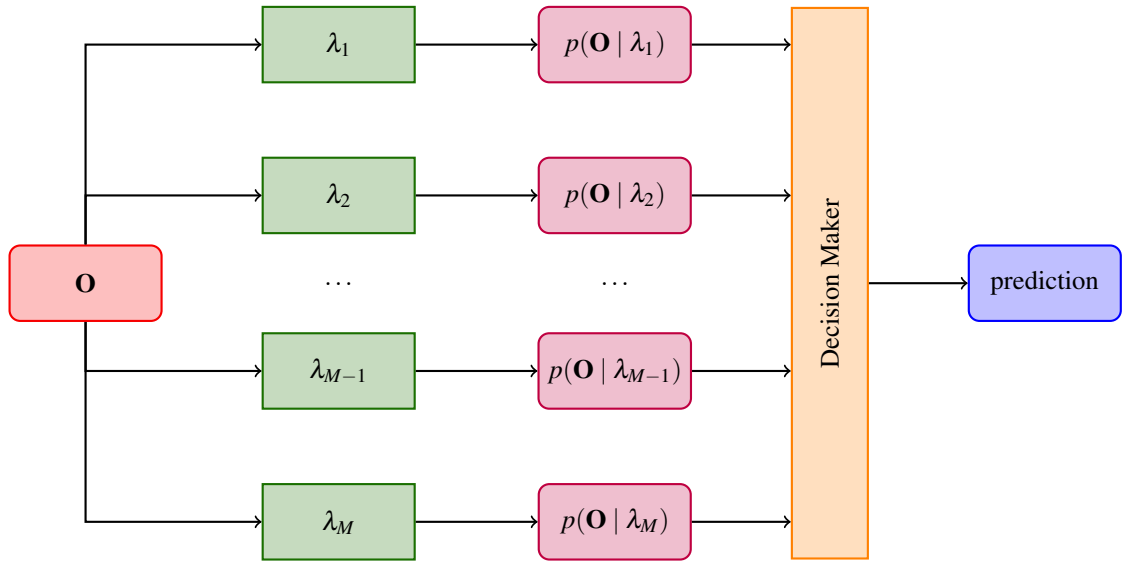
\begin{figure}[h]
    \centering
    \begin{tikzpicture}[->, node distance=2cm, thick]
      \tikzstyle{motion} = [rectangle, rounded corners, minimum width=2cm, minimum height=1cm,text centered, draw=red!100, fill=red!25]
      \tikzstyle{hmm} = [rectangle, minimum width=2cm, minimum height=1cm, text centered, draw=green!100, fill=green!25]
      \tikzstyle{ll} = [rectangle, rounded corners, minimum width=2cm, minimum height=1cm,text centered, draw=purple!100, fill=purple!25]
      \tikzstyle{dm} = [rectangle, minimum width=1cm, minimum height=7.3cm, text centered, draw=orange!100, fill=orange!25]
      \tikzstyle{pred} = [rectangle, rounded corners, minimum width=2cm, minimum height=1cm,text centered, draw=blue!100, fill=blue!25]
      \node (motion) [motion] {$\vec{O}$};

      \node (hmm_dot) [right=2cm of motion] {$\ldots$};
      \node (hmm_2) [hmm, above=0.5cm of hmm_dot] {$\lambda_2$};
      \node (hmm_1) [hmm, above of=hmm_2] {$\lambda_1$};
      \node (hmm_M1) [hmm, below=0.5cm of hmm_dot] {$\lambda_{M-1}$};
      \node (hmm_M) [hmm, below of=hmm_M1] {$\lambda_M$};

      \node (ll_dot) [right=3cm of hmm_dot] {$\ldots$};
      \node (ll_2) [ll, above=0.5cm of ll_dot] {$p(\vec{O} \mid \lambda_2)$};
      \node (ll_1) [ll, above of=ll_2] {$p(\vec{O} \mid \lambda_1)$};
      \node (ll_M1) [ll, below=0.5cm of ll_dot] {$p(\vec{O} \mid \lambda_{M-1})$};
      \node (ll_M) [ll, below of=ll_M1] {$p(\vec{O} \mid \lambda_M)$};

      \node (dm) [dm, right=2cm of ll_dot] {\rotatebox{90}{Decision Maker}};

      \node (pred) [pred, right=1.3cm of dm] {prediction};

      \draw (motion) |- (hmm_1);
      \draw (motion) |- (hmm_2);
      \draw (motion) |- (hmm_M1);
      \draw (motion) |- (hmm_M);

      \draw (hmm_1) -- (ll_1);
      \draw (hmm_2) -- (ll_2);
      \draw (hmm_M1) -- (ll_M1);
      \draw (hmm_M) -- (ll_M);

      \draw (ll_1.east) -- (dm.west|-ll_1.center);
      \draw (ll_2.east) -- (dm.west|-ll_2.center);
      \draw (ll_M1.east) -- (dm.west|-ll_M1.center);
      \draw (ll_M.east) -- (dm.west|-ll_M.center);

      \draw (dm) -- (pred);

    \end{tikzpicture}
    \caption{Overview of the classification process. For an unknown observation sequence $\vec{O}$ (in {\color{red}red}), the likelihood
    under each HMM (in {\color{green}green}) is computed. Each HMM corresponds to a single class. The likelihoods (in {\color{purple}purple})
    are then fed into a decision maker (in {\color{orange}orange}) that computes the prediction (in {\color{blue}blue}).}
    \label{fig:overview-classification}
\end{figure}

In the previous chapter motion recognition was discussed. This discussion is now extended to a
\emph{classification problem}. In a classification problem an unknown observation sequence, in this case a motion,
must be assigned to a finite set of classes $\mathcal{L}$. For example, assume that the set of classes
is \emph{run}, \emph{jump} and \emph{kick}.
The goal is then to find the correct \emph{label} $y \in \mathcal{L} = \{\text{run}, \text{jump}, \text{kick}\}$ for an
unknown motion $\vec{O}$ by making a prediction $p$. The classification is correct if $y = p$. Figure~\ref{fig:overview-classification}
provides an overview of the classification process. This process is used throughout this work, not just
for single-label classification.

If only a single label is assigned to a motion, the classification is straightforward. Instead
of using words, $y$ is encoded by natural numbers where each number corresponds to a class:
\begin{equation}
	y \in \{1, \ldots, M\},
\end{equation}
where $M$ is the number of classes that need to be recognized. The mapping between word and class can
be chosen arbitrarily as long as it is bijective. In the above example, a possible mapping would be
$\text{run} \rightarrow 1$, $\text{jump} \rightarrow 2$ and $\text{kick} \rightarrow 3$.

Additionally, assume that $M$ HMMs have been trained per class: $\lambda_1, \ldots, \lambda_M$.
The first HMM was only trained with motions of class $1$, the second one only with motions of class $2$ and so on.
In contrast to the learning algorithm described in the previous chapter, the labels of the training data
are now used to select the appropriate HMM during training. While the training of the HMM itself is still unsupervised,
the selection happens in supervised fashion.

After training, the prediction of an unknown motion $\vec{O}$ must be computed. To do so, the likelihoods
of the motion under each individual HMM are evaluated. The last step is then performed by
what this work refers to as a \emph{decision maker}. The job of a decision maker is to find a mapping from the likelihoods of the HMMs
to the prediction $p$. In case of the single-label classification problem, the prediction is done by a
very simple decision maker that chooses the label that corresponds to the HMM with the maximum likelihood:
\begin{equation}
	p = \argmax_{m \in \{1,\ldots,M\}}{p(\vec{O} \mid \lambda_m)}.
	\label{eq:single-classification}
\end{equation}

\section{Multi-Label Classification}
\label{section:multi-label-classification}
Let us now extend the classification problem to a problem where multiple labels can be assigned to a 
motion. This can be motivated by recalling the Motion Description Tree (see chapter~\ref{database}):
A motion is specified by the leave nodes of the tree, e.g.
\begin{inparaenum}[(1)]
    \item \emph{locomotion}~$\rightarrow$~\emph{bipedal}~$\rightarrow$~\emph{walk},
	\item \emph{speed}~$\rightarrow$~\emph{fast},
	\item \emph{direction}~$\rightarrow$~\emph{left},
	\item \emph{perturbation}~$\rightarrow$~\emph{result}~$\rightarrow$~\emph{failing}, and
	\item \emph{perturbation}~$\rightarrow$~\emph{source}~$\rightarrow$~\emph{passive}.
\end{inparaenum}
It should be obvious from this example that most motions cannot be adequately described by a single class.
However, if a motion can have multiple labels, the simple classification described by equation~\ref{eq:single-classification}
does not work anymore since it only computes the single label with the maximum likelihood.

To work around this problem, two different approaches can be identified: problem transformation using
the power set method or the use of more advanced decision makers. In the latter case, the decision maker
needs to truly handle multi-label classification. This can either be done by using the \emph{binary relevance method}
or \emph{algorithm modification}~\cite{read:2011classifier}.

\subsection{Power Set Method}
\label{section:classification-power-set-method}
A first attempt to solve this problem is to simply treat each possible combination of
labels as a single class. More formally, if $\mathcal{L}$ is the set of all labels,
compute the power set as the substitute label set:
\begin{equation}
	\hat{\mathcal{L}} = \mathcal{P}(\mathcal{L}) = \{\mathcal{U} \mid \mathcal{U} \subseteq \mathcal{L}\} \setminus \emptyset.
\end{equation}
If each element in $\hat{\mathcal{L}}$ is then mapped to a natural number, the multi-label problem
has been transformed into a single-label problem. This means that equation~\ref{eq:single-classification}
can be used to predict the substitute label with the maximum likelihood. Since the substitute label
represents multiple labels internally, reversing the transformation after the classification step solves the multi-label problem
using the already discussed approach for single-label classification. In the literature,
this idea is usually referred to as the \emph{power-set method}~\cite{boutell:2004learning, read:2011classifier}.

The power set method, however, has several downsides. Firstly, the number
of possible label combinations is $2^M-1$, where $M$ is the number of classes. Recall that for each class a Hidden
Markov Model must be trained. However, having sufficient training data now becomes a problem. If,
for example, $200$ walking motions exist but of those $200$ motions only a single a has the labels \emph{walk}
and \emph{fast} whereas all others have the labels \emph{walk} and \emph{slow}, the HMM for the fast walking
motion would only be trained on a single training sample.
This is inefficient since the HMM would benefit from the samples of the similar slow walking motion.
Secondly, to perform classification, the likelihood under each model must be calculated.
It can be easily seen that, for the worst case, this requires $\mathcal{O}(2^M)$ computations of the loglikelihood
for each unknown motion,
which makes this approach infeasible for even moderately large $M$.

\subsection{Binary Relevance Method}
\label{section:binary-relevance-method}
Instead of treating all possible combinations of labels as a single label, an approach where a motion can truly
have multiple labels is interesting. One way to achieve this is to encode the label as a binary
vector:
\begin{equation}
	\vec{y} \in \{0, 1\}^M,
\end{equation}
where $y_m$ is set to $1$ if the $m$-th label is active and set to $0$ otherwise.
To give an explanatory example, assume that the classes \emph{walk}, \emph{fast} and \emph{slow} exist.
The label of a fast walking motion would then be encoded as $\vec{y}=(1,1,0)$ whereas motion
where the subject walks slowly would would be encoded as $\vec{y}=(1,0,1)$.

The training then works as follows:
$M$ HMMs are initialized, one for each class. Given a training sequence $\vec{O}$, use it to train
all HMMs which correspond to the set classes in the sequence's label $\vec{y}$. If, for example,
$\vec{y}=(1,0,1)$, train $\lambda_1$ and $\lambda_3$ on the same sequence. An advantage over the previously described method
is that only $M$ instead of $2^M-1$ HMMs (worst case) must be kept in memory. Additionally, the utilization of the available training
data is vastly improved. In the above example, the HMM that corresponds to the \emph{walk} class now
benefits from both, the \emph{slow} and \emph{fast} walks. In the previously described approach, 
both would only be considered in isolation. However, this can also become a problem
if classes are too generic. Take for example motion of a throw performed with the left hand and 
a kick with the left foot. If both would be associated with the same class \emph{left}, it is unlikely
that common patterns can be learned properly. Instead, \emph{hand-left} and \emph{foot-left} can be used
to distinguish the two, which is a more reasonable choice in this case.

However, it is unclear how the decision maker can be realized in this case. The previous approach
where the class with the maximum likelihood was selected cannot be used anymore. One way to perform classification in this case is to
use some fixed value as the \emph{decision boundary}. Let $\bar{x}_m$ be such a decision boundary for the $m$-th label.
The multi-label classification is then achieved by computing all likelihoods of the unknown motion $\vec{O}$ under each 
HMM and selecting all that are equal to or greater than the respective decision boundary:
\begin{align}
	p_m = \begin{cases}
    	1, & \text{if } p(\vec{O} \mid \lambda_m) \geq \bar{x}_m \\
    	0, & \text{otherwise}\end{cases}
    \qquad m = 1,\ldots,M,
\end{align}
where $y_m$ denotes the $m$-th element of the prediction $\vec{p}$. This approach is
promising, but unfortunately it all depends on a good choices of $\bar{x}_m$.

Instead of trying to find good values for all $\bar{x}_m$ by hand, it would be preferable if the decision boundaries could be determined automatically.
Since the approach described in this work is supervised, the association between the likelihoods and the
corresponding class is actually known for the training dataset. It seems like a good idea to use this knowledge to learn a decision
boundary from the training data instead of finding it by hand. This has the advantage that the model easily adapts to the current
dataset at hand without the need to fine-tune the decision boundary manually. The problem can indeed be seen
as \emph{binary classification}: For each class, the features for the binary classifier are the likelihoods which
should be classified into two half-spaces that correspond to the labels $0$ and $1$ respectively. Since in this case, the 
feature space is $1$-dimensional (just a scalar likelihood under one specific model), the hyperplane that separates the two
half-spaces is a simple point and corresponds to the decision boundary $\bar{x}_m$ described previously.
This means that any binary classifier like \emph{Logistic
Regression}~\cite{jordan:2002discriminative} or \emph{Support Vector Machines}~\cite{hearst:1998support}
can be used to learn a mapping from the likelihoods to $y_m$ for each class.

Instead of only considering the $m$-th likelihood, in a slightly more advanced version of this basic idea
the binary classifier considers all likelihoods for its binary decision. This approach is commonly referred to as the \emph{binary relevance method}~\cite{read:2011classifier}.
More concretely, multi-label motion classification is realized with this method as follows:

\begin{minipage}{\textwidth} 
\begin{enumerate}
	\item Given the training data $\vec{O}^{(1)}, \ldots, \vec{O}^{(N)}$ and respective labels $\vec{y}^{(1)}, \ldots, \vec{y}^{(M)}$, calculate the likelihoods for each motion under each model:
	\begin{equation}
		\label{eq:likelihood-mat}
		\vec{x}^{(n)} = (p(\vec{O}^{(n)} \mid \lambda_1), p(\vec{O}^{(n)} \mid \lambda_2), \ldots, p(\vec{O}^{(n)} \mid \lambda_M)),
		\qquad n \in \{1,\ldots,N\}.
	\end{equation}
	\item Train $M$ binary classifiers. For each classifier, the features are $\vec{x}^{(1)}, \ldots, \vec{x}^{(N)}$. However, the labels are varied across classifiers, with the $m$-th classifier using labels $y^{(1)}_m, \ldots, y^{(N)}_m$.
	\item Given an unknown motion $\vec{O}$, calculate the likelihoods $\vec{x}$ under each of the $M$ models analogously to
	equation~\ref{eq:likelihood-mat}. Afterwards, perform $M$ predictions using the previously trained classifiers to
	obtain the complete multi-label prediction:
	\begin{equation}
		\vec{p} = (h^{(1)}(\vec{x}), \ldots, h^{(M)}(\vec{x})),
	\end{equation}
	where $h^{(m)}(\vec{x}) \in \{0,1\}$ denotes a function that computes the prediction of the $m$-th classifier.
\end{enumerate}
\end{minipage}

In this case, the decision maker uses multiple binary classifiers internally to learn a mapping from
the likelihoods of the HMMs to the multi-label prediction using the binary relevance method. It is important
to stress again that the ability to use any binary classifier for multi-label predictions is a big
advantage of this method since a wide and well-understood variety of such classifiers exist.

However, this approach also has a downside: Each class is considered
in isolation. This means that information is lost since the correlation between classes in the label
vector $\vec{y}$ carries information as well. More concretely, it is likely that certain label combinations correspond
to certain patterns in the likelihoods. Such patterns cannot be detected by the approach
described above, since each classifier only ``sees'' a single class~\cite{read:2011classifier}.

\subsection{Modified Algorithms}
\label{section:modified-algos}
Some learning algorithms have been modified to support multi-label classification ``out of the box''.
In such a case, the multi-label classification is straightforward: Similarly to the previous approaches,
the likelihoods under each model are calculated for each of the $N$ training motions and the likelihood
vectors $\vec{x}^{(1)},\ldots,\vec{x}^{(N)}$ are obtained (compare equation~\ref{eq:likelihood-mat}).
In contrast to the binary relevance method, the classifier can now be trained on the entire label
vector $\vec{y}$ instead of training individual classifiers on the individual classes. This potentially
allows the learning algorithm to find patterns between the likelihoods and the classes since the correlation
between classes can now also be considered. In this case the decision maker is thus simply a classifier
that is capable of learning multi-label classification.

Multiple algorithms exist that have been adopted to the multi-label problem. \emph{Decision Trees}
have been extended to work in multi-label classification problems~\cite{vens:2008decision}. With this
extension, \emph{Random Forests}~\cite{breiman:2001random} can also be used to perform multi-label
classification since a Random Forest is an \emph{ensemble classifier} that uses multiple Decision Trees
internally. While this work focuses on the mentioned algorithms, it should be noted that other algorithms have been modified
as well to support multi-label predictions, e.g. \emph{AdaBoost}~\cite{schapire:2000boostexter} or \emph{$k$-Nearest Neighbors}~\cite{zhang:2007ml}.

\chapter{Evaluation}
\label{chapter:evaluation}
The evaluation of the previously discussed concepts is challenging. This is due to the fact
that a huge number of system parameters can be identified: A set of features must be selected,
the hyperparameters of the Hidden Markov Models must be chosen, a comparison between FHMMs and
HMMs is necessary and a variety of different classifiers that perform the mapping from likelihoods to final
labels are available which also potentially have many hyperparameters that need fine-tuning. It should
be clear that evaluating all those DoFs simultaneously is impossible due to the large number of
combinations. Instead, the problem is split into smaller problems that are evaluated and optimized
individually. Only a small subset of the best parameter choices is then used in the next problem, and so on.

The structure of this chapter is based on this idea: To get started, a brief overview of the tools used and
developed for this work is given. The next section describes the used evaluation dataset. For the first evaluation step, feature selection is performed to find
a ``good'' subset of the wide variety of available features (see chapter~\ref{chapter:features}).
Next, Hidden Markov Models and their set of hyperparameters are evaluated (see chapter~\ref{section:motion-recognition}). The discussion is continued
by a comparison between Hidden Markov Models and Factorial Hidden Markov Models (see chapter~\ref{section:recognition-fhmm}).
Different approaches to predict the labels are evaluated (see chapter \ref{section:multi-label-classification}).
Finally, an end-to-end evaluation of two classifier systems is conducted using the results from the previous sections.

\section{Tools}
A toolkit was developed to help to evaluate this work. The toolkit is mostly written in Python with some
modules using C and C++ either for speed or for interoperability. Additionally, the excellent \emph{scikit-learn}
framework~\cite{pedregosa:2011scikit} was used throughout the toolkit. The toolkit
is used via multiple command-line scripts. This allows to easily automate long-running tasks and handles
the wide variety of different actions and parameters efficiently. 
Areas of responsibility are split into 3 main modules which are each briefly covered in the remainder of this section.

\subsection{\texttt{dataset} Module}
The \texttt{dataset} module provides classes and functions to load motion from different file formats
(MMM and C3D files). Furthermore, the derived features described in chapter~\ref{section:features-derived} can be computed as well.
The functionality of the \emph{Simox}\footnote{\url{http://simox.sourceforge.net}} and
\emph{MMMTools}\footnote{\url{https://gitlab.com/mastermotormap/mmmtools}} libraries are used to aid in the computation of more advanced
features like the angular momentum. \emph{SWIG}\footnote{\url{http://swig.org}} is used to bridge
between Python and the C++ API of the two mentioned libraries\footnote{\url{https://gitlab.com/cmandery/pySimox}}\footnote{\url{https://gitlab.com/cmandery/pyMMM}}. The module also provides functionality
to normalize features to obtain rotation and translation invariant representations. Features can also be smoothed
by applying a moving average filter and transformed to have similar scales across features. In short,
the \texttt{dataset} module implements everything that was discussed in chapter~\ref{chapter:features}.

Another important responsibility of the \texttt{dataset} module is the handling of datasets. A dataset
is defined by a simple \emph{manifest}. The manifest is realized as a JSON file that contains
an array of folders. Each folder can be label with multiple classes to perform supervised learning.
A dataset can be loaded from such a JSON manifest file. Multiple file formats (MMM and C3D) can be merged
into a combined dataset. A dataset can be checked for inconsistencies (e.g. missing files), a report
can be generated (e.g. the number of samples and the labels present in the dataset), selected
features can be plotted to visualize the data and the entire dataset can be exported into a single file
(different file formats are supported). Especially the export functionality is important since the
computation of the derived features and the normalization is computationally expensive. All evaluation
steps therefore load such an exported dataset instead of computing the dataset from scratch every time.

Lastly, the \texttt{dataset} module also provides functionality that helps with splitting the dataset
into two separate (and disjunct) datasets: one for training, one for testing. This is done using
\emph{stratified $k$-fold} for multi-label data as proposed by Sechidis~et~al.~\cite{sechidis:2011stratification}. The basic idea of a $k$-fold
evaluation is to split the dataset into $k$ disjunct subsets or \emph{folds}. $k$ rounds can then be performed by
using $k-1$ folds for training and the remaining fold for testing. In a stratified $k$-fold, each
fold attempts to have the same distribution of data across classes as the entire dataset.

\subsection{\texttt{hmm} Module}
The \texttt{hmm} module implements everything necessary to train and evaluate Hidden Markov Models.
The basic HMM functionality is provided by the \emph{hmmlearn} library\footnote{\url{https://github.com/hmmlearn/hmmlearn}}.
Other libraries were considered as well, namely \emph{pomegranate}\footnote{\url{https://github.com/jmschrei/pomegranate}} and
\emph{GHMM}\footnote{\url{http://ghmm.org}}. However, hmmlearn proved to be the most reliable and fastest
library.

For this evaluation, a fork\footnote{\url{https://github.com/matthiasplappert/hmmlearn}} of the hmmlearn library
was created and used. It fixes some issues and, more importantly, implements functionality to use
and train Factorial Hidden Markov Models. The implementation allows to compute exact likelihoods
as described by Ghahramani~et~al.~\cite{Ghahramani:1997id}. Additionally, it supports the sequential
training algorithm developed by Kulić~et~al.~\cite{Kulic:2008ib} (see chapter~\ref{section:recognition-fhmm})

The module also provides functionality to combine the individual HMMs into an ensemble of HMMs.
It allows training of the individual HMMs on labeled training data as well as computing the loglikelihoods
of unknown motions under each model (see chapter~\ref{section:single-label-classification} and \ref{section:multi-label-classification}). The training
and evaluation of likelihoods is performed in parallel for each model to maximize performance.

Lastly, the \texttt{hmm} module provides utility functions to compute the initial transition matrix
and start probabilities using different topologies (see chapter~\ref{section:topologies}). Functions to estimate the initial means and covariance matrices
using different strategies and constraints as described in chapter~\ref{section:parameter-initialization} are included as well.

\subsection{\texttt{misc} Module}
The \texttt{misc} module is a collection of smaller components. The most important one are the
\emph{decision makers}. A decision maker is a classifier that performs the mapping from the likelihoods
as calculated by the HMM ensemble to the binary label vector. As described in chapter~\ref{section:multi-label-classification},
different approaches exist. The \texttt{misc} module implements a decision maker that always picks
the maximum likelihood (chapter~\ref{section:single-label-classification}) and one that uses a fixed decision boundary
(chapter~\ref{section:binary-relevance-method}). Additionally, decision makers
that use Logistic Regression and Support Vector Machines are implemented using the binary relevance method (chapter~\ref{section:binary-relevance-method}).
Decision Trees and Random Forests are used with adapted learning algorithms to handle the multi-label problem directly
(chapter~\ref{section:modified-algos}). The decision makers make especially heavy use of the \emph{scikit-learn} framework.

\section{Dataset}
\label{section:dataset}
Each motion in the dataset was recorded using the motion capture system described in chapter~\ref{motion-capture}. The raw positions of the
motion markers were stored in a C3D file. Each motion was recorded using the KIT reference marker set consistent of $56$
markers placed at well-defined anatomical locations and a sampling rate
of \SI{100}{\Hz}. The C3D files are converted to the MMM reference model using $40$ joints (each with a single DoF) and a non-linear optimization
algorithm (see chapter~\ref{mmm}). The joint angles as well as the root position and root rotation were stored in an XML file.
The C3D and XML files are used to extract and compute all features described in chapter~\ref{chapter:features}.
For all features, the normalized or unnormalized form can be used. During normalization, the root rotation
of the first frame was constrained such that the roll and pitch angles were set to zero. This is because
a small error in either of the two angles can cause an improperly rotated coordinate system. The features can also be smoothed using a moving average filter with $W=3$
(compare chapter~\ref{section:feature-smoothing}). Table~\ref{table:features} lists all $29$ available features without differentiating
between normalized/unnormalized and smoothed/not smoothed.

\begin{table}[h]
  \tvspace{1.07}
	\begin{center}
    	\begin{tabular}{@{}rlp{7cm}@{}} \toprule
    	\textbf{Dimension} 		   & \textbf{Feature Name} & \textbf{Description} \\ \midrule
    	40 & joint\_pos	  		   				& angles of all $40$ joints \\ 
  		40 & joint\_vel	    	  				    & velocities of each joint \\ 
  		1 & joint\_vel\_norm 	   				    & Euclidean norm of the joint velocities \\ 
  		40 & joint\_acc 			   			    & acceleration of each joint \\ 
  		1 & joint\_acc\_norm 	   				    & combined Euclidean norm of the joint accelerations \\ 
  		3 & root\_pos 			    					& root position of the subject in Cartesian space \\ 
  		3 & root\_vel 			    					& directed root velocity \\ 
  		1 & root\_vel\_norm 	 					& Euclidean norm of the root velocity \\ 
  		3 & root\_acc 			   					& directed root acceleration \\ 
  		1 & root\_acc\_norm 	 					& Euclidean norm of the directed root acceleration \\ 
  		3 & root\_rot 			   					& pitch, roll and yawn angles of the subject's root \\ 
  		1 & root\_rot\_norm 	 					& Euclidean norm of the root rotation \\ 
  		12 & extremities\_pos  				& position of the hands and feet in Cartesian space \\ 
  		12 & extremities\_vel	 				& directed velocities of the hands and feet \\ 
  		4 & extremities\_vel\_norm  					& Euclidean norm of the directed extremity velocities per hand/foot \\ 
  		12 & extremities\_acc		    				& directed accelerations of the hands and feet \\ 
  		4 & extremities\_acc\_norm     				& Euclidean norm of the directed extremity velocities per hand/foot \\ 
  		3 & com\_pos 			      				& position of the center of mass in Cartesian space \\ 
  		3 & com\_vel 			      				& directed velocity of the center of mass \\ 
  		1 & com\_vel\_norm 	   				& Euclidean norm of the CoM's velocity \\ 
  		3 & com\_acc 			  	 				& directed acceleration of the center of mass \\ 
  		1 & com\_acc\_norm 		      				& Euclidean norm of the CoM's acceleration \\ 
  		3 & angular\_momentum 	    				& whole-body angular momentum in $x$, $y$ and $z$ direction \\ 
  		1 & angular\_momentum\_norm   				& Euclidean norm of the whole-body angular momentum \\ 
  		168 & marker\_pos 			    				& position of all $56$ markers in Cartesian space \\ 
  		168 & marker\_vel 			    				& directed velocities of all markers \\ 
		  1 & marker\_vel\_norm 	      				& Euclidean norm of all markers' velocities \\ 
  		168 & marker\_acc 			    				& directed acceleration of all markers \\ 
  		1 & marker\_acc\_norm 	     				& Euclidean norm of all markers' accelerations \\ \bottomrule
    	\end{tabular}
    \end{center}
    \caption{List of all $29$ available features with a total of $702$ dimensions.}
    \label{table:features}
\end{table}

A total of $454$ motions were selected from the motion database. $10$ different human subjects performed
the motions. Of those $10$ subjects, $3$ are female and $7$ are male. Each motion was performed by a single
human subject. If a motion involves an object (e.g. playing the guitar), the object was not physically present
but instead imagined by the subject. The only exception to this is a set of upwards walking motions where
actual stairs were used since it is quite hard to climb imaginary ones. Each recording in the dataset
contains only a single motion. Additionally, each subject begins the motion in an upward-standing pose similar to the zero-pose of the MMM model
and finishes the motion in the same pose. For periodic motions such as walking, stirring or dancing,
a fixed number of repetitions were selected across the dataset. The motions in the dataset were manually labeled
with $49$ different labels. Table~\ref{table:labels} lists all labels and the number of samples in the dataset that are assigned to each label.
Since motions can have multiple labels, table~\ref{table:label-combinations} lists all label combinations
that are present in the evaluation dataset.
\begin{table}[h]
  \tvspace{1.11}
  \thspace{0.5cm}
	\begin{center}
    	\begin{tabular}{@{}rl rl@{}}  \toprule
      \multicolumn{2}{c}{\textbf{Motions}} & \multicolumn{2}{c}{\textbf{Attributes}} \\ \cmidrule(r){1-2} \cmidrule(l){3-4}
    	\textbf{Samples} & \textbf{Label} & \textbf{Samples} & \textbf{Label} \\ \midrule
    	180 & walk 		   & 31 & turn-right 				\\ 
    	35 & push-recovery & 28 & turn-left 				\\ 
    	41 & run 		   & 9 & speed-fast 				\\ 
    	28 & kick         & 10 & speed-normal 			\\ 
    	10 & throw 	  & 10 & speed-slow 				\\ 
    	10 & bow       &  84 & direction-forward 		\\ 
    	5 & squat 		 &  14 & direction-backward 		\\ 
    	10 & punch 	  & 15 & direction-left 			\\ 
    	10 & stomp     &   12 & direction-right 			\\ 
    	25 & jump       &  15 & direction-upward 		\\ 
    	11 & golf        & 35 & direction-circle 		\\ 
    	20 & tennis     &  27 & direction-slalom 		\\ 
    	15 & wave        & 10 & bend-right				\\ 
    	11 & play-guitar & 10 & bend-left 				\\ 
  		10 & play-violin & 18 & counter-clockwise 		\\ 
  		11 & stir        & 17 & clockwise 				\\ 
  		11 & wipe        & 23 & foot-right				\\ 
  		11 & dance       & 15 & foot-left  				\\ 
  		   &			  & 45 & hand-right 				\\ 
  		   &			  & 48 & hand-left 				\\ 
  		   &			  & 5 & hand-both 				\\ 
  		   &			  & 5 & deep 						\\ 
  		   &			  & 5 & slight 					\\ 
  		   &			  & 10 & high 					\\ 
  		   &			  & 10 & low 						\\ 
  		   &			  & 5 & putting 					\\ 
  		   &			  & 6 & drive 					\\ 
  		   &			  & 10 & smash 						\\ 
  		   &			  & 10 & forehand 					\\ 
  		   &			  & 6 & waltz 						\\ 
  		   &			  & 5 & chachacha 					\\ \bottomrule
  		\end{tabular}
  	\end{center}
  	\caption{List of all $49$ different labels used in the dataset.}
    \label{table:labels}
\end{table}

\begin{table}[h]
  \tvspace{1.2}
	\begin{center}
    	\begin{tabular}{@{}rl rl@{}} \toprule
      \textbf{Samples} & \textbf{Label Combination} & \textbf{Samples} & \textbf{Label Combination} \\ \cmidrule(r){1-2} \cmidrule(l){3-4}
    	10 & walk, direction-forward, speed-normal & 5 & stomp, foot-right  \\ 
 		10 & walk, direction-forward, speed-slow & 5 & stomp, foot-left  \\ 
 		9  & walk, direction-forward, speed-fast & 5 & jump, direction-forward  \\ 
    	31 & walk, turn-right & 5 & jump, direction-backward  \\ 
 		28 & walk, turn-left & 5 & jump, direction-right  \\ 
 		10 & walk, bend-left & 5 & jump, direction-upward  \\ 
 		10 & walk, bend-right & 5 & jump, direction-left  \\ 
 		18 & walk, direction-circle, counter-clockwise & 5 & golf, putting  \\ 
 		17 & walk, direction-circle, clockwise & 6 & golf, drive  \\ 
 		27 & walk, direction-slalom  & 5 & tennis, hand-right, smash  \\ 
 		10 & walk, direction-upward  & 5 & tennis, hand-right, forehand \\ 
 		9 & push-recovery, direction-forward & 5 & tennis, hand-left, smash  \\ 
 		9 & push-recovery, direction-backward & 5 & tennis, hand-left, forehand  \\ 
 		7 & push-recovery, direction-right & 5 & wave, hand-right  \\ 
 		10 & push-recovery, direction-left & 5 & wave, hand-left  \\ 
 		41 & run, direction-forward  & 5 & wave, hand-both  \\ 
 		8 & kick, foot-right  & 5 & play-guitar, hand-right  \\ 
 		5 & kick, foot-right, high & 6 & play-guitar, hand-left  \\ 
 		5 & kick, foot-right, low  & 5 & play-violin, hand-right  \\ 
 		5 & kick, foot-left, high  & 5 & play-violin, hand-left  \\ 
 		5 & kick, foot-left, low  & 5 & stir, hand-right  \\ 
 		5 & throw, hand-right  & 6 & stir, hand-left  \\ 
 		5 & throw, hand-left  & 5 & wipe, hand-right  \\ 
 		5 & bow, deep  & 6 & wipe, hand-left  \\ 
 		5 & bow, slight  & 6 & dance, waltz  \\ 
 		5 & squat  & 5 & dance, chachacha  \\ 
 		5 & punch, hand-right & &  \\ 
 		5 & punch, hand-left & & \\ \bottomrule
    	\end{tabular}
	\end{center}
	\caption{All $54$ label combinations used in the dataset.}
	\label{table:label-combinations}
\end{table}

\section{Feature Selection}
\label{section:feature-selection}
In feature selection, the goal is to find a subset of features that are relevant to the problem at hand.
The selection process is a crucial first step. This is due to the fact that any classifier can only compute
predictions from the features it receives. If those features do not contain the relevant features, the classifier
obviously cannot produce optimal results. Additionally, a reduction of the feature dimensionality has
several advantages: Firstly, the computational complexity is usually proportional to the dimensionality
of the feature space. Secondly, if the feature space has very high dimensionality but the dataset is
relatively small, overfitting becomes more likely. Lastly, reducing the feature
set can also bring interesting insights to better understand a problem at hand~\cite{guyon:2003introduction}.

Unfortunately, the optimal feature set cannot be computed directly by evaluating all possible subsets.
This is due to the fact that, for a feature set with $N$ different features, $2^N-1$ non-empty subsets exist.
As previously described, a set of $N=29$ different features can be identified in this work with
results in $2^{29}-1 = 536\,870\,911$ subsets. Evaluating each of these subsets is infeasible.
In the literature, three popular heuristics can be identified to solve this problem:
the \emph{filter method}, the \emph{wrapper method} and the \emph{embedded method}.
The filter method selects features by ranking them with correlation coefficients. The wrapper method
evaluates the usefulness of a subset using a classifier. The embedded method works similarly to the
wrapper method in the sense that it uses a predictor. However, while the wrapper methods relies
on an external measure to assess the quality of the feature set, the embedded method relies on feature
selection that is inherent to the learning algorithms. To give a concrete example, Support Vector Machines
with L1 regularization can be used to perform feature selection using the embedded method. This is possible
because L1 regularization yields sparse weights. The selection can then be done
by simply picking all features that correspond to non-zero weights without the need of an externally defined measure~\cite{guyon:2003introduction}.

In this work, feature selection is performed using the \emph{wrapper method} with \emph{backwards elimination}.
In backwards elimination, one starts with the full feature set. Each feature is temporarily excluded from the set once and a model
is trained on each of the subsets. For each subset, the model is evaluated on the test set and a measure
is computed. The feature with the \emph{least} effect on the measure is then removed from the feature set. This process
is repeated until the feature set is empty (or until a stopping criterion is reached).
The feature selection was performed on the dataset described in section~\ref{section:dataset}. Of the $29$ available features, the features
\emph{marker\_pos}, \emph{marker\_vel} and \emph{marker\_acc} were initially excluded. This is due
to the very high dimensionality of the three mentioned features ($168$ dimensions each) that resulted in
numerical problems. The three excluded features will be revisited later. Additionally, to avoid too much complexity, all features
under consideration were normalized, smoothed (using the previously described moving average with $W=3$) and scaled to be
in range $[-1, 1]$. The reasoning behind
this decision is that it seems very unlikely that unnormalized features perform well since they are,
as previously discussed, neither invariant to translation nor rotation. Another concern was that
the unnormalized features can cause overfitting. The smoothed features were
selected as the default for similar reasons since it seems unlikely that an error-prone signal with strong jitter
performs better than a slightly smoothed representation thereof. Scaling is important to ensure a good
initialization when using $k$-means clustering. Since the dataset contains $49$ different
classes, $49$ HMM models with $5$ states each and the \emph{left-to-right} topology with $\Delta = 1$ were
used. Each model was trained using the Baum-Welch algorithm for $10$ iterations. The transition matrix
and start probabilities were initialized uniformly whereas $k$-means clustering was used to initialize the
means and covariance matrices. The covariance matrix was further constrained to be diagonal and the diagonal
elements were constrained to be larger than $0.0001$ to avoid numerical instabilities. The entire dataset was
initially shuffled to break correlations between nearby training samples. A stratified $3$-fold was used to 
train and evaluate the HMMs three times per feature subset. In each round, the likelihoods under each model were
calculated for the test split. The likelihoods were additionally split into a set of positive and negative likelihoods.
The positive set contains all likelihoods of the motions that should have been recognized by the respective model whereas the
negative set contains the likelihoods of the motions that should have been rejected. Since $49$ different HMMs are used this results in
$98$ sets of likelihoods: $49$ sets of positive instances and $49$ sets of negative instances. The mean and standard deviation
were then calculated per set over all three rounds, resulting in $98$ means of the likelihoods and their respective standard deviations.

A possible way to measure the results in each round is to compute the distance between the distribution of positive and negative likelihoods
on a per-class basis. The \emph{Wasserstein metric}~\cite{givens:1984class, wasserstein}
can be used to compute the distance between two Gaussian distributions:
\begin{equation}
  m_i = \sqrt{|\bar{\mu}_{pos_i} - \bar{\mu}_{neg_i}| + \left(\bar{\sigma}_{pos_i}^2 + \bar{\sigma}_{neg_i}^2 - 2 \sqrt{\bar{\sigma}_{pos_i}^2 \bar{\sigma}_{neg_i}^2}\right)},
\end{equation}
where $i$ denotes the $i$-th class, $\bar{\mu}_{pos_i}$ is the mean over all positive
and $\bar{\mu}_{neg_i}$ the mean over all negative likelihoods \emph{for the $i$-th class}. $\bar{\sigma}_{pos_i}$ and
$\bar{\sigma}_{neg_i}$ denote the respective standard deviations. It was further assumed that the positive and negative likelihood distributions are
Gaussian. This assumption is not formally proven here but seems likely to hold given the central limit theorem.

After computing $m_1, \ldots, m_{49}$,
the individual distance measures must be combined into a single measure. Since all
labels are equally important, the median is used here. This ensures that
the distance between positive and negative distribution is balanced across all $49$ classes. The total dimension of the features
should be considered as well. If, for example, roughly the same distance has been computed for a feature set with $100$ and
$10$ dimensions respectively, the feature set with only $10$ dimensions should be preferred. A simple way to achieve
this is to compute a measure \emph{per feature}. Based on these considerations, the following combined measure is devised:
\begin{equation}
  m = \frac{1}{D} \tilde{m},
  \label{eq:wasserstein-metric}
\end{equation}
where $D$ denotes current number of feature dimensions and $\tilde{m}$ is the median over individual
Wasserstein metrics $m_1, \ldots, m_{49}$. Other measures
were considered (e.g. \emph{AICc} and the \emph{Mahalanobis metric}) but proofed to not work as well
as the Wasserstein-based metric.

\begin{table}[h]
  \tvspace{1.12}
  \thspace{0.5cm}
  \begin{center}
      \begin{tabular}{@{}rrrl@{}} \toprule
      \textbf{Round} & \textbf{Score} & \textbf{Dimension} & \textbf{Deleted Feature} \\ \midrule
       1 &   5\,899  & 198 & ---                                          \\ 
       2 &   6\,615  & 158 & joint\_acc                       \\ 
       3 &   9\,009  & 118 & joint\_vel                       \\ 
       4 &  10\,686  &  78 & joint\_pos                       \\ 
       5 &  12\,602  &  66 & extremities\_acc                   \\ 
       6 &  15\,508  &  54 & extremities\_vel                   \\ 
       7 &  17\,176  &  53 & com\_vel\_norm                    \\ 
       8 &  18\,740  &  49 & extremities\_vel\_norm              \\ 
       9 &  20\,856  &  46 & angular\_momentum                \\ 
      10 &  23\,644  &  42 & extremities\_acc\_norm              \\ 
      11 &  26\,253  &  41 & root\_vel\_norm                   \\ 
      12 &  32\,415  &  38 & com\_vel                         \\ 
      13 &  35\,563  &  35 & com\_acc                         \\ 
      14 &  38\,544  &  32 & root\_acc                        \\ 
      15 &  40\,214  &  31 & joint\_acc\_norm                  \\ 
      16 &  42\,362  &  30 & joint\_vel\_norm                  \\ 
      17 &  44\,416  &  29 & com\_acc\_norm                    \\ 
      18 &  46\,866  &  28 & root\_acc\_norm                   \\ 
      19 &  48\,849  &  27 & marker\_acc\_norm                 \\ 
      20 &  50\,754  &  26 & angular\_momentum\_norm           \\ 
      21 &  52\,949  &  25 & marker\_vel\_norm                 \\ 
      22 &  54\,781  &  22 & com\_pos                         \\ 
      23 &  51\,928  &  19 & root\_pos                        \\ 
      24 &  54\,525  &  16 & root\_vel                        \\ 
      25 &  48\,524  &  13 & root\_rot                        \\ 
      26 & 127\,598  &   1 & extremities\_pos                   \\ \midrule
      \multicolumn{3}{c}{\textbf{Remaining Feature}}    & root\_rot\_norm                   \\ \bottomrule
      \end{tabular}
  \end{center}
  \caption{The results of the feature selection using backward elimination. The table lists all elimination
  rounds, the feature that was deleted in each round, the measure that was computed in each round without the deleted
  feature and the dimension of the combined features without the deleted feature.}
  \label{table:feature-selection-results}
\end{table}

Table~\ref{table:feature-selection-results} lists the results per elimination round. The best feature was found in the last round: \emph{root\_rot\_norm}. However, the individual
likelihoods reveal that the single feature does not capture enough information to recognize all motion
types. For example, the \emph{stir} motion had a mean loglikelihood of $-12\,046 \pm 43\,080$, which
clearly indicates that this motion cannot be recognized using only the normalized norm of the root rotation.
Some direction labels achieve similarly bad results. The label \emph{clockwise} and \emph{counter-clockwise}
for example achieved a loglikelihood score of $-311 \pm 67$ and $-275 \pm 103$ respectively. However, the
mean loglikelihoods for motions that should \emph{not} be recognized by the models are $401 \pm 701$
and $405 \pm 725$ respectively. This feature set was therefore discarded as an outlier.

The second best result was achieved in round $22$ with the following set of features: \emph{root\_pos},
\emph{root\_vel}, \emph{extremities\_pos}, \emph{root\_rot} and \emph{root\_rot\_norm}.
This is a more reasonable result. The velocity of the subject's root certainly is important
to detect the direction the subject moves in and to decide if the motion is of stationary or dynamic nature.
The position of the extremities proved to be an especially important feature since it survived all elimination
rounds until the last one. This also makes sense: A lot of motions in the dataset involve the subject's hands
or feet. Examples are stirring, waving, throwing a ball, playing the guitar or violin and kicks. The
position of the extremities presumably also help with identifying which hand or foot was used to perform
the motion. Including the root rotation and its norm makes sense as well: Since the root position
and the root rotation are both included, it should be possible to adequately determine the subject's position
and rotation in space from the feature set. This is obviously important information. An interesting
observation is that the joint angles are not that important to recognize motions. The joint angles
were already discarded in the fourth round. This is presumably due to the relatively low
information contents per feature. While some joints might carry important information, this
information is also redundantly present in other features like the position of the extremities. Since
other feature sets achieved similar scores, they will be revisited in the last
section during the end-to-end evaluation.

All things considered, the selected features seem like a good choice. This claim is backed by the data in table~\ref{table:feature-selection-likelihoods} (at
the very end of this section)
which lists the mean likelihoods and their standard deviation per label over the entire test dataset
for HMMs that were trained using the previously mentioned feature set from elimination round $22$.
The likelihoods of positive samples are consistently high with a reasonable standard deviation. Additionally, the models
only respond to the motions that they were trained to recognize. The only exception is the model
that represents walking motions, which seems likely to produce false positives. This is presumably due
to the fact that very different walking motions (e.g. walking and making a 90 degree turn vs. walking
forward) were used to train the same model. The high standard deviation of the negative
likelihoods can be explained by the fact that an extremely wide variety of very different motions were
all combined into a single score. In other words: The positive likelihoods were only computed for motions
of the same type whereas the negative likelihoods combines motion of all other types into a single score.

\begin{table}[h]
  \tvspace{1.3}
  \thspace{0.5cm}
  \begin{center}
      \begin{tabular}{@{}rrl@{}} \toprule
      \textbf{Score} & \textbf{Dimension} & \textbf{Additional Feature} \\ \midrule
      54\,781        & 22                 & ---                           \\
       4\,456        & 181                & marker\_pos        \\
       4\,459        & 181                & marker\_vel        \\
      NaN            & 181                & marker\_acc        \\ \bottomrule
      \end{tabular}
  \end{center}
  \caption{Comparison between the baseline feature set and the previously excluded features. Each of the previously excluded
  features is added once to the baseline set and the performance is evaluated.}
  \label{table:feature-selection-marker-results}
\end{table}

Recall that three features were excluded from the previous feature selection process: \emph{marker\_pos},
\emph{marker\_vel} and \emph{marker\_acc}. This was necessary because of numerical problems when including these features in the initial feature set during backwards elimination.
More concretely, the probabilities during inference would become so small that an underflow occurred even though
the HMM implementation used during the evaluation already uses logarithmic probabilities to counter the problem. However,
these features are now revisited. To do so, the best feature set as previously determined was used as a baseline.
Each of the three left out features was then added once. The achieved scores (using the same measure
and training procedure as before) are listed in table~\ref{table:feature-selection-marker-results}.

It is quite obvious from the data that the marker positions, velocities and accelerations do not
contribute anything to the feature set; instead the scores worsen significantly. Notice that the marker
accelerations even cause numerical instabilities due to vanishing probabilities, hence a score cannot be computed in this case.
The three temporarily excluded features are thus excluded permanently from further consideration.

\begin{table}[h]
  \small
  \tvspace{1.07}
  \thspace{0.5cm}
  \begin{center}
      \begin{tabular}{@{}lrr@{}} \toprule
      \textbf{Class}  & \textbf{Positive Loglikelihood}  & \textbf{Negative Loglikelihood} \\ \midrule
      walk                & $12\,858 \pm 4\,924$      &  $10\,095 \pm 16\,747$                       \\ 
      push-recovery       & $84\,073 \pm 26\,271$     &  $-830\,206 \pm 1\,314\,193$                       \\ 
      run                 & $13\,655 \pm 3\,937$      &  $-161\,784 \pm 344\,406$                       \\ 
      kick                & $24\,898 \pm 5\,235$      & $-207\,653 \pm 422\,938$                        \\ 
      throw               & $25\,122 \pm 6\,796$      & $-4\,982\,845 \pm 8\,584\,547$                        \\
      bow                 & $31\,251 \pm 4\,014$      & $-5\,946\,674 \pm 9\,198\,336$                        \\
      squat               & $32\,149 \pm 6\,640$      & $-2\,198\,971 \pm 3\,600\,540$                       \\ 
      punch               & $35\,073 \pm 11\,606$     & $-1\,828\,241 \pm 3\,125\,022$                        \\ 
      stomp               & $34\,982 \pm 3\,633$      & $-3\,409\,349 \pm 5\,592\,487$                        \\ 
      jump                & $24\,124 \pm 6\,256$      & $-293\,847 \pm 612\,549$                        \\ 
      golf                & $37\,559 \pm 8\,758$      & $-1\,880\,819 \pm 3\,414\,369$                        \\ 
      tennis              & $34\,408 \pm 3\,933$      & $-303\,574 \pm 551\,566$                        \\ 
      wave                & $58\,801 \pm 7\,035$      & $-14\,784\,535 \pm 21\,892\,345$                        \\ 
      play-guitar         & $49\,155 \pm 4\,275$      & $-3\,531\,256 \pm 6\,955\,093$                        \\ 
      play-violin         & $55\,510 \pm 3\,780$      & $-6\,878\,584 \pm 10\,443\,505$                        \\
      stir                & $28\,269 \pm 52\,856$     & $-5\,574\,138 \pm 8\,338\,055$                        \\
      wipe                & $48\,898 \pm 4\,734$      & $-8\,223\,779 \pm 13\,225\,863$                        \\
      dance               & $33\,981 \pm 5\,709$      & $-178\,554 \pm 259\,631$                        \\ 
      turn-right          & $18\,467 \pm 3\,158$      & $-58\,574 \pm 111\,112$                        \\ 
      turn-left           & $18\,594 \pm 7\,066$      & $-36\,449 \pm 69\,685$                        \\ 
      speed-fast          & $20\,673 \pm 3\,640$      & $-333\,766 \pm 531\,406$                        \\ 
      speed-normal        & $27\,002 \pm 8\,508$      & $-393\,349 \pm 618\,222$                        \\ 
      speed-slow          & $28\,689 \pm 2\,372$      & $-410\,014 \pm 559\,746$                        \\ 
      direction-forward   & $20\,472 \pm 17\,043$     & $-159\,039 \pm 328\,429$                        \\ 
      direction-backward  & $54\,408 \pm 24\,864$     & $-720\,136 \pm 1\,467\,702$                        \\ 
      direction-left      & $86\,904 \pm 52\,238$     & $-949\,005 \pm 1\,478\,147$                        \\ 
      direction-right     & $50\,982 \pm 24\,431$     & $-1\,271\,632 \pm 1\,879\,763$                        \\ 
      direction-upward    & $27\,477 \pm 4\,773$      &  $-194\,048 \pm 366\,744$                       \\ 
      direction-circle    & $17\,183 \pm 6\,479$      &  $-24\,067 \pm 75\,746$                       \\ 
      direction-slalom    & $23\,776 \pm 4\,992$      &  $-38\,560 \pm 106\,968$                       \\ 
      bend-right          & $28\,950 \pm 3\,030$      & $-191\,518 \pm 237\,383$                        \\ 
      bend-left           & $23\,457 \pm 4\,346$      & $-206\,328 \pm 262\,216$                        \\ 
      counter-clockwise   & $21\,281 \pm 5\,297$      & $-50\,763 \pm 115\,171$                        \\ 
      clockwise           & $20\,271 \pm 8\,250$      &   $-45\,160 \pm 83\,697$                        \\
      foot-right          & $25\,384 \pm 5\,914$      &  $-229\,056 \pm 472\,367$                        \\
      foot-left           & $31\,525 \pm 2\,765$      & $-1\,583\,832 \pm 2\,753\,615$                        \\ 
      hand-right          & $39\,043 \pm 11\,234$     &  $-457\,165 \pm 819\,820$                        \\ 
      hand-left           & $40\,720 \pm 11\,209$     &  $-380\,057 \pm 723\,547$                        \\ 
      hand-both           & $62\,305 \pm 4\,304$      & $-8\,376\,056 \pm 12\,155\,675$                        \\
      deep                & $29\,047 \pm 2\,347$      & $-5\,173\,123 \pm 7\,942\,443$                       \\ 
      slight              & $31\,561 \pm 6\,827$      & $-3\,272\,445 \pm 6\,239\,772$                        \\ 
      high                & $29\,758 \pm 4\,651$      & $-1\,173\,850 \pm 2\,088\,698$                        \\ 
      low                 & $27\,388 \pm 3\,102$      & $-1\,653\,420 \pm 2\,928\,772$                        \\ 
      putting             & $39\,531 \pm 4\,220$      & $-4\,056\,969 \pm 6\,825\,540$                        \\ 
      drive               & $36\,932 \pm 11\,987$     & $-1\,846\,573 \pm 3\,369\,606$                        \\ 
      smash               & $35\,313 \pm 4\,018$      &  $-328\,343 \pm 665\,057$                        \\ 
      forehand            & $31\,963 \pm 4\,084$      &  $-826\,177 \pm 1\,417\,883$                        \\ 
      waltz               & $36\,618 \pm 10\,136$     &  $-396\,325 \pm 634\,596$                        \\ 
      chachacha           & $29\,362 \pm 8\,173$      &  $-797\,359 \pm 1\,213\,353$                        \\ \bottomrule
      \end{tabular}
  \end{center}
  \caption{The mean positive and negative loglikelihoods under each of the $49$ classes for the feature set selected during round $22$.}
  \label{table:feature-selection-likelihoods}
\end{table}

\section{Hidden Markov Models}
\label{section:evaluation-hmm}
As discussed in chapter~\ref{section:motion-recognition}, Hidden Markov Models can be used to perform motion
recognition. However, HMMs have multiple hyperparameters that need to be selected, namely the number
of states and the topology. Additionally, different initialization strategies can be used. As mentioned
in chapter~\ref{section:parameter-initialization}, proper initialization is a crucial step before training.
Moreover, a comparison between the performance of HMMs and FHMMs makes sense. FHMMs have another
important hyperparameter, the number of chains.

Throughout this section, the experimental setup from the previous feature selection section was used again.
More concretely, the same dataset was used. To make the results comparable between this and the previous section,
the same permutation was used to shuffle the dataset. $3$-fold cross-validation was used with $49$ HMMs, one for each
class. Each HMM was trained for $10$ iterations. The best feature set from the previous section was used throughout
the following experiments: \emph{root\_pos}, \emph{root\_vel}, \emph{extremities\_pos}, \emph{root\_rot} and \emph{root\_rot\_norm}.
All features were normalized, smoothed and scaled. The same Wasserstein-based metric (equation~\ref{eq:wasserstein-metric})
from the previous chapter was used to score the results. The default parameters of the HMMs were as follows: $5$ states, \emph{left-to-right} topology
with $\Delta = 1$, uniform initialization of the transition \& start probabilities and $k$-means initialization
of the Gaussian emission distribution parameters with the covariance matrices constrained to be diagonal. Notice that these are only defaults. Each experiment
will vary either one or multiple of these parameters, which will be clearly stated in each subsection.

\subsection{Hyperparameters}
\label{section:evaluation-hmm-hyper}
The optimal number of states and topology are found using \emph{grid search}. In grid search, each possible
combination of hyperparameters is used to train a model on the training dataset which is then evaluated
on the test dataset. A measure is computed per combination and the best combination is selected. Such a search
is feasible in this case since only two different hyperparameters are evaluated.

In theory any natural number can be used for the number of states and countless different topologies are possible.
In practice however, it makes sense to limit the number of states $K$ to be between $5$ and $20$ states
when recognizing human motions~\cite{Kulic:2007clustering, Kulic:2008ib}. In this evaluation $K~\in~\{3,\ldots,20\}$,
resulting in $18$ different values for the number of states.
The topologies (chapter~\ref{section:topologies}) under consideration are the following:
\begin{itemize}
  \item \emph{fully-connected},
  \item \emph{left-to-right} without a $\Delta$ constraint,
  \item \emph{left-to-right} with $\Delta = 1$, and
  \item \emph{left-to-right} with $\Delta = 2$.
\end{itemize}
The left-to-right topologies proved to be especially well-suited for motion recognition~\cite{Kulic:2008ib}. Different
variations of this topology are evaluated by varying the $\Delta$ parameter. For completeness, the fully-connected topology is considered
as well.

\begin{figure}[h]
    \centering
    \includegraphics[width=\textwidth]{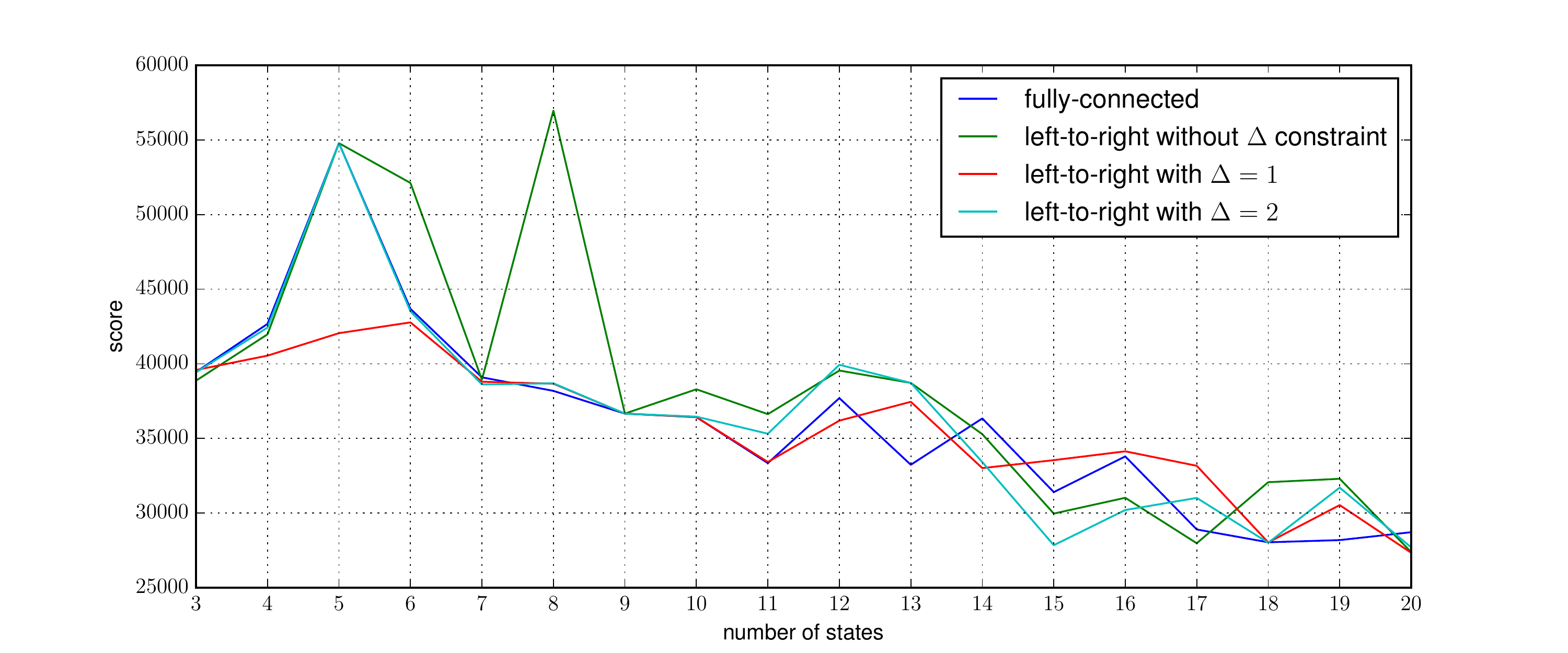} 
    \caption{The results of the hyperparameter grid search. The four different topologies are plotted individually. For each
    topology the number of states are plotted against the respectively achieved score.}
    \label{fig:results-hmm-hyperparams}
\end{figure}

Hence $18$ different values for the number of states and $4$ different topologies were under evaluation.
This results in a total of $18~\cdot~4~=~72$ combinations, making the grid search easily feasible.
The previously described experimental setup was used, with all parameters fixed and set to their default values except
for the number of states and the topology. The results are depicted in figure~\ref{fig:results-hmm-hyperparams}.

As can be seen from the results, it is indeed desirable to have a low number of states. Good choices
are between $5$ and $8$ states, depending on the topology. Another interesting observation is that
the \emph{left-to-right} topology with $\Delta = 1$ is the only topology that performed consistently
without large jumps. However, the best result was achieved by the \emph{left-to-right} topology without
a $\Delta$ constraint and with $8$ states. The result support the claim that a low number of states
is preferable for motion recognition and that the \emph{left-to-right} topology is a good choice when
dealing with motions. However, except for a few cases, the performance across topologies was similar.

\subsection{Parameter Initialization}
\label{section:evaluation-hmm-init}
Different strategies can be used to initialize the parameters of a Hidden Markov Model before starting
the training (see chapter~\ref{section:parameter-initialization}). To quickly recap, the transition
matrix and the start probabilities can either be initialized uniformly or by randomization. The implementation
of a uniform initialization is trivial. The randomization can be easily achieved as well: The transition
and start probabilities are initialized uniformly. Next, each element from the transition matrix is multiplied by a pseudo-random
number (in this case the random number was drawn uniformly from the interval $[0,1]$). Notice that for each
multiplication a new random number must be used. After randomization, the probabilities need to be normalized
such that they sum to $1$. This can be easily done by summing over each row of the transition matrix
and dividing each element in that row by the the respective sum. The randomization of the start probabilities vector works
analogously. This randomization automatically ensures that the desired topology is maintained.

The means and covariance matrices can either be randomized or estimated using the $k$-means clustering algorithm.
The randomization of the mean vectors is straightforward: $D$ pseudo-random numbers (in this case drawn uniformly from $[-1,1]$ to match
the feature scaling) are combined to form the mean vector ($D$ denotes the number of features). This is repeated for each state to obtain $K$ random mean vectors.
The randomization of the covariance matrix is a bit more complicated since a covariance matrix
is by definition symmetric and positive semi-definite. However, this can be easily achieved as well:
First, initialize a randomized matrix $\vec{R}$ of the desired dimension (that is $D \times D$).
A symmetric and positive semi-definite matrix is then be obtained as follows~\cite{psd}:
\begin{equation}
  \vecsym{\Sigma} = \vec{R} \vec{R}^T.
\end{equation}
Again, this is repeated $K$ times to obtain one covariance matrix per state.

In contrast, the initialization using $k$-means is done as follows: First $K$ clusters are found in the training
dataset using the $k$-means algorithm as implemented by \emph{scikit-learn} (the implementation uses \emph{Lloyd's algorithm}~\cite{kanungo:2002efficient}).
More concretely, all training observation sequences are ``stacked'' vertically. The clustering algorithm is then run 
on this stacked matrix. Also recall that
$K$ is the number of states, in this case $K=5$. Next, the mean vectors of each state are simply
set to their respective $D$-dimensional cluster centers. The computation of the covariance matrix is a
bit more complicated: For each state, each row of the stacked observation sequences is assigned to a cluster. This
is simply done by computing the Euclidean distance between each cluster center and the current sample
and selecting the cluster with the smallest distance. Next, a maximum likelihood covariance estimator
is used to estimate the $K$ covariance matrices.
In this case, the \texttt{EmpiricalCovariance} estimator as implemented by \emph{scikit-learn} was used.

Finally, the covariance matrices can be constrained to be diagonal. This is done during initialization
by simply setting every element to $0$ except for the ones on the diagonal. During training, the maximization
of the covariance is adopted to the diagonal case as described in~\cite{huang:2001spoken}. The \emph{hmmlearn}
library implements this approach.

To summarize, the following different initializations can be identified: The transition and start probabilities
can be randomized or initialized uniformly. The means and covariances of the emission distribution can be
randomized as well or estimated using the $k$-means clustering algorithm. The covariance matrices can be unconstrained (``full'')
or constrained to be diagonal. This results in $2 \cdot 2 \cdot 2 = 8$ different combinations. Similarly
to the previous subsection, grid search was used to evaluate each possible combination. The previously described experimental setup was used again,
with all parameters fixed and set to their default values except for the initialization strategies.
The results are given in table~\ref{table:results-parameter-initialization}.

\begin{table}[h]
  \tvspace{1.2}
  \thspace{0.3cm}
  \begin{center}
      \hspace*{-25pt}
      \begin{tabular}{@{}lllrrr@{}} \toprule
      & & & & \multicolumn{2}{c}{\textbf{Loglikelihood}} \\ \cmidrule(l){5-6}
      \textbf{Transition \& Start} & \textbf{Emission}  & \textbf{Covariance} & \textbf{Score} & \textbf{Positive} & \textbf{Negative} \\ \midrule
      randomized                   & randomized         & full                & NaN      & --- & ---                     \\ 
      randomized                   & randomized         & diagonal            &  10\,117 &  $29\,587 \pm 5\,495$ & $-102\,020 \pm 226\,759$         \\ 
      randomized                   & $k$-means            & full                & 641\,257 & $-24\,822 \pm 60\,461$ & $-9\,623\,115 \pm 14\,139\,634$       \\
      randomized                   & $k$-means            & diagonal            &  54\,782 &  $31\,268 \pm 5\,299$ & $-721\,413 \pm 1\,213\,389$        \\ 
      uniform                      & randomized         & full                & NaN      & ---  & ---                    \\ 
      uniform                      & randomized         & diagonal            &   9\,392 &  $30\,444 \pm 6\,038$ & $-130\,813 \pm 211\,034$         \\ 
      uniform                      & $k$-means            & full                & 658\,865 & $-24\,746 \pm 60\,202$ & $-9\,594\,632 \pm 14\,526\,834$      \\ 
      uniform                      & $k$-means            & diagonal            &  54\,781 &  $31\,242 \pm 5\,302$ & $-782\,554 \pm 1\,213\,368$ \\ \bottomrule
      \end{tabular}
    \end{center}
    \caption{The achieved scores and median loglikelihoods across all classes for each possible combination of initialization strategies.}
    \label{table:results-parameter-initialization}
\end{table}

A first conclusion that can be drawn from the results is that a randomizing the emission distribution parameters is not
a good initialization strategy. If the covariance matrices are unconstrained, the score cannot even be computed due to vanishing probabilities and resulting numerical underflows.
If the covariance matrices are constrained to be diagonal, a score can be computed. Notice however that the achieved
score is significantly worse than the scores achieved using $k$-means initialization. This nicely illustrates
that Baum-Welch does not necessary converge to a global maximum. Having a good initial estimate is therefore
indeed a crucial step, as previously claimed. Another interesting conclusion is that the initialization
of the transition and start probabilities is less important. Take for example the results of the fourth
and last row: The achieved scores and loglikelihoods are almost identical. Lastly, consider the difference
between full and constrained covariance matrices for the emission distribution. While the scores for
the full covariance matrices are higher, the mean loglikelihood scores plummet. Compared
to the previously achieved results (table~\ref{table:feature-selection-likelihoods}) using diagonally constrained
matrices, the full covariance matrices deliver far worse results overall.

It therefore is clear that the best choice is to uniformly initialize the transition and start probabilities with
$k$-means-based estimations for the means and covariances of the emission distribution. Additionally, the covariance
matrices needs to be constrained to be diagonal.

\subsection{Factorial Hidden Markov Models}
\label{section:evaluation-hmm-fhmms}
Factorial Hidden Markov models are an important extension to HMMs when dealing with motions.
However, this benefit comes at the cost of being computationally more expensive than regular HMMs.
It therefore makes sense to not only consider the score that FHMMs achieve but to also consider the
time it takes to train and evaluate them.

During this evaluation, a total of 4 different models were evaluated:
\begin{itemize}
  \item regular HMM with $5$ states,
  \item FHMM with $5$ states and $2$ chains,
  \item FHMM with $5$ states and $3$ chains, and
  \item FHMM with $5$ states and $4$ chains
\end{itemize}
The FHMMs were trained using the sequential training algorithm (see chapter~\ref{section:recognition-fhmm}).
Each chain was trained for $10$ Baum-Welch iterations. The emission distribution parameters of the subsequent chains
were initialized on the residual error using the already discussed $k$-means approach. All other parameters
were set to the default values as defined above. The duration of each round was recorded as well since
computational feasibility should also be a consideration. For reference, the training was performed
on a machine with an 8-core \emph{Intel Core i7-4770} CPU clocked at \SI{3.40}{\GHz}. The results are
listed in table~\ref{table:fhmm-results}.

\begin{table}[h]
  \tvspace{1.2}
  \thspace{0.3cm}
  \begin{center}
      \begin{tabular}{@{}lrrrrr@{}} \toprule
      \multicolumn{2}{c}{\textbf{Model}} & & & \multicolumn{2}{c}{\textbf{Loglikelihood}} \\ \cmidrule(r){1-2} \cmidrule(l){5-6}
      \textbf{Type} & \textbf{Chains} & \textbf{Score} & \textbf{Duration} & \textbf{Positive} & \textbf{Negative} \\ \midrule
      HMM  & --- & 54\,782 &       49 sec & $31\,256 \pm 5\,296$ & $-722\,971 \pm 1\,213\,360$ \\ 
      FHMM & $2$ & 29\,369 &      741 sec & $32\,663 \pm 5\,189$ & $-356\,975 \pm 656\,688$ \\ 
      FHMM & $3$ & 18\,404 &   5\,137 sec & $28\,829 \pm 3\,873$ & $-208\,319 \pm 410\,939$ \\ 
      FHMM & $4$ & 12\,369 &  34\,890 sec & $25\,317 \pm 3\,572$ & $-132\,873 \pm 276\,476$ \\ \bottomrule
      \end{tabular}
    \end{center}
    \caption{Results of the comparison between FHMMs and HMMs. The given durations measure the time
    it took to train and evaluate the models over $3$ rounds each.}
    \label{table:fhmm-results}
\end{table}

The results are somewhat surprising: The scores of the models decrease as more chains are added. However,
the FHMM with $2$ chains shows promise. The median loglikelihood has increased significantly while the
standard deviation has decreased. This indicates that the model is capable of better fitting the data
at hand. FHMMs with more than $2$ chains, however, are not a good option. First, they do not seem to provide
a significant benefit over an HMM or an FHMM with $2$ chains. Additionally, the training and evaluation
times increase significantly. A possible explanation for this result is that only very few
states are required to discriminate motions (see section~\ref{section:evaluation-hmm-hyper}). The additional
history information that can be encoded by FHMMs with more than two chains appears to be counter-productive
in this case.

\section{Decision Makers}
\label{section:evaluation-decision-makers}
Recall that decision makers are used to find a mapping from the likelihoods of the (F)HMMs to the final
predictions. Since decision makers are usually classifiers, their hyperparameters must be selected as well.
This section therefore focuses on each individual decision maker whereas the next section focuses on
covering the entire classification process as a whole.

In the following subsections, four different decision makers will be evaluated: \emph{Logistic Regression}
and \emph{Support Vector Machines} are two binary classifiers that can be used with the binary relevance method (chapter~\ref{section:binary-relevance-method}).
In contrast, \emph{Decision Trees} and \emph{Random Forests} can directly be trained on the multi-label
problem and are therefore instances of models that use a modified learning algorithm (chapter~\ref{section:modified-algos}).
Notice that simpler decision makers like always picking the model with the maximum likelihood are not
considered in this section since they are parameter-free. They will be, however, considered during the
end-to-end evaluation that follows in the next section. Since all decision makers are evaluated similarly, this section starts by describing the general
evaluation process. This includes the definition of new measure since now a classification problem
is considered. The four decision makers are discussed during the remainder of this section.

The evaluation was performed on the usual dataset (section~\ref{section:dataset}). A different permutation
was used to shuffle the dataset. The best set of
features (section~\ref{section:feature-selection}) were then used to train $49$ Factorial Hidden Markov Models:
\emph{root\_pos}, \emph{root\_vel}, \emph{extremity\_pos}, \emph{root\_rot} and \emph{root\_rot\_norm}.
All features were normalized, smoothed and scaled as previously described.
For each FHMM, the optimal configuration as described in section~\ref{section:evaluation-hmm} was used:
each FHMM used $8$ states and $2$ chains with the unconstrained \emph{left-to-right} topology. The
initialization for the transition and start probabilities was uniform, with the means and covariances
of the emission distribution being initialized with the $k$-means method. The covariance matrices were constrained
to be diagonal. Each FHMM was trained using the sequential training algorithm with each chain being 
trained for $100$ Baum-Welch iterations.
Stratified $3$-fold was used to split the dataset into three training and test datasets. The loglikelihoods
under each model were then computed for each of the test and train splits. This data forms the basis for the following evaluation:
Each decision maker is trained on the train loglikelihoods and then evaluated on the respective test loglikelihoods.
Since a $3$-fold was used to split the dataset, the same split applies to the loglikelihoods; hence a total of
$3$ rounds can be performed. In each round, the decision maker is first trained on the train loglikelihoods.
After training, the decision maker is evaluated on the corresponding test loglikelihoods. The corresponding
labels of the dataset are available as well. Notice that the FHMMs were only trained and evaluated once for every label.
This re-using of loglikelihoods significantly speeds up the evaluation of the decision makers.

The evaluation of the decision makers can thus simply be seen as a $3$-fold evaluation of a supervised classifier where the loglikelihoods
are the features and the labels of the motions can be re-used as the targets. It therefore makes sense
to use metrics that are commonly used for classification problems to measure the results. To define measures,
it is useful to first define the following quantities:
\begin{itemize}
  \item The number of \emph{true positives} is the number of samples that were correctly classified as positive. To give an example,
  if the label of a samples is $y = 1$ and the classifier predicts $p = 1$, the sample is counted as true positive. The number of
  true positives is denoted as $\tp$.
  \item In contrast to a true positive, a sample is counted as a \emph{true negative} if it is correctly classified as negative.
  This would be the case for a sample with label $y = 0$ and prediction $p = 0$. The number of true negatives is denoted as $\tn$.
  \item If a sample is wrongly classified as positive but is actually labeled as negative, a \emph{false positive} occurs.
  To give an example, a sample with $y = 0$ and $p = 1$ would be counted as such. The number of false positives is denoted as $\fp$.
  \item Lastly, if a sample is wrongly classified as negative but is actually positive, a \emph{false negative} occurs. An example
  for this case would be a sample with $y = 1$ and $p = 0$. The number of false negatives is denoted as $\fn$.
\end{itemize}
These four quantities cover every possible outcome for binary classification (which is applicable here as well since
each label is encoded as a binary vector). Using these quantities, a couple of measures can be defined.

Firstly, the \emph{accuracy} is an extremely common measure. It is defined as:
\begin{equation}
  \accuracy = \frac{\tp + \tn}{\tp + \fp + \tn + \fn}.
\end{equation}
The accuracy simply measures the percentage of correctly classified samples. However, while being intuitive,
the accuracy also has a severe flaw. To illustrate this, consider the following example: Assume that a dataset
with $100$ samples is classified. Of the $100$ samples, only $1$ have a positive and the remaining $99$
have a negative label. Now consider a classifier that always predicts $p = 0$ regardless of the sample's
features. In this case, the classifier would achieve an accuracy of $\frac{99}{100} = 0.99$. It should be obvious
that, in this case, the accuracy is not a good metric to measure to quality of the classifier.

To overcome the shortcomings of the accuracy, the \emph{precision} and \emph{recall} are popular metrics:
\begin{equation}
  \precision = \frac{\tp}{\tp + \fp} \qquad \text{and} \qquad 
  \recall = \frac{\tp}{\tp + \fn}.
\end{equation}
Precision and recall are often combined into a single metric called the \emph{$F_1$ score}:
\begin{equation}
  F_1 = 2 \; \frac{\precision \cdot \recall}{\precision + \recall}.
\end{equation}
Like the accuracy, precision, recall and $F_1$ score are in the interval $[0,1]$. Generally speaking,
a value close to $1$ for all three scores is desirable~\cite{van:1979information}. In the remainder of this section, the $F_1$ score
is used to assess the decision makers. Since the $F_1$ score, precision and recall are computed per class, a possible way to combine the individual scores
is to take the mean over all classes per metric.

\subsection{Logistic Regression}
Logistic Regression is a very simple but popular binary classifier. It essentially combines linear regression
and the logistic function to perform binary classification. The reader is referred to other works
(e.g. \cite{Bishop:2006pattern}) for a full discussion.

\begin{figure}[h]
    \centering
    \includegraphics[width=\textwidth]{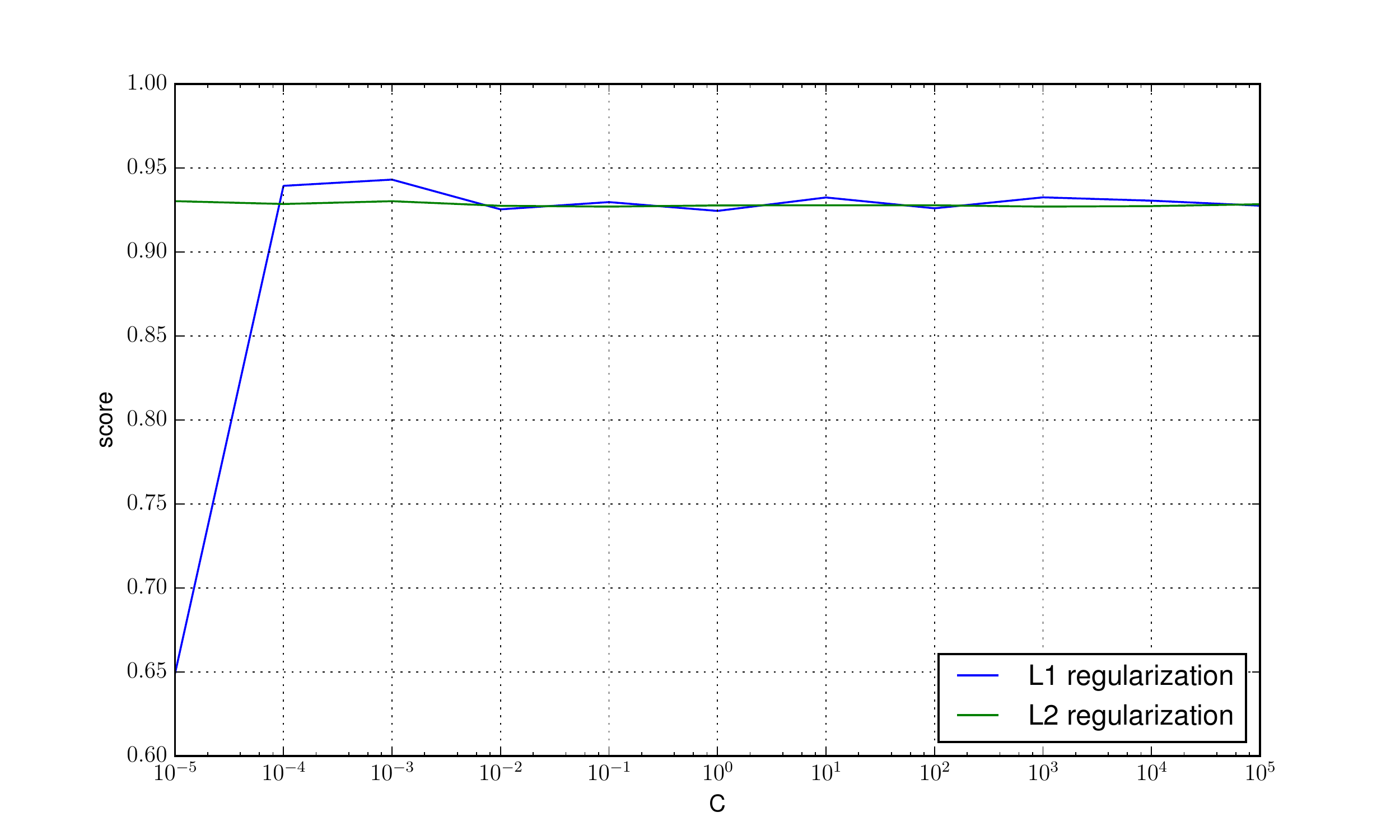}
    \caption{The results of the hyperparameter grid search for Logistic Regression. For each regularization type, the regularization coefficient is plotted against the achieved $F_1$ score.}
    \label{fig:log-regression-results}
\end{figure}

In this work, the \texttt{LogisticRegression}
implementation found in \emph{scikit-learn} with the \emph{liblinear} solver~\cite{fan:2008liblinear} is used.
Multi-label classification is achieved using the binary relevance method.
Since Logistic Regression is a rather simple model, the only hyperparameters considered in this work
are those that control regularization. More concretely, Logistic Regression can be used with L1 or
L2 regularization. Regularization is essentially an additional term in the cost function that penalizes large weights
to avoid overfitting. The difference between L1 and L2 regularization lies in the way the penalty term
is calculated: For L1 regularization, this happens using the $||\cdot||_1$ norm whereas L2 regularization
uses $||\cdot||_2$ instead. A coefficient $C$ controls the ``strength'' of the regularization. In case of the \emph{scikit-learn}
implementation, a small (that is close to zero) coefficient
corresponds to strong regularization whereas a larger coefficient relaxes it. To summarize, the following
hyperparameters and values are considered:
\begin{itemize}
  \item L1 vs. L2 regularization
  \item $C \in \{10^{-5}, 10^{-4}, \ldots, 10^4, 10^5\}$
\end{itemize}
This results in a total of $2 \cdot 11 = 22$ combinations, which were evaluated using grid search.
Each parameter combination was evaluated using a $3$-fold as described above. The $F_1$ score was computed
per class and then combined by taking the mean over all classes. Figure~\ref{fig:log-regression-results} depicts
the results.

As can be seen from the diagram, L1 regularization yields better results in this case. More concretely,
the best result with Logistic Regression is achieved with said regularization and $C = 10^{-3}$. This
makes sense since L1 regularization allows that many weights can become zero making it more suitable for sparse signals. This is the case for this dataset
since only a few labels will be set to $1$. Overall, Logistic Regression works very well, indicating
that the problem is linearly separable.

\subsection{Support Vector Machine}
Support Vector Machines~\cite{hearst:1998support} are another very popular binary classifier. Only
the very basics of Support Vector Machines are discussed here. Like Logistic Regression, SVMs try to
fit a hyperplane that separates the positive and negative samples in the feature space. The hyperplane
is fitted such that the distance between it and the samples that are close to it is maximized (\emph{large margin classifier}).
Additionally, only the feature vectors of close-by samples (so-called \emph{support vectors}) must be considered when fitting the hyperplane
which makes SVMs especially efficient on large datasets (since the majority of the samples can be ignored). Lastly, SVMs use what is commonly referred to as the
\emph{``kernel trick''} to fit problems that are not linearly separable. The fundamental idea here is to transform
the features to a very high-dimensional space. If the dimension is high enough, the samples will eventually be linearly
separable. However, such a transformation is computationally infeasible. The kernel trick uses kernel
functions which enables the classifier to operate in high-dimensional space without the need to actually compute
the coordinates~\cite{Bishop:2006pattern}.

\begin{figure}[h]
    \centering
    \includegraphics[width=\textwidth]{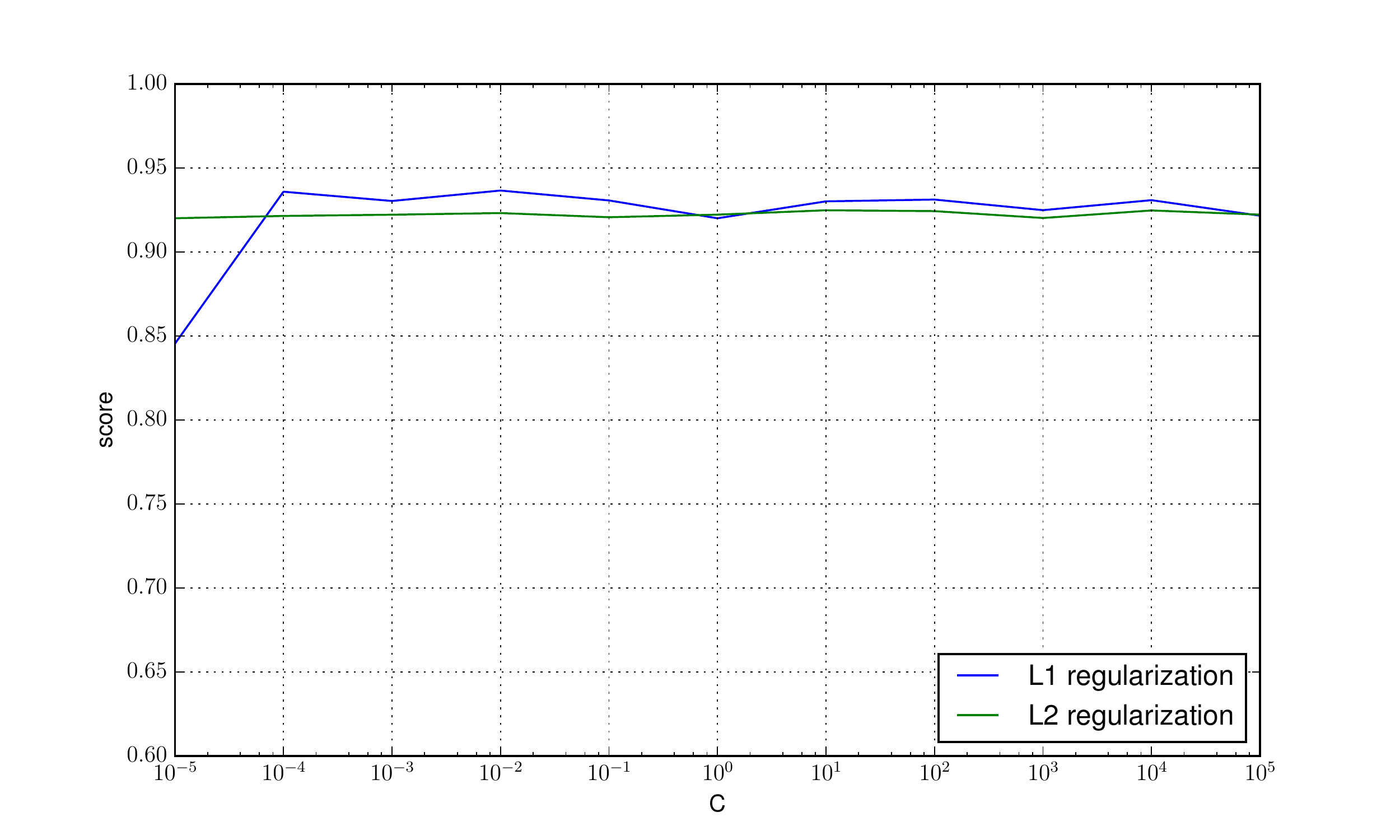}
    \caption{The results of the hyperparameter grid search for Support Vector Machine with a linear kernel. For each regularization type, the regularization coefficient is plotted against the achieved $F_1$ score.}
    \label{fig:svm-results}
\end{figure}

In this work, the linear SVM implementation found in \emph{scikit-learn} is used: \texttt{LinearSVC}. Under the hood,
\texttt{LinearSVC} uses the already mentioned \emph{liblinear} solver.
Notice that \texttt{LinearSVC} only supports a linear kernel. In this case, however, this is very much
sufficient since Logistic Regression already achieves very good results, indicating that the data
is linearly separable. The binary classifier is applied to the multi-label problem using the binary relevance method.
The following hyperparameters are under consideration:
\begin{itemize}
  \item L1 vs. L2 regularization
  \item $C \in \{10^{-5}, 10^{-4}, \ldots, 10^4, 10^5\}$
\end{itemize}
Notice the similarity to the hyperparameter of Logistic Regression. Furthermore, the
squared hinge loss was used instead of the more common hinge loss. This is necessary because the hinge loss cannot be used with L1 regularization
in the \emph{scikit-learn} implementation. The results of the grid search are depicted in figure~\ref{fig:svm-results}.

The results for the Support Vector Machine are very similar to the results achieved with Logistic
Regression, although Logistic Regression performed slightly better. The best hyperparameter combination as
measured by the $F_1$ score is achieved for L1 regularization with $C=10^{-2}$. As already discussed
during the evaluation of Logistic Regression, the L1 regularization allows sparse weights making
it a good choice for the dataset at hand.

\subsection{Decision Tree}
Decision Trees can be used to perform classification. In this case, the leaves of the tree represent
the classes and the inner nodes correspond to an input variable. The branches between nodes
represent decisions that are based on the input variables. A Decision Tree can then be traversed from
the root while a decision needs to be made at each inner node (e.g. if input variable $x$ is greater than some
threshold follow the left branch; follow the right branch otherwise) until a leaf node is reached.
The learning of such a model is performed by splitting the set of training samples
at each node. The relevant attribute and the threshold to make this decision are selected using a criterion,
e.g. the maximum information gain. This process is then repeated recursively for each child node until
either all samples belong to the same class (in which case the node turns into a leaf) or until some
other stopping criterion is reached (e.g. a set maximum depth)~\cite{Bishop:2006pattern}. Decision Trees can be extended to support
multi-label classification, which makes them an instance of a modified algorithm~~\cite{vens:2008decision}.

\begin{figure}[h]
    \centering
    \includegraphics[width=\textwidth]{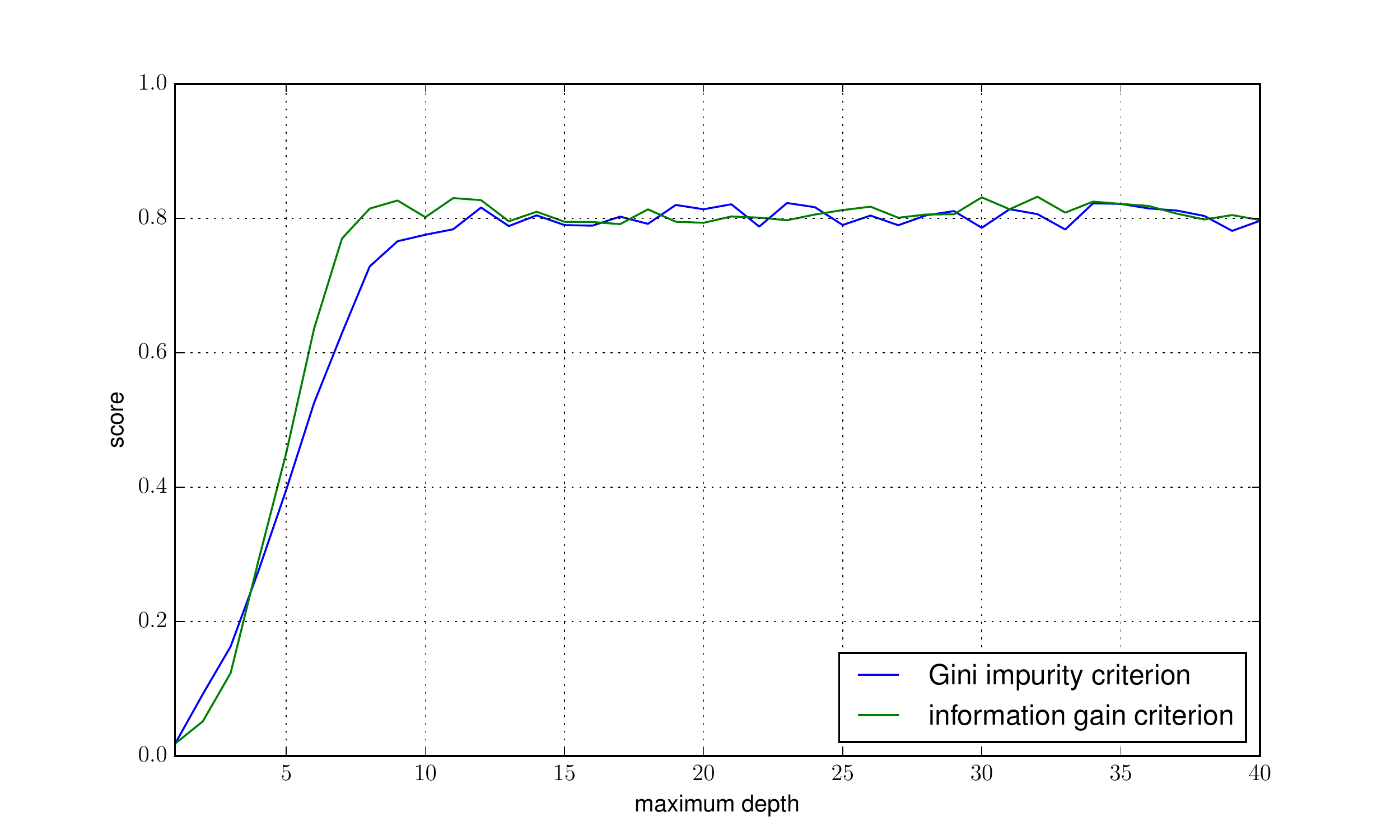}
    \caption{The results of the hyperparameter grid search for Decision Tree. For each splitting criterion the maximum allowed depth is plotted against the achieved $F_1$ score.}
    \label{fig:decision-tree-results}
\end{figure}

In this case, the \emph{scikit-learn} implementation is used: \texttt{DecisionTreeClassifier}. The implementation
uses the \emph{CART} algorithm~\cite{breiman:1984classification} to perform learning, which
has been modified to also support multi-label problems. The following important hyperparameters can be identified:
\begin{itemize}
  \item The criterion that measures the quality of a split. The implementation supports the \emph{Gini impurity}
  and the \emph{information gain}.
  \item The maximum depth that the Decision Tree can reach. In this case, the following values were considered: $\{1,2,\ldots,40\}$.
\end{itemize}
The $2 \cdot 40 = 80$ possible combinations were explored using grid search. The evaluation was performed
as previously described and the averaged $F_1$ score was used as a measure. Figure~\ref{fig:decision-tree-results} depicts the
results.

As can be seen from the results, the Decision Tree does not need to be very deep to work properly. Trees
that use the information gain splitting criterion required less depth than trees where the Gini
impurity was used. However, both criteria eventually converge to approximately the same score of
$F_1 = 0.8$. Since the information gain criterion reaches this convergence faster, it should be preferred.
In either case, a tree with a maximum depth of $15$ is sufficient to consistently learn
the problem at hand. However, the Decision Tree classifier is outperformed
by Logistic Regression and SVMs on this dataset.

\subsection{Random Forest}
The last classifier that is discussed here is called \emph{Random Forest}~\cite{breiman:2001random}. A Random Forest is
a so-called \emph{ensemble classifier}: This means that it uses multiple, potentially weak, classifiers
internally and combines their predictions into a final prediction. Random Forests do this by
using multiple Decision Trees internally. The final prediction is then computed by taking a majority vote over
the prediction of each tree. To avoid ending up with almost identical trees, randomness is introduced
during training. More concretely, this is achieved by two factors: \emph{Bootstrapping}~\cite{efron:1979bootstrap}
is used to fit trees on re-sampled training examples. Additionally, Decision Trees do not select the
best split but randomize this process by only considering a random subset of the available features.
Since Random Forests use Decision Trees internally, they can be used to perform multi-label classification
as well if the trees support it.

\begin{figure}[h]
    \centering
    \includegraphics[width=\textwidth]{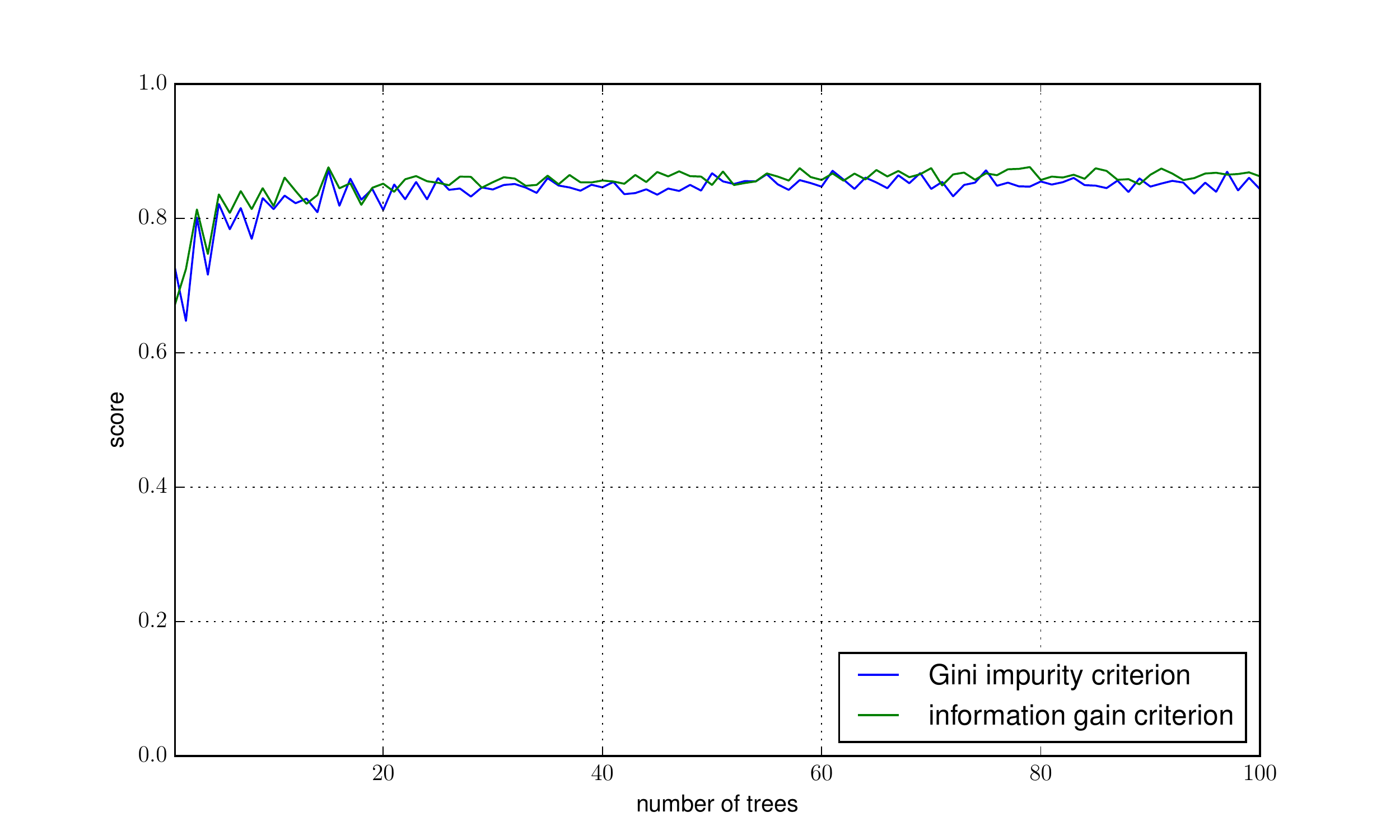}
    \caption{The results of the hyperparameter grid search for Random Forest. For each splitting criterion, the number of internally used Decision Trees is plotted against the achieved $F_1$ score.}
    \label{fig:random-forest-results}
\end{figure}

In this case, the \emph{scikit-learn} implementation is used: \texttt{RandomForestClassifier}. The implementation
uses \texttt{DecisionTreeClassifier}s internally using the same training algorithm as described above.
As already mentioned, the splitting decision is now randomized. Since Decision Trees were already
considered in the previous section, a maximum depth of $15$ was selected. However, the following additional hyperparameters
are considered:
\begin{itemize}
  \item The criterion that measures the quality of a split. The implementation supports the \emph{Gini impurity}
  and the \emph{information gain}. This is re-evaluated since the splitting decision is now randomized.
  \item The number of Decision Trees that are grown in the Random Forest classifier: $\{1,2,\ldots,100\}$.
\end{itemize}
The $2 \cdot 100 = 200$ possible combinations were explored using grid search. The evaluation was performed
as previously described and the averaged $F_1$ score was used as a measure. Figure~\ref{fig:random-forest-results} depicts the
results.

As can be seen from the data, Random Forests improve as more and more Decision Trees are used
until they eventually converge to a similar $F_1$ score. A couple of interesting conclusions can be drawn from the results: Firstly, the
information gain splitting criterion generally performs better than the Gini impurity on this problem. This was already
the case during the evaluation of the individual Decision Trees. Secondly, a Random
Forest that uses a single Decision Tree internally is worse than a single Decision Tree although they
share the same parameters. This nicely illustrates that trees in a Random Forest are, well, randomized.
However, as more and more suboptimal trees are added, the Random Forest suddenly outperforms the best
Decision Tree from the previous section. This is the crucial property that makes ensemble methods so successful.
Still, Logistic Regression and SVMs achieved better results on this dataset.

\section{Classification Systems}
After having evaluated the individual components of the system in the previous sections, this section
brings it all together by evaluating the entire system end-to-end. To do so, two fundamental approaches
must be distinguished first. Recall from chapter~\ref{section:multi-label-classification} that the multi-label
classification problem can either be transformed to a single-label problem by treating each combination
of labels as a single substitute label. In this case, each HMM represents such a substitute class and
the classification is then simply a matter of selecting the class that corresponds to the model with the maximum likelihood.
The second option truly handles the multi-label problem by using one HMM per label. This means that a single
motion is potentially used to train multiple HMMs. However, in this case a more advanced decision maker
is needed to map the likelihoods of the different HMMs to the multi-label prediction since simply picking
the maximum is not an option anymore. Since these two approaches are inherently different they are treated
as two distinct systems: The system that uses the former method is referred to as the \emph{power set system}
while the latter system is referred to as the \emph{multi-label system}. Notice however that both
systems can be used to solve the same multi-label classification problem.

This section also makes use of an additional measure: the \emph{total accuracy}. In contrast
to the accuracy and the $F_1$ score which are computed on a per-class basis, the total accuracy measures
how often the entire binary label vector $\vec{y}$ was correctly predicted. If, for example, the prediction
is $\vec{p} = (1, 0, 0, 1, 1, 1)$ and the correct label is $\vec{y} = (1, 0, 0, 1, 1, 0)$, the total
accuracy would be $0$ despite the fact that the classifier ``almost'' got it right. It should be obvious
from this example that the total accuracy is a rather harsh and unforgiving measure. Only the total
accuracy is used in this chapter.

During this section, the results of the two systems are first considered individually with a direct comparison
following subsequently. Both systems were evaluated on the same dataset that was already used for
all other evaluations. A new permutation was used to initially shuffle the dataset. However, to make the two
systems comparable, the permutation was the same for both systems. Stratified
$3$-fold was used to split the dataset into training and test folds. All measures were then computed
over the combined test splits.

From each section, the best results were used to perform the end-to-end evaluation. However, often
multiple configurations showed promise in which case the best couple of configurations were selected.
More concretely, the following best feature sets were selected from the results in section~\ref{section:feature-selection}:
\begin{enumerate}
  \item \emph{root\_rot\_norm}
  \item \emph{root\_pos}, \emph{root\_vel}, \emph{extremities\_pos}, \emph{root\_rot} and \emph{root\_rot\_norm}
  \item \emph{extremities\_pos}, \emph{root\_rot} and \emph{root\_rot\_norm}
  \item \emph{root\_pos}, \emph{root\_vel}, \emph{com\_pos}, \emph{extremities\_pos}, \emph{root\_rot} and \emph{root\_rot\_norm}
  \item \emph{root\_vel}, \emph{extremities\_pos}, \emph{root\_rot} and \emph{root\_rot\_norm}
  \item \emph{root\_pos}, \emph{root\_vel}, \emph{com\_pos}, \emph{extremities\_pos}, \emph{root\_rot}, \emph{root\_rot\_norm} and \emph{marker\_vel\_norm}
\end{enumerate}
All features were normalized, smoothed with $W = 3$ and scaled to be in the interval $[-1, 1]$. During this evaluation,
the feature sets will be referenced by their enumeration number.

Both HMMs and FHMMs were used during this evaluation. Since the results in section~\ref{section:evaluation-hmm-fhmms}
clearly showed that using FHMMs with more than two chains do not offer any significant advantage,
only HMMs and FHMMs with $2$ chains were considered in the following evaluation. All HMMs were trained for $100$ Baum-Welch iterations.
The FHMMs were trained using the sequential training algorithm and $100$ Baum-Welch iterations were performed
per chain. The transition and start probabilities were initialized uniformly whereas the emission distribution parameters
were initialized using the $k$-means approach. The covariance matrices were all constrained to be diagonal. The
decision to only use this initialization strategy was made based on the results in section~\ref{section:evaluation-hmm-init} which
showed that the other combinations do not have any advantages.

For each of the four different topologies evaluated in section~\ref{section:evaluation-hmm-hyper},
the best number of states were selected:
\begin{enumerate}
  \item \emph{fully-connected} topology and $K=8$ states
  \item \emph{left-to-right} topology without $\Delta$ constraint and $K=5$ states
  \item \emph{left-to-right} topology with $\Delta = 1$ and $K=6$ states
  \item \emph{left-to-right} topology with $\Delta = 2$ and $K=5$ states
\end{enumerate}

\subsection{Power Set System}
For the power set system, a total of $6 \cdot 2 \cdot 4 = 48$ different configuration combinations
were evaluated as discussed above. Each combination of labels in the original dataset was replaced with a
single substitute label. Since $54$ different combinations can be identified in the dataset (see section~\ref{section:dataset}),
a total of $54$ of such substitute labels were used. Analogously $54$ HMMs or FHMMs (depending on the configuration)
were trained. Classification was then performed by selecting the model under which the unknown motion
had the highest likelihood. The $20$ best results as measured by their $F_1$ score are listed in
table~\ref{table:results-overview-power-set}. The $F_1$ scores, precisions and recalls were averaged
over all classes. Averaging is not necessary for the total accuracy since this measure already considers
all classes combined.

\begin{table}[h]
  \tvspace{1.2}
  \thspace{0.3cm}
  \begin{center}
      \begin{tabular}{@{}rllrrrrr@{}} \toprule
      \textbf{Feature Set} & \textbf{Model} & \textbf{Topology} & \textbf{K} & \textbf{$F_1$ score} & \textbf{Precision} & \textbf{Recall} & \textbf{Accuracy} \\ \midrule
      2 & HMM & fully connected & $8$ & \underline{0.9742} & 0.9797 & \underline{0.9727} & 0.9780 \\ 
      4 & HMM & fully connected & $8$ & 0.9739 & 0.9788 & \underline{0.9727} & 0.9780 \\ 
      6 & FHMM & fully connected & $8$ & 0.9736 & \underline{0.9800} & \underline{0.9727} & 0.9780 \\ 
      5 & HMM & fully connected & $8$ & 0.9723 & 0.9786 & 0.9709 & 0.9780 \\ 
      4 & FHMM & left-to-right $\Delta=1$ & 6 & 0.9720 & 0.9777 & 0.9709 & 0.9780 \\ 
      6 & FHMM & left-to-right & $5$ & 0.9712 & 0.9774 & 0.9711 & 0.9802 \\ 
      6 & FHMM & left-to-right $\Delta=1$ & 6 & 0.9712 & 0.9774 & 0.9711 & \underline{0.9802} \\ 
      4 & FHMM & fully connected & $8$ & 0.9704 & 0.9762 & 0.9690 & 0.9736 \\ 
      5 & FHMM & left-to-right $\Delta=1$ & 6 & 0.9675 & 0.9736 & 0.9672 & 0.9758 \\ 
      6 & FHMM & left-to-right $\Delta=2$ & $5$ & 0.9672 & 0.9752 & 0.9674 & 0.9780 \\ 
      6 & HMM & fully connected & $8$ & 0.9667 & 0.9735 & 0.9653 & 0.9736 \\ 
      5 & HMM & left-to-right $\Delta=1$ & 6 & 0.9656 & 0.9740 & 0.9625 & 0.9714 \\ 
      5 & FHMM & fully connected & $8$ & 0.9653 & 0.9737 & 0.9653 & 0.9736 \\ 
      4 & HMM & left-to-right $\Delta=1$ & 6 & 0.9651 & 0.9737 & 0.9635 & 0.9714 \\ 
      2 & FHMM & fully connected & $8$ & 0.9632 & 0.9705 & 0.9635 & 0.9714 \\ 
      6 & HMM & left-to-right $\Delta=1$ & 6 & 0.9632 & 0.9709 & 0.9616 & 0.9714 \\ 
      2 & FHMM & left-to-right $\Delta=2$ & $5$ & 0.9615 & 0.9715 & 0.9600 & 0.9714 \\ 
      3 & HMM & fully connected & $8$ & 0.9615 & 0.9709 & 0.9580 & 0.9648 \\ 
      5 & FHMM & left-to-right & $5$ & 0.9610 & 0.9721 & 0.9600 & 0.9714 \\ 
      4 & FHMM & left-to-right & $5$ & 0.9598 & 0.9678 & 0.9600 & 0.9692 \\ \bottomrule
      \end{tabular}
  \end{center}
  \caption{The $20$ best configurations of the power set system as determined by their $F_1$ score. For each score, the best
  result is underlined. \emph{Accuracy} lists the total accuracies.}
  \label{table:results-overview-power-set}
\end{table}

A first conclusion that can be drawn from the data is that the overall performance of the system is
very good. The best configuration achieved an $F_1$ score of $0.9742$ and a total accuracy of $0.9727$.
Another important observation is that the precision and recall scores are nicely balanced across
configurations, hence the system does not show a bias towards the one or the other. No clear best option
can be identified for the different feature sets. Set $2$, $4$, $5$ and $6$ all delivered very good
performance with feature set $3$ doing slightly worse. However, feature set $1$ performed very poorly,
with not a single instance of this configuration in the top $20$. Another interesting result is that
FHMMs did not provide a significant advantage over HMMs. The best total accuracy was achieved by an FHMM whereas the best
$F_1$ score was achieved by an HMM. As can be seen from the data, both HMMs and FHMMs achieved very similar
scores overall. Since FHMMs are computationally more expensive and more complex to implement,
HMMs should be preferred. Lastly, the different topologies and states did not have a big impact
on the performance. All different combinations appear in the top $20$. However, it should be noted
that the best results were achieved with fully-connected HMMs and left-to-right FHMMs respectively.
Despite this apparent correlation, the effect on the overall performance is only very small.

Table~\ref{table:results-details-power-set} contains the detailed results for the configuration that
achieved the best $F_1$ score.
Notice again that multiple labels have been combined into a single substitute label. The data also
shows that the classification scores are consistently very high across all classes.

\begin{table}[H]
  \small
  \tvspace{1.02}
  \thspace{0.3cm}
  \begin{center}
      \begin{tabular}{@{}lrrr@{}} \toprule
      \textbf{Class} & \textbf{$F_1$ score} & \textbf{Precision} & \textbf{Recall} \\ \midrule
      walk, turn-right                                        & 0.984      & 0.969     & 1.000 \\ 
      walk, turn-left                                         & 0.982      & 1.000     & 0.964 \\ 
      walk, speed-normal, direction-forward                   & 0.900      & 0.900     & 0.900 \\ 
      walk, direction-forward, speed-slow                     & 0.900      & 0.900     & 0.900 \\ 
      walk, direction-forward, speed-fast                     & 1.000      & 1.000     & 1.000 \\ 
      walk, bend-left                                         & 1.000      & 1.000     & 1.000 \\ 
      walk, bend-right                                        & 1.000      & 1.000     & 1.000 \\ 
      push-recovery, direction-backward                       & 1.000      & 1.000     & 1.000 \\ 
      direction-forward, push-recovery                        & 0.941      & 1.000     & 0.889 \\ 
      push-recovery, direction-left                           & 1.000      & 1.000     & 1.000 \\ 
      push-recovery, direction-right                          & 1.000      & 1.000     & 1.000 \\ 
      walk, direction-circle, counter-clockwise               & 1.000      & 1.000     & 1.000 \\ 
      walk, direction-circle, clockwise                       & 0.971      & 0.944     & 1.000 \\ 
      walk, direction-slalom                                  & 1.000      & 1.000     & 1.000 \\ 
      direction-forward, run                                  & 0.988      & 0.976     & 1.000 \\ 
      walk, direction-upward                                  & 1.000      & 1.000     & 1.000 \\ 
      kick, foot-right                                        & 0.933      & 1.000     & 0.875 \\ 
      throw, hand-right                                       & 0.909      & 0.833     & 1.000 \\ 
      throw, hand-left                                        & 1.000      & 1.000     & 1.000 \\ 
      bow, deep                                               & 1.000      & 1.000     & 1.000 \\ 
      bow, slight                                             & 0.889      & 1.000     & 0.800 \\ 
      kick, foot-right, high                                  & 0.909      & 0.833     & 1.000 \\ 
      kick, high, foot-left                                   & 1.000      & 1.000     & 1.000 \\ 
      kick, foot-right, low                                   & 1.000      & 1.000     & 1.000 \\ 
      kick, foot-left, low                                    & 1.000      & 1.000     & 1.000 \\ 
      squat                                                   & 1.000      & 1.000     & 1.000 \\ 
      hand-right, punch                                       & 0.889      & 1.000     & 0.800 \\ 
      hand-left, punch                                        & 1.000      & 1.000     & 1.000 \\ 
      foot-right, stomp                                       & 1.000      & 1.000     & 1.000 \\ 
      foot-left, stomp                                        & 1.000      & 1.000     & 1.000 \\ 
      direction-upward, jump                                  & 0.889      & 1.000     & 0.800 \\ 
      direction-forward, jump                                 & 1.000      & 1.000     & 1.000 \\ 
      direction-backward, jump                                & 1.000      & 1.000     & 1.000 \\ 
      direction-left, jump                                    & 0.909      & 0.833     & 1.000 \\ 
      direction-right, jump                                   & 1.000      & 1.000     & 1.000 \\ 
      golf, putting                                           & 1.000      & 1.000     & 1.000 \\ 
      golf, drive                                             & 1.000      & 1.000     & 1.000 \\ 
      hand-right, tennis, smash                               & 1.000      & 1.000     & 1.000 \\ 
      hand-left, tennis, smash                                & 1.000      & 1.000     & 1.000 \\ 
      hand-right, tennis, forehand                            & 1.000      & 1.000     & 1.000 \\ 
      hand-left, tennis, forehand                             & 1.000      & 1.000     & 1.000 \\ 
      hand-right, wave                                        & 0.909      & 0.833     & 1.000 \\ 
      hand-left, wave                                         & 1.000      & 1.000     & 1.000 \\ 
      wave, hand-both                                         & 1.000      & 1.000     & 1.000 \\ 
      hand-right, play-guitar                                 & 1.000      & 1.000     & 1.000 \\ 
      hand-left, play-guitar                                  & 1.000      & 1.000     & 1.000 \\ 
      hand-right, play-violin                                 & 1.000      & 1.000     & 1.000 \\ 
      hand-left, play-violin                                  & 1.000      & 1.000     & 1.000 \\ 
      hand-right, stir                                        & 1.000      & 1.000     & 1.000 \\ 
      hand-left, stir                                         & 1.000      & 1.000     & 1.000 \\ 
      hand-right, wipe                                        & 0.889      & 1.000     & 0.800 \\ 
      hand-left, wipe                                         & 1.000      & 1.000     & 1.000 \\ 
      dance, waltz                                            & 1.000      & 1.000     & 1.000 \\ 
      dance, chachacha                                        & 1.000      & 1.000     & 1.000 \\ \bottomrule
    \end{tabular}
  \end{center}
  \caption{The results of the power set system on a per-class basis for the best combination.}
  \label{table:results-details-power-set}
\end{table}

\subsection{Multi-Label System}
The multi-label system was evaluated similarly. However, this system requires more sophisticated
decision makers to map the likelihood scores of the HMMs to the binary prediction vector. For each
of the previously evaluated decision makers, the best configurations (see section~\ref{section:evaluation-decision-makers}) was used:
\begin{enumerate}
  \item \emph{Logistic Regression} with L1 penalty and $C=10^{-3}$ using the binary relevance method
  \item \emph{Support Vector Machine} with L1 penalty and $C=10^{-2}$ using the binary relevance method
  \item \emph{Decision Tree} with information gain splitting criterion and a maximum depth of $15$
  \item \emph{Random Forest} with information gain splitting criterion, a maximum depth of $15$ and $40$ internal Decision Trees
\end{enumerate}
Additionally, two much simpler decision makers were evaluated as well: The \emph{zero} decision maker
always predicts all classes that correspond to HMMs with a likelihood score greater than or equal to zero.
The \emph{maximum} decision maker works as described in the previous section by always selecting only a
single class which corresponds to the HMM with the highest likelihood.
Therefore a total of $6~\cdot~2~\cdot~4~\cdot~6~=~288$ different configuration combinations
were evaluated as discussed above. Since $49$ different classes exist (see section~\ref{section:dataset})
a total of $49$ HMMs or FHMMs (depending on the configuration) were trained. Classification was then performed
using the decision maker of the respective configuration. The $20$ best results as measured by their $F_1$ score are listed in
table~\ref{table:results-overview-multi-label}. 

\begin{table}[h]
  \tvspace{1.2}
  \thspace{0.2105cm}
  \begin{center}
      \hspace*{-20pt}
      \begin{tabular}{@{}rllrlrrrr@{}} \toprule
      \textbf{F.-Set} & \textbf{Model} & \textbf{Topology} & \textbf{K} & \textbf{Decision Maker} & \textbf{$F_1$ score} & \textbf{Precision} & \textbf{Recall} & \textbf{Accuracy} \\ \midrule
      5 & HMM & left-to-right $\Delta=2$ & 5 & Logistic Regression &  \underline{0.9662}  & 0.9680  & 0.9662  & 0.9229    \\ 
      2 & HMM & left-to-right $\Delta=2$ & 5 & Logistic Regression &  0.9660  & 0.9642  & 0.9696  & 0.9229    \\ 
      4 & HMM & left-to-right & 5 & Logistic Regression & 0.9634  & 0.9598  & 0.9690  & 0.9229    \\ 
      2 & HMM & left-to-right $\Delta=1$ & 6 & Logistic Regression  & 0.9631  & 0.9629  & 0.9660  & \underline{0.9339}    \\ 
      5 & HMM & left-to-right & 5 & Logistic Regression & 0.9628  & 0.9635  & 0.9652  & 0.9207    \\ 
      5 & HMM & left-to-right & 5 & SVM & 0.9628  & 0.9595  & 0.9691  & 0.9031    \\ 
      2 & FHMM & left-to-right $\Delta=1$ & 6 & Logistic Regression & 0.9627  & 0.9692  & 0.9603  & 0.9295    \\ 
      4 & HMM & left-to-right $\Delta=2$ & 5 & Logistic Regression & 0.9621  & 0.9576  & 0.9682  & 0.9229     \\ 
      5 & FHMM & fully-connected & 8 & Logistic Regression & 0.9620  & 0.9647  & 0.9620  & 0.9207     \\ 
      6 & HMM & left-to-right $\Delta=1$ & 6 & SVM & 0.9618  & 0.9533  & \underline{0.9735}  & 0.9163     \\ 
      5 & HMM & left-to-right $\Delta=2$ & 5 & SVM & 0.9616  & 0.9576  & 0.9694  & 0.9097     \\ 
      5 & FHMM & left-to-right $\Delta=2$ & 5 & Logistic Regression & 0.9610  & 0.9651  & 0.9616  & 0.9251    \\ 
      2 & FHMM & left-to-right $\Delta=2$ & 5 & Logistic Regression & 0.9609  & 0.9640  & 0.9618  & 0.9229    \\ 
      6 & FHMM & left-to-right $\Delta=1$ & 6 & Logistic Regression & 0.9600  & 0.9674  & 0.9564  & 0.9273     \\ 
      5 & HMM & left-to-right $\Delta=1$ & 6 & Logistic Regression & 0.9597  & \underline{0.9694}  & 0.9531  & 0.9141     \\ 
      5 & FHMM & left-to-right & 5 & Logistic Regression & 0.9595  & 0.9628  & 0.9617  & 0.9207     \\ 
      2 & FHMM & fully-connected & 8 & Logistic Regression & 0.9592  & 0.9632  & 0.9599  & 0.9163     \\ 
      6 & HMM & left-to-right $\Delta=2$ & 5 & Logistic Regression & 0.9590  & 0.9526  & 0.9682  & 0.9251     \\ 
      6 & HMM & fully-connected & 8 & Logistic Regression & 0.9586  & 0.9661  & 0.9576  & 0.9185    \\ 
      2 & HMM & left-to-right $\Delta=1$ & 6 & SVM & 0.9578  & 0.9567  & 0.9632  & 0.9097     \\ \bottomrule
      \end{tabular}
  \end{center}
  \caption{The $20$ best configurations of the multi-label system as determined by their $F_1$ score. For each score, the best
  result is underlined. \emph{Accuracy} lists the total accuracies.}
  \label{table:results-overview-multi-label}
\end{table}

The results of the multi-label system are also very good. The best achieved $F_1$ score was $0.9662$ with
the best total accuracy at $0.9339$. Precision and recall are once again balanced, so a bias towards
one or the other does not exist. An important observation is that HMMs slightly outperform FHMMs. In general
FHMMs do not offer any noticeable advantage of HMMs on this dataset. Another result is that
the left-to-right topologies dominate the top $20$ results. However, since the $F_1$ scores are still
very similar across topologies, they are still not very important for the overall performance of the classifier.
Similar to the previously discussed system, the feature sets $2$, $4$, $5$ and $6$ all achieve good results.
Feature set $1$ and $3$ did not make it into the top $20$. The best achieved $F_1$ score for feature
set $3$ was $0.9513$ and for the first set the best score was $0.5084$. Again, feature set $1$ is obviously
not a good choice.

A very important aspect of the classifier is the choice of the decision maker. As can be seen from the above results,
the top $20$ best scores were all achieved using either Logistic Regression or Support Vector Machines.
Logistic Regression consistently outperformed SVMs but the effect on the $F_1$ score is very small.
The best achieved $F_1$ score for Random Forests was $0.9029$ and the best score using Decision Trees
was $0.8799$. In general, the results are consistent with the results of section~\ref{section:evaluation-decision-makers}.
The best maximum decision maker achieved an $F_1$ score of $0.7248$. It should be noted however that
the total accuracy dropped to an abysmal score of $0.0110$. This result makes sense since the maximum
decision maker only selects a single class but most motions belong to multiple classes. This results
in a somewhat decent $F_1$ score since the prediction likely contains one correct label. However, a prediction
made by the maximum decision maker can only be completely correct if a motion only has a single label--which is not the case
for most motions in the dataset. The zero decision maker finally achieved an $F_1$ score of $0.3711$
and a total accuracy of $0.2445$. To summarize, a linear model is obviously the best choice for the decision maker.
Logistic Regression proved to perform especially well.

Similar to the previous section, table~\ref{table:results-details-multi-label} (at the very end of this section) contains a more detailed
characterization of the best classifier configuration as measured by the $F_1$ score. Overall, the $F_1$ score is consistently high
across classes.

\begin{table}[h]
  \small
  \tvspace{1.1}
  \thspace{0.3cm}
  \begin{center}
      \begin{tabular}{@{}lrrr@{}} \toprule
      \textbf{Class} & \textbf{$F_1$ score} & \textbf{Precision} & \textbf{Recall} \\ \midrule
      walk                             &  1.000    &   1.000   &  1.000   \\ 
      turn-right                       &  1.000    &   1.000   &  1.000   \\ 
      turn-left                        &  0.982    &   1.000   &  0.964   \\ 
      speed-normal                     &  0.700    &   0.700   &  0.700   \\ 
      direction-forward                &  0.988    &   0.977   &  1.000   \\ 
      speed-slow                       &  0.857    &   0.818   &  0.900   \\ 
      speed-fast                       &  1.000    &   1.000   &  1.000   \\ 
      bend-left                        &  1.000    &   1.000   &  1.000   \\ 
      bend-right                       &  1.000    &   1.000   &  1.000   \\ 
      push-recovery                    &  1.000    &   1.000   &  1.000   \\ 
      direction-backward               &  0.963    &   1.000   &  0.929   \\ 
      direction-left                   &  0.968    &   0.938   &  1.000   \\ 
      direction-right                  &  1.000    &   1.000   &  1.000   \\ 
      direction-circle                 &  1.000    &   1.000   &  1.000   \\ 
      counter-clockwise                &  1.000    &   1.000   &  1.000   \\ 
      clockwise                        &  1.000    &   1.000   &  1.000   \\ 
      direction-slalom                 &  0.981    &   1.000   &  0.963   \\ 
      run                              &  0.964    &   0.952   &  0.976   \\ 
      direction-upward                 &  1.000    &   1.000   &  1.000   \\ 
      kick                             &  1.000    &   1.000   &  1.000   \\ 
      foot-right                       &  0.979    &   0.958   &  1.000   \\ 
      throw                            &  0.857    &   0.818   &  0.900   \\ 
      hand-right                       &  0.989    &   1.000   &  0.978   \\ 
      hand-left                        &  0.989    &   1.000   &  0.979   \\ 
      bow                              &  1.000    &   1.000   &  1.000   \\ 
      deep                             &  0.909    &   0.833   &  1.000   \\ 
      slight                           &  1.000    &   1.000   &  1.000   \\ 
      high                             &  1.000    &   1.000   &  1.000   \\ 
      foot-left                        &  0.933    &   0.933   &  0.933   \\ 
      low                              &  1.000    &   1.000   &  1.000   \\ 
      squat                            &  0.909    &   0.833   &  1.000   \\ 
      punch                            &  0.952    &   0.909   &  1.000   \\ 
      stomp                            &  0.952    &   0.909   &  1.000   \\ 
      jump                             &  0.962    &   0.926   &  1.000   \\ 
      golf                             &  1.000    &   1.000   &  1.000   \\ 
      putting                          &  1.000    &   1.000   &  1.000   \\ 
      drive                            &  0.923    &   0.857   &  1.000   \\ 
      tennis                           &  0.923    &   0.947   &  0.900   \\ 
      smash                            &  0.842    &   0.889   &  0.800   \\ 
      forehand                         &  0.778    &   0.875   &  0.700   \\ 
      wave                             &  1.000    &   1.000   &  1.000   \\ 
      hand-both                        &  1.000    &   1.000   &  1.000   \\ 
      play-guitar                      &  1.000    &   1.000   &  1.000   \\ 
      play-violin                      &  1.000    &   1.000   &  1.000   \\ 
      stir                             &  0.800    &   0.889   &  0.727   \\ 
      wipe                             &  0.857    &   0.900   &  0.818   \\ 
      dance                            &  1.000    &   1.000   &  1.000   \\ 
      waltz                            &  1.000    &   1.000   &  1.000   \\ 
      chachacha                        &  1.000    &   1.000   &  1.000   \\ \bottomrule
      \end{tabular}
  \end{center}
  \caption{The results of the multi-label system on a per-class basis for the best combination.}
  \label{table:results-details-multi-label}
\end{table}

\subsection{Comparison}
Both systems
did not benefit from the use of FHMMs in any significant way. Additionally, feature sets $2$, $4$, $5$ and $6$
proofed to be good choices for either system. The choice of topology and number of states did not have
a significant effect on the performance although the left-to-right topology achieved better results
in the multi-label system.

The power set system achieved an $F_1$ score of $0.9742$ whereas the multi-label system achieved $0.9662$.
A more significant difference can be seen in the total accuracy: $0.9802$ was the best achieved score
for the power set system whereas the multi-label system achieved $0.9339$. It is interesting that the
$F_1$ scores are relatively similar whereas there's a noticeable difference in total accuracy. An explanation
for these results is that the power set system is less likely to make a mistake since only ever one label
is selected. Since most labels are properly recognized, the total accuracy is very high. In general, the power set
system can only get a prediction right or wrong. The multi-label system on the other hand has a trickier job:
The system does not know in advance how many labels a motion has. This additional uncertainty results in a slightly worse total accuracy.
This means that the system will produce some predictions that are ``almost right''. However, these are counted
as wrong by the total accuracy, resulting in a worse score. In summary the recognition
performance across classes is similar for both systems, hence the $F_1$ score is also very similar.
Due to the difference in how the prediction is made, the total accuracy varies. However, it is important
to stress that the superior performance of the power set method is bought with additional resources: Since
each label combination is treated as a single substitute label, more HMMs need to be kept in memory. During classification
of an unknown motion, the likelihoods under each HMM must be evaluated, resulting in higher computational complexity.
Additionally, sparse training data can become a problem (also see discussion in chapter~\ref{section:classification-power-set-method}).

On this dataset, the power set system does not run into any of these problems since only $54$ label combinations exist and enough training data is available
per combination. However, as the dataset grows and more and more label combinations become possible, the power set method
will eventually become infeasible. The multi-label system does not have this problem and therefore certainly
has its place.

\cleardoublepage

\chapter{Conclusion}
\label{chapter:conclusion}
The goal of this thesis was to develop a system that can accurately classify human whole-body
motion. A key consideration was that multiple labels are needed to describe a single motion. During the course of this work,
the different parts necessary to perform multi-label classification were introduced, discussed and evaluated.
An especially important aspect of this thesis was the discussion of different representations of the motions as features as well
as the extraction of novel features from the raw data. The section on how multi-label classification
can be performed was equally important to achieve the required multi-label classification. Two different
systems were discussed: The power set system transforms the multi-label problem to a single-label problem
by treating each label combination as a substitute class. The multi-label system was developed using an approach novel
in the field of motion recognition for robotics that truly handles multi-label problems\footnote{based upon fundamental research from the field of machine learning}.

The systems and their components were evaluated using a dataset that consists of $454$ motions. $49$ different labels were
used to describe each motion with $54$ unique label combinations. Due to the high number of parameters
and components, individual building blocks of the system were evaluated and optimized in isolation before an end-to-end evaluation
of the whole proposed approach was conducted.

First, feature selection was performed to find the best set of features. An important insight was that the joint angles and marker positions
are not necessary to recognize the different motions. Instead the root position, root velocity, root rotation and the
positions of the subject's extremities proved to be an especially good set of features.

Different configurations and variations of HMMs were evaluated. The main results of this section were that
proper initialization of the emission distribution parameters is very important; less so the initialization
of the transition and start probabilities. The best results were achieved when the means and covariance matrices over
the emission distribution were initialized by first clustering the data using the $k$-means algorithm and then estimating
the means and covariances over these clusters. It further proved necessary to constrain the covariance matrices
to be diagonal. Different topologies and numbers of states were evaluated as well. While a low number of states
($5$ to $8$) produced the best results, the choice of topology was less important. However, the left-to-right
topology proved to be a good choice overall. The performance of FHMMs and HMMs was also compared.
FHMMs did not provide a significant advantage over HMMs for motion classification
on the evaluation dataset. Since FHMMs are computationally more expensive to train and evaluate, HMMs are the better
choice for both systems.

Different decision makers which map the likelihood scores of the HMMs to the multi-label prediction and their hyperparameters were evaluated as well. In general, linear models,
namely Logistic Regression and Support Vector Machines, proved to be especially good choices. Logistic Regression
was further shown to slightly outperform SVMs on the evaluation dataset. An important insight was that both models
worked well if L1 regularization was used. The best regularization coefficients were $C = 10^{-3}$ and $C = 10^{-2}$
for Logistic Regression and SVMs respectively. Decision Trees and Random Forests were also evaluated.
Both were capable of learning the multi-label mapping from likelihoods to labels and Random Forests outperformed
a single Decision Tree. The best splitting criterion turned out to be the information gain and a maximum tree
depth of $15$ delivered consistently good results. $40$ Decision Trees were used in a Random Forest to achieve
good results. However, the linear models clearly outperformed the Decision Tree and Random Forest during
evaluation.

The end-to-end evaluation of the two systems revealed that both can be used to accurately classify
human whole-body motion into multiple classes. The power set system achieved a total accuracy of $98.02\%$ on the test
dataset whereas the multi-label system achieved $93.39\%$. However, while the power set is limited
by the number of label combinations within a dataset, the multi-label system does not have this constraint.
This makes the multi-label system a potentially very interesting approach for classification tasks
where the number of label combinations is much greater than the number of labels.

In future work, the system devised in this thesis could be extended in a couple of ways.
An interesting extensions would be Parametric Hidden Markov Models. PHMMs could be used similarly
to the work in~\cite{herzog:2008motion} to recognize different variations of the same motion. More concretely,
instead of using different HMMs for the classes ``fast'', ``medium'' and ``slow'', a single PHMM could potentially
be used to replace three HMMs.

Another extension would be to use a hierarchical tree structure of HMMs similar to~\cite{Kulic:2008ib}. This tree structure
would have several advantages: Most importantly, the classification speed could be reduced if not all but only a couple of HMMs
need to be considered for each unknown motion. However, while traversing the tree from root to leaf is trivial
for single-label classification, it becomes less obvious how the tree search could be realized for a multi-label problem.
A first idea would be to use supervised learning to not only train the HMMs but to also learn rules when to cut off
a subtree from further consideration. Using a tree structure would also allow to use a hybrid of HMMs and FHMMs.
The sequential learning algorithm could then be used to train additional chains if and only if necessary.

Bayesian optimization is an interesting topic that could be used to find better feature subsets similar to the work in~\cite{inza2000feature}.
The same basic idea can be applied to tune the hyperparameters of the classifier~\cite{NIPS2012_4522}.

The two systems could also be further evaluated on a larger dataset with more motions performed by
more subjects. An especially interesting evaluation
would be on a dataset where much more label combinations than labels exist. On such a dataset, the multi-label
system could potentially outperform the power set system due to the high number of label combinations. Additionally,
the effect of sparse training data could be evaluated properly on such a dataset.

Lastly, the developed system could be deployed for usage in the KIT Whole-Body Human Motion Database.
The system could then be used to automatically label new motion data and maybe even to detect inconsistencies
in the already labeled data. Furthermore, the integration would offer interesting insights on how the system
performs on a real-world problem. Additional considerations would be how the system could be integrated
into the existing database source code as well as adding user interface elements to check and, if necessary,
correct the predictions of the classifier.

\bibliographystyle{alpha}
\bibliography{thesis}

\newcommand{\etalchar}[1]{$^{#1}$}
\begin{thebibliography}{KMN{\etalchar{+}}02}

\bibitem[AAD07]{Azad:2007unifiedrepr}
Pedram Azad, Tamim Asfour, and R{\"u}diger Dillmann.
\newblock Toward an unified representation for imitation of human motion on
  humanoids.
\newblock In {\em Robotics and Automation, 2007 IEEE International Conference
  on}, pages 2558--2563. IEEE, 2007.

\bibitem[ARA{\etalchar{+}}06]{Asfour:2006armar}
Tamim Asfour, Kristian Regenstein, Pedram Azad, Joachim Schr{\"o}der, Alexander
  Bierbaum, Niko Vahrenkamp, and R{\"u}diger Dillmann.
\newblock {ARMAR-III: An integrated humanoid platform for sensory-motor
  control}.
\newblock In {\em Humanoid Robots, 2006 6th IEEE-RAS International Conference
  on}, pages 169--175. IEEE, 2006.

\bibitem[B{\etalchar{+}}06]{Bishop:2006pattern}
Christopher~M Bishop et~al.
\newblock {\em {Pattern Recognition and Machine Learning}}, volume~4.
\newblock springer New York, 2006.

\bibitem[BAG92]{Buchholz:1992anthropometric}
Bryan Buchholz, Thomas~J Armstrong, and Steven~A Goldstein.
\newblock Anthropometric data for describing the kinematics of the human hand.
\newblock {\em Ergonomics}, 35(3):261--273, 1992.

\bibitem[Bak76]{bakis:1976continuous}
Raimo Bakis.
\newblock Continuous speech recognition via centisecond acoustic states.
\newblock {\em The Journal of the Acoustical Society of America},
  59(S1):S97--S97, 1976.

\bibitem[BBG{\etalchar{+}}05]{breazeal:2005learning}
Cynthia Breazeal, Daphna Buchsbaum, Jesse Gray, David Gatenby, and Bruce
  Blumberg.
\newblock Learning from and about others: Towards using imitation to bootstrap
  the social understanding of others by robots.
\newblock {\em Artificial life}, 11(1-2):31--62, 2005.

\bibitem[BCDS08]{Billard:2008kb}
Aude Billard, Sylvain Calinon, R{\"u}diger Dillmann, and Stefan Schaal.
\newblock Robot programming by demonstration.
\newblock In {\em Springer handbook of robotics}, pages 1371--1394. Springer,
  2008.

\bibitem[BE{\etalchar{+}}67]{Baum:1967bf1}
Leonard~E Baum, John~Alonzo Eagon, et~al.
\newblock An inequality with applications to statistical estimation for
  probabilistic functions of markov processes and to a model for ecology.
\newblock {\em Bull. Amer. Math. Soc}, 73(3):360--363, 1967.

\bibitem[BFSO84]{breiman:1984classification}
Leo Breiman, Jerome Friedman, Charles~J Stone, and Richard~A Olshen.
\newblock {\em Classification and regression trees}.
\newblock CRC press, 1984.

\bibitem[BLSB04]{boutell:2004learning}
Matthew~R Boutell, Jiebo Luo, Xipeng Shen, and Christopher~M Brown.
\newblock Learning multi-label scene classification.
\newblock {\em Pattern recognition}, 37(9):1757--1771, 2004.

\bibitem[BPP{\v{S}}14]{babivc:2014effects}
Jan Babi{\v{c}}, Tadej Petri{\v{c}}, Luka Peternel, and Nejc {\v{S}}arabon.
\newblock Effects of supportive hand contact on reactive postural control
  during support perturbations.
\newblock {\em Gait \& posture}, 40(3):441--446, 2014.

\bibitem[BPSW70]{Baum:1970maximization}
Leonard~E Baum, Ted Petrie, George Soules, and Norman Weiss.
\newblock A maximization technique occurring in the statistical analysis of
  probabilistic functions of markov chains.
\newblock {\em The annals of mathematical statistics}, pages 164--171, 1970.

\bibitem[Bre01]{breiman:2001random}
Leo Breiman.
\newblock Random forests.
\newblock {\em Machine learning}, 45(1):5--32, 2001.

\bibitem[BS{\etalchar{+}}68]{Baum:1968bf2}
Leonard~E Baum, George~R Sell, et~al.
\newblock Growth transformations for functions on manifolds.
\newblock {\em Pacific J. Math}, 27(2):211--227, 1968.

\bibitem[c3d]{c3d}
{The C3D file format}.
\newblock \url{http://c3d.org}.
\newblock [accessed May 20th 2015].

\bibitem[CGB07]{calinon:2007learning}
Sylvain Calinon, Florent Guenter, and Aude Billard.
\newblock On learning, representing, and generalizing a task in a humanoid
  robot.
\newblock {\em Systems, Man, and Cybernetics, Part B: Cybernetics, IEEE
  Transactions on}, 37(2):286--298, 2007.

\bibitem[Cra05]{craig:2005introduction}
John~J Craig.
\newblock {\em Introduction to robotics: mechanics and control}, volume~3.
\newblock Pearson Prentice Hall Upper Saddle River, 2005.

\bibitem[DRE{\etalchar{+}}00]{dillmann:2000learning}
R{\"u}diger Dillmann, Oliver Rogalla, Markus Ehrenmann, R~Zollner, and Monica
  Bordegoni.
\newblock Learning robot behaviour and skills based on human demonstration and
  advice: the machine learning paradigm.
\newblock In {\em ROBOTICS RESEARCH-INTERNATIONAL SYMPOSIUM-}, volume~9, pages
  229--238, 2000.

\bibitem[EAM08]{Elliott:1995HMM}
Robert~J Elliott, Lakhdar Aggoun, and John~B Moore.
\newblock {\em Hidden Markov models: estimation and control}, volume~29.
\newblock Springer Science \& Business Media, 2008.

\bibitem[Efr79]{efron:1979bootstrap}
Bradley Efron.
\newblock Bootstrap methods: another look at the jackknife.
\newblock {\em The annals of Statistics}, pages 1--26, 1979.

\bibitem[FCH{\etalchar{+}}08]{fan:2008liblinear}
Rong-En Fan, Kai-Wei Chang, Cho-Jui Hsieh, Xiang-Rui Wang, and Chih-Jen Lin.
\newblock Liblinear: A library for large linear classification.
\newblock {\em The Journal of Machine Learning Research}, 9:1871--1874, 2008.

\bibitem[GE03]{guyon:2003introduction}
Isabelle Guyon and Andr{\'e} Elisseeff.
\newblock An introduction to variable and feature selection.
\newblock {\em The Journal of Machine Learning Research}, 3:1157--1182, 2003.

\bibitem[GJ97]{Ghahramani:1997id}
Zoubin Ghahramani and Michael~I Jordan.
\newblock Factorial hidden markov models.
\newblock {\em Machine learning}, 29(2-3):245--273, 1997.

\bibitem[GS{\etalchar{+}}84]{givens:1984class}
Clark~R Givens, Rae~Michael Shortt, et~al.
\newblock A class of wasserstein metrics for probability distributions.
\newblock {\em Michigan Math. J}, 31(2):231--240, 1984.

\bibitem[HAHR01]{huang:2001spoken}
Xuedong Huang, Alex Acero, Hsiao-Wuen Hon, and Raj Reddy.
\newblock {\em Spoken language processing: A guide to theory, algorithm, and
  system development}.
\newblock Prentice Hall PTR, 2001.

\bibitem[HDO{\etalchar{+}}98]{hearst:1998support}
Marti~A. Hearst, Susan~T Dumais, Edgar Osman, John Platt, and Bernhard
  Scholkopf.
\newblock Support vector machines.
\newblock {\em Intelligent Systems and their Applications, IEEE}, 13(4):18--28,
  1998.

\bibitem[HUK08]{herzog:2008motion}
Dennis Herzog, AleNs Ude, and Volker Kr{\"u}ger.
\newblock Motion imitation and recognition using parametric hidden markov
  models.
\newblock In {\em Humanoid Robots, 2008. Humanoids 2008. 8th IEEE-RAS
  International Conference on}, pages 339--346. IEEE, 2008.

\bibitem[ice]{ice}
{The Internet Communications Engine (Ice)}.
\newblock \url{https://zeroc.com/ice.html}.
\newblock [accessed May 22th 2015].

\bibitem[ILES00]{inza2000feature}
I{\~n}aki Inza, Pedro Larra{\~n}aga, Ram{\'o}n Etxeberria, and Basilio Sierra.
\newblock Feature subset selection by bayesian network-based optimization.
\newblock {\em Artificial intelligence}, 123(1):157--184, 2000.

\bibitem[JJT02]{Jacobs:2002fhmmbackfitting}
Robert~A Jacobs, Wenxin Jiang, and Martin~A Tanner.
\newblock Factorial hidden markov models and the generalized backfitting
  algorithm.
\newblock {\em Neural computation}, 14(10):2415--2437, 2002.

\bibitem[Jor02]{jordan:2002discriminative}
A~Jordan.
\newblock On discriminative vs. generative classifiers: A comparison of
  logistic regression and naive bayes.
\newblock {\em Advances in neural information processing systems}, 14:841,
  2002.

\bibitem[KHB{\etalchar{+}}10]{kruger:2010learning}
Volker Kr{\"u}ger, Dennis~L Herzog, Sanmohan Baby, Ales Ude, and Danica Kragic.
\newblock Learning actions from observations.
\newblock {\em Robotics \& Automation Magazine, IEEE}, 17(2):30--43, 2010.

\bibitem[KMN{\etalchar{+}}02]{kanungo:2002efficient}
Tapas Kanungo, David~M Mount, Nathan~S Netanyahu, Christine~D Piatko, Ruth
  Silverman, and Angela~Y Wu.
\newblock An efficient k-means clustering algorithm: Analysis and
  implementation.
\newblock {\em Pattern Analysis and Machine Intelligence, IEEE Transactions
  on}, 24(7):881--892, 2002.

\bibitem[KOL{\etalchar{+}}11]{Kulic:2011incremental}
Dana Kuli{\'c}, Christian Ott, Dongheui Lee, Junichi Ishikawa, and Yoshihiko
  Nakamura.
\newblock Incremental learning of full body motion primitives and their
  sequencing through human motion observation.
\newblock {\em The International Journal of Robotics Research}, page
  0278364911426178, 2011.

\bibitem[KTN07a]{Kulic:2007clustering}
Dana Kuli{\'c}, Wataru Takano, and Yoshihiko Nakamura.
\newblock Incremental on-line hierarchical clustering of whole body motion
  patterns.
\newblock In {\em Robot and Human interactive Communication, 2007. RO-MAN 2007.
  The 16th IEEE International Symposium on}, pages 1016--1021. IEEE, 2007.

\bibitem[KTN07b]{Kulic:2007bf}
Dana Kuli{\'c}, Wataru Takano, and Yoshihiko Nakamura.
\newblock Representability of human motions by factorial hidden markov models.
\newblock In {\em Intelligent Robots and Systems, 2007. IROS 2007. IEEE/RSJ
  International Conference on}, pages 2388--2393. IEEE, 2007.

\bibitem[KTN08]{Kulic:2008ib}
Dana Kuli{\'c}, Wataru Takano, and Yoshihiko Nakamura.
\newblock Incremental learning, clustering and hierarchy formation of whole
  body motion patterns using adaptive hidden markov chains.
\newblock {\em The International Journal of Robotics Research}, 27(7):761--784,
  2008.

\bibitem[Man06]{mann:2006numerically}
Tobias~P Mann.
\newblock Numerically stable hidden markov model implementation.
\newblock {\em An HMM scaling tutorial}, pages 1--8, 2006.

\bibitem[MBJA15]{mandery:2015analyzing}
Christian Mandery, J{\'u}lia Borr{\`a}s, Mirjam J{\"o}chner, and Tamim Asfour.
\newblock Analyzing whole-body pose transitions in multi-contact motions.
\newblock {\em arXiv preprint arXiv:1507.08799}, 2015.

\bibitem[mmm]{mmm:dataformat}
{The MMM motion data format}.
\newblock
  \url{http://h2t-projects.webarchiv.kit.edu/Projects/MMM/Core/dataformat.html}.
\newblock [accessed May 21th 2015].

\bibitem[MTD{\etalchar{+}}15]{Mandery:2015db}
Christian Mandery, {\"O}mer Terlemez, Martin Do, Nikolaus Vahrenkamp, and Tamim
  Asfour.
\newblock The kit whole-body human motion database.
\newblock In {\em IEEE International Conference on Robotics and Automation
  (ICRA), Seattle, USA,(submitted)}, 2015.

\bibitem[NME{\etalchar{+}}04]{nakanishi:2004learning}
Jun Nakanishi, Jun Morimoto, Gen Endo, Gordon Cheng, Stefan Schaal, and Mitsuo
  Kawato.
\newblock Learning from demonstration and adaptation of biped locomotion.
\newblock {\em Robotics and Autonomous Systems}, 47(2):79--91, 2004.

\bibitem[OST05]{ogata:2005open}
Tetsuya Ogata, Shigeki Sugano, and Jun Tani.
\newblock Open-end human--robot interaction from the dynamical systems
  perspective: mutual adaptation and incremental learning.
\newblock {\em Advanced Robotics}, 19(6):651--670, 2005.

\bibitem[PE04]{popovic:2004angular}
Mako Popovic and Amy Englehart.
\newblock Angular momentum primitives for human walking: biomechanics and
  control.
\newblock In {\em Intelligent Robots and Systems, 2004.(IROS 2004).
  Proceedings. 2004 IEEE/RSJ International Conference on}, volume~2, pages
  1685--1691. IEEE, 2004.

\bibitem[psd]{psd}
A simple algorithm for generating positive semi-definite matrices.
\newblock \url{http://stackoverflow.com/questions/619335}.
\newblock [accessed August 31st 2015].

\bibitem[PVG{\etalchar{+}}11]{pedregosa:2011scikit}
Fabian Pedregosa, Ga{\"e}l Varoquaux, Alexandre Gramfort, Vincent Michel,
  Bertrand Thirion, Olivier Grisel, Mathieu Blondel, Peter Prettenhofer, Ron
  Weiss, Vincent Dubourg, et~al.
\newblock Scikit-learn: Machine learning in python.
\newblock {\em The Journal of Machine Learning Research}, 12:2825--2830, 2011.

\bibitem[Rab89]{Rabiner:1989hs}
Lawrence~R Rabiner.
\newblock A tutorial on hidden markov models and selected applications in
  speech recognition.
\newblock {\em Proceedings of the IEEE}, 77(2):257--286, 1989.

\bibitem[RPHF11]{read:2011classifier}
Jesse Read, Bernhard Pfahringer, Geoff Holmes, and Eibe Frank.
\newblock Classifier chains for multi-label classification.
\newblock {\em Machine learning}, 85(3):333--359, 2011.

\bibitem[Sim12]{simonoff:2012smoothing}
Jeffrey~S Simonoff.
\newblock {\em Smoothing methods in statistics}.
\newblock Springer Science \& Business Media, 2012.

\bibitem[SLA12]{NIPS2012_4522}
Jasper Snoek, Hugo Larochelle, and Ryan~P Adams.
\newblock Practical bayesian optimization of machine learning algorithms.
\newblock In F.~Pereira, C.J.C. Burges, L.~Bottou, and K.Q. Weinberger,
  editors, {\em Advances in Neural Information Processing Systems 25}, pages
  2951--2959. Curran Associates, Inc., 2012.

\bibitem[SS00]{schapire:2000boostexter}
Robert~E Schapire and Yoram Singer.
\newblock Boostexter: A boosting-based system for text categorization.
\newblock {\em Machine learning}, 39(2):135--168, 2000.

\bibitem[STV11]{sechidis:2011stratification}
Konstantinos Sechidis, Grigorios Tsoumakas, and Ioannis Vlahavas.
\newblock On the stratification of multi-label data.
\newblock In {\em Machine Learning and Knowledge Discovery in Databases}, pages
  145--158. Springer, 2011.

\bibitem[TH09]{taylor:2009factored}
Graham~W Taylor and Geoffrey~E Hinton.
\newblock Factored conditional restricted boltzmann machines for modeling
  motion style.
\newblock In {\em Proceedings of the 26th annual international conference on
  machine learning}, pages 1025--1032. ACM, 2009.

\bibitem[THN15]{Takano:2015ca}
Wataru Takano, Seiya Hamano, and Yoshihiko Nakamura.
\newblock Correlated space formation for human whole-body motion primitives and
  descriptive word labels.
\newblock {\em Robotics and Autonomous Systems}, 66:35--43, 2015.

\bibitem[THR06]{taylor:2006modeling}
Graham~W Taylor, Geoffrey~E Hinton, and Sam~T Roweis.
\newblock Modeling human motion using binary latent variables.
\newblock In {\em Advances in neural information processing systems}, pages
  1345--1352, 2006.

\bibitem[TIKN10]{takano:2010organization}
Wataru Takano, Hirotaka Imagawa, Dana Kuli{\'c}, and Yoshihiko Nakamura.
\newblock Organization of behavioral knowledge from extraction of
  temporal-spatial features of human whole body motions.
\newblock In {\em Biomedical Robotics and Biomechatronics (BioRob), 2010 3rd
  IEEE RAS and EMBS International Conference on}, pages 52--57. IEEE, 2010.

\bibitem[TN15a]{takano:2015construction}
Wataru Takano and Yoshihiko Nakamura.
\newblock Construction of a space of motion labels from their mapping to
  full-body motion symbols.
\newblock {\em Advanced Robotics}, 29(2):115--126, 2015.

\bibitem[TN15b]{takano:2015statistical}
Wataru Takano and Yoshihiko Nakamura.
\newblock Statistical mutual conversion between whole body motion primitives
  and linguistic sentences for human motions.
\newblock {\em The International Journal of Robotics Research}, page
  0278364915587923, 2015.

\bibitem[TUM{\etalchar{+}}14]{Terlemez:2014master}
{\"O}mer Terlemez, Stefan Ulbrich, Christian Mandery, Martin Do, Nikolaus
  Vahrenkamp, and Tamim Asfour.
\newblock {Master Motor Map (MMM)--framework and toolkit for capturing,
  representing, and reproducing human motion on humanoid robots}.
\newblock In {\em Humanoid Robots (Humanoids), 2014 14th IEEE-RAS International
  Conference on}, pages 894--901. IEEE, 2014.

\bibitem[TYS{\etalchar{+}}06]{Takano:2006primitive}
Wataru Takano, Katsu Yamane, Tomomichi Sugihara, Kou Yamamoto, and Yoshihiko
  Nakamura.
\newblock {Primitive Communication based on Motion Recognition and Generation
  with Hierarchical Mimesis Model}.
\newblock In {\em Robotics and Automation, 2006. ICRA 2006. Proceedings 2006
  IEEE International Conference on}, pages 3602--3609. IEEE, 2006.

\bibitem[vic]{vicon}
{The VICON motion capture system}.
\newblock \url{http://www.vicon.com}.
\newblock [accessed May 20th 2015].

\bibitem[Vit67]{Viterbi:1967decoding}
Andrew~J Viterbi.
\newblock Error bounds for convolutional codes and an asymptotically optimum
  decoding algorithm.
\newblock {\em Information Theory, IEEE Transactions on}, 13(2):260--269, 1967.

\bibitem[vR79]{van:1979information}
CJ~van Rijsbergen.
\newblock {\em Information Retrieval. 1979}.
\newblock Butterworth, 1979.

\bibitem[VSS{\etalchar{+}}08]{vens:2008decision}
Celine Vens, Jan Struyf, Leander Schietgat, Sa{\v{s}}o D{\v{z}}eroski, and
  Hendrik Blockeel.
\newblock Decision trees for hierarchical multi-label classification.
\newblock {\em Machine Learning}, 73(2):185--214, 2008.

\bibitem[was]{wasserstein}
{Wasserstein distance between two Gaussians}.
\newblock
  \url{http://djalil.chafai.net/blog/2010/04/30/wasserstein-distance-between-two-gaussians}.
\newblock [accessed August 24th 2015].

\bibitem[WB99]{wilson:1999phmm}
Andrew~D Wilson and Aaron~F Bobick.
\newblock Parametric hidden markov models for gesture recognition.
\newblock {\em Pattern Analysis and Machine Intelligence, IEEE Transactions
  on}, 21(9):884--900, 1999.

\bibitem[Win79]{Winter:1979biomechanics}
David~A Winter.
\newblock {\em Biomechanics of Human Movement}.
\newblock Wiley New York, 1979.

\bibitem[Win09]{Winter:2009biomechanics}
David~A Winter.
\newblock {\em Biomechanics and motor control of human movement}.
\newblock John Wiley \& Sons, 2009.

\bibitem[YYN09]{Yamane:RSS09}
K.~Yamane, Y.~Yamaguchi, and Y.~Nakamura.
\newblock Human motion database with a binary tree and node transition graphs.
\newblock In {\em Proceedings of Robotics: Science and Systems}, Seattle, USA,
  June 2009.

\bibitem[ZZ07]{zhang:2007ml}
Min-Ling Zhang and Zhi-Hua Zhou.
\newblock Ml-knn: A lazy learning approach to multi-label learning.
\newblock {\em Pattern recognition}, 40(7):2038--2048, 2007.

\end{thebibliography}
\cleardoublepage
\end{document}